\documentclass{tlp}
\usepackage{amsfonts}
\usepackage{aopmath}
\usepackage{epsfig,graphicx}

\newtheorem{definition}{Definition} 
\newtheorem{example}{Example} 

\begin{document}
\bibliographystyle{tlp}

\long\def\comment#1{}

\title{Handling Defeasibilities in Action Domains}

\author[Y. Zhang]
{YAN ZHANG \\
School of Computing and Information Technology\\
University of Western Sydney\\
Locked Bag 1797, Penrith South DC\\
NSW 1797, Australia    \\
E-mail: yan@cit.uws.edu.au
}

\pagerange{\pageref{firstpage}--\pageref{lastpage}}
\volume{\textbf{10} (3):}
\jdate{March 2002}
\setcounter{page}{1}
\pubyear{2002}

\maketitle
\label{firstpage}

\begin{abstract}

Representing defeasibility is an important 
issue in common sense reasoning. In reasoning about action and change, 
this issue becomes more difficult because 
domain and action related
defeasible information may conflict with general 
inertia rules. Furthermore, different types of defeasible information
may also interfere with each other during the reasoning.  
In this paper, we develop a prioritized logic programming
approach to handle defeasibilities in reasoning about action. 
In particular, we propose three action languages 
${\cal AT}^{0}$, ${\cal AT}^{1}$ and ${\cal AT}^{2}$ which handle
three types of defeasibilities in action domains named 
defeasible constraints, defeasible observations and actions
with defeasible and abnormal effects respectively.
Each language with a higher superscript can be viewed as an extension of 
the language with a lower superscript.
These action languages inherit the simple syntax of ${\cal A}$ 
language but their 
semantics is developed in terms of transition systems where 
transition functions are defined based on
prioritized logic programs.
By illustrating various examples, we show that our approach
eventually provides a powerful mechanism to handle
various defeasibilities in
temporal prediction and postdiction.
We also investigate
semantic properties of these three action languages and characterize
classes of action domains that present more desirable solutions in 
reasoning about action within the underlying action languages. 


\end{abstract}

\section{Introduction}

Representing defeasibility is an important issue in common sense reasoning.
In reasoning about action, this issue becomes more difficult
because domain and action related defeasible information
may conflict with general inertia rules -- that are necessary to specify
things that persist with respect to actions and usually defeasible as well.
Furthermore, different types of defeasible information may also
interfere with each other during the reasoning.
Therefore, most previous 
action theories usually ignored such defeasible information in
problem domains.
However, recent work on causality reveals that
in many situations defeasibility plays an important role in 
temporal prediction and postdiction 
and ignoring this issue may cause difficulties 
in deriving correct solutions in reasoning about action. 

Let us consider the Switch-Power domain that was first 
addressed in \cite{z:comm96}, where two domain constraints were
taken into account:
\begin{eqnarray}
\mbox{\em if the switch is on, then the light is usually on;}
\label{1.1}
\end{eqnarray}
\begin{eqnarray}
\mbox{\em if there is no power, then the light is not on.}
\label{1.2}
\end{eqnarray}
Intuitively, the first constraint is defeasible from our common sense. 
For instance , even if the switch is on, the light might not be on if there is
no power, or there is a problem in the circuit, and so on. But 
if this constraint is not expressed as a defeasible rule, we may have
a difficulty in our reasoning. Suppose we simply represent the above two
constraints as logical implications
$Switch \supset On$ and $\neg Power\supset \neg On$ respectively. If the initial 
state is $\{On, Power, Switch\}$ and the robot is asked to perform an action
$Cut$-$Power$ with effect $\neg Power$
(e.g. a fire alarm leads the robot to perform this action).
Clearly, $Cut$-$Power$ will cause a direct effect $\neg Power$, and then from 
constraint $Switch\supset On$ and $\neg Power\supset \neg On$, an indirect
effect $\neg Switch$ is derived. Obviously, this effect is not
quite reasonable from our intuition as cutting off the power 
should be irrelevant to the switch's position.

People may argue that the above problem is due to the duality of
logical implication
(i.e. $A\supset B\equiv \neg B\supset\neg A$). Now suppose we
adopt McCain and Turn's causal theory \cite{mt:ijcai95} where
constraints (\ref{1.1}) and (\ref{1.2}) are represented as inference rules
$Switch\Rightarrow On$ and $\neg Power\Rightarrow\neg On$
respectively\footnote{Informally, $A\Rightarrow B$ represents a semantics
like ``if $A$ then $B$'', from which we cannot derive $\neg B\Rightarrow \neg A$.
See \cite{mt:ijcai95} for detail.}.
Then under the same initial state as above, it turns out that action
$Cut$-$Power$ becomes unexecutable because the effect $\neg Power$
together with rule $\neg Power\Rightarrow\neg On$ contradicts fact $On$ which
is derivable from fact $Switch$ and rule $Switch\Rightarrow On$. 
This is not a desirable solution either.

The above example just
illustrates one type of defeasibility - defeasible constraints,
which causes difficulties in reasoning about action.  
In fact, there are other types of defeasible information, 
such as defeasible observations and
actions with defeasible and abnormal effects, 
that also significantly influence
temporal prediction and postdiction.
Although the problem of defeasibilities
has been investigated by some researchers recently, e.g. 
\cite{cl:action97,gef:ijcai97,jlm:action96,z:comm96}, none of the 
previous proposals
is completely satisfactory in terms of
representing and handling different types of defeasibilities 
in temporal reasoning
(we will discuss this issue in section 7).

In this paper, we address three basic types of defeasible information
related to temporal prediction and postdiction where incomplete
information is allowable:
defeasible constraints, defeasible observations and actions with
defeasible and abnormal effects.
Our goal is to handle these three types of defeasibilities 
in reasoning about action under a unified framework of logic programming.

The issue of representing action in 
logic programming languages is not new. It
was explored by some researchers previously, e.g. \cite{ko:89}. However,
probably Gelfond and Lifschitz's work 
\cite{gl:action} was the first time to
make a major progress in this direction. By introducing a simple
action language ${\cal A}$, 
Gelfond and Lifschitz's action formulation was able to
deal with both temporal prediction and postdiction, 
while properties of actions were characterized by translating
action language ${\cal A}$ into the language of
extended logic programs \cite{gl:elp}. In other words, in 
Gelfond and Lifschitz's formulation, extended logic program was
used as an implementation of the high level
action language ${\cal A}$. 

It has been recognized that logic programming can not only be used as
the implementation of a high level action language, but also can be
used as a basis for providing a formal semantics of the high level
language, e.g. \cite{cl:action97}. In this paper, we further 
demonstrate that prioritized logic programming has a great flexibility to
serve as a semantic basis to develop high level action languages that
handle various information conflicts in reasoning about action.  
The paper is organized as follows. Section 2 briefly reviews
the concept of prioritized logic programs.
Section 
3 proposes a simple action language ${\cal AT}^{0}$ which can represent
actions in domains with defeasible constraints. The syntax of ${\cal AT}^{0}$
is similar to that of ${\cal A}$ style action languages. A transition
system is proposed to provide a formal semantics of ${\cal AT}^{0}$, where
a corresponding
prioritized logic program is employed as a basis for defining 
such a transition system.
Section 4 then extends action language ${\cal AT}^{0}$ to ${\cal AT}^{1}$
so that it can represent
defeasible observations and shows how it handles the problem of
temporal postdiction under the occurrence of defeasible observations.
Section 5 further generalizes ${\cal AT}^{1}$ to action language
${\cal AT}^{2}$ to represent actions with defeasible and abnormal effects.
Section 6 then investigates various properties of action languages
${\cal AT}^{0}$, ${\cal AT}^{1}$ and ${\cal AT}^{2}$ and characterize specific
classes of action domains 
that may present desirable solutions in reasoning about
action. 
Section 7 discusses related work, and finally section 8 concludes
the paper with some remarks.  

\section{Prioritized Logic Programs (PLPs): An Overview}

We first
introduce the extended logic program and its answer set semantics
developed by Gelfond and Lifschitz \cite{gl:elp}.
A language $\cal L$ of extended
logic programs is determined by its object constants, function
constants and predicate constants. {\em Terms} are
built as in the corresponding first order language; {\em atoms}
have the form $P(t_{1},\cdots,t_{n})$, where
$t_{i}$ ($1\leq i\leq n$) is a term and $P$ is a predicate constant of arity $n$;
a {\em literal} is either an atom $P(t_{1},\cdots,t_{n})$ or
a negative atom $\neg P(t_{1},\cdots,t_{n})$.
A {\em rule} is an expression of the form:
\begin{equation}
L_{0}\leftarrow L_{1},\cdots,L_{m}, not L_{m+1},\cdots, not L_{n},
\label{eq2}
\end{equation}
where each $L_{i}$ ($0\leq i\leq n$) is a literal.
$L_{0}$ is called the {\em head} of the rule,
while $\{L_{1},\cdots,L_{m}$, $not L_{m+1},\cdots$,
$not L_{n}\}$ is called the {\em body} of the rule.
Obviously, the body of a
rule could be empty. We also allow the head of a rule to be empty. In
this case, the rule with an empty head is called {\em constraint}.
A term, atom, literal, or rule is {\em ground} if no variable
occurs in it.
An {\em extended logic program} $\Pi$ is a collection of rules.
$\Pi$ is {\em ground} if each rule in $\Pi$ is
ground.

\comment{

Let $r$ be a ground rule of the form (\ref{eq2}),
we use $pos(r)$ to denote the set of literals in the body of $r$
without negation as failure
$\{L_{1},\cdots,L_{m}\}$, and $neg(r)$ the set of literals
in the body of $r$ with negation as failure $\{L_{m+1},\cdots,L_{n}\}$. We specify
$body(r)$ to be
$pos(r)\cup neg(r)$. We also use $head(r)$ to denote the head of $r$: $\{L_{0}\}$.
Then we use $lit(r)$ to denote $head(r)\cup body(r)$. By extending these
notations, we use $pos(\Pi)$, $neg(\Pi)$,
$body(\Pi)$, $head(\Pi)$, and $lit(\Pi)$ to denote the unions of
corresponding components of all rules in the ground program $\Pi$, e.g.
$body(\Pi)=\bigcup_{r\in\Pi} body(r)$. If $\Pi$ is a non-ground program,
then notions $pos(\Pi)$, $neg(\Pi)$,
$body(\Pi)$, $head(\Pi)$, and $lit(\Pi)$ are defined based on the ground
instantiation (see below definition) of $\Pi$.

}

To evaluate an extended logic program, Gelfond and
Lifschitz proposed an answer set semantics for extended logic
programs.
Let $\Pi$ be a ground extended logic program not containing {\em not}
and {\em Lit} the set of all ground literals in the language of $\Pi$.
An {\em answer set} of $\Pi$
is the smallest subset $S$ of {\em Lit} such that
(i) for any rule $L_{0}\leftarrow L_{1},\cdots,L_{m}$ from
$\Pi$, if $L_{1},\cdots,L_{m}\in S$, then $L_{0}\in S$; and (ii)
if $S$ contains a pair of complementary literals, then
$S=Lit$.
Now let $\Pi$ be a ground arbitrary extended logic program. For any subset $S$
of $Lit$, let $\Pi^{S}$ be the logic program obtained
from $\Pi$ by deleting
(i) each rule that has a formula
{\em not} $L$ in its body with $L\in S$, and
(ii) all formulas of the form {\em not} $L$ in the bodies
of the remaining rules\footnote{We also
call $\Pi^{S}$ the Gelfond-Lifschitz transformation of
$\Pi$ in terms of $S$.}.
We define that $S$ is an {\em answer set} of $\Pi$
iff $S$ is an answer set of $\Pi^{S}$.

For a non-ground extended logic program $\Pi$, we usually view a rule
in $\Pi$ containing variables to be the set of all ground instances of this
rule formed from the set of ground literals in the language. The
collection of all these ground rules forms the
{\em ground instantiation} $\Pi'$ of $\Pi$.
Then a set of ground literals is an answer set of
$\Pi$ if and only if it is an answer set of $\Pi'$.
It is easy to see that an
extended logic program may have one, more than one, or
no answer set at all.

A {\em prioritized logic program} (PLP) ${\cal P}$ is a triple
$(\Pi,{\cal N},<)$, where
$\Pi$ is an extended logic program, $\cal N$ is a
naming function
mapping each rule in $\Pi$ to a name, and $<$ is a strict
partial ordering on names.
The partial ordering $<$ in ${\cal P}$ plays an essential role in
the evaluation of ${\cal P}$.
We also use ${\cal P}(<)$ to denote the set of $<$-relations of ${\cal P}$.
Intuitively $<$ represents a preference
of applying rules during the evaluation of the program.
In particular, if
${\cal N}(r)<{\cal N}(r')$ holds in ${\cal P}$, rule $r$ would be
preferred to apply over rule $r'$ during the evaluation of ${\cal P}$
(i.e. rule $r$ is more preferred than rule $r'$).
Consider the following
classical example represented in our formalism:
\begin{quote}
${\cal P}_{1}=(\Pi,{\cal N},<)$:\\
\hspace*{.1in} $N_{1}: Fly(x)\leftarrow Bird(x)$, {\em not} $\neg Fly(x)$,\\
\hspace*{.1in} $N_{2}: \neg Fly(x)\leftarrow Penguin(x)$, {\em not} $Fly(x)$, \\
\hspace*{.1in} $N_{3}: Bird(Tweety)\leftarrow$,\\
\hspace*{.1in} $N_{4}: Penguin(Tweety)\leftarrow$,\\
\hspace*{.1in} $N_{2}<N_{1}$.
\end{quote}
Obviously, rules $N_{1}$ and $N_{2}$ conflict with each other
as their heads are complementary literals,
and applying $N_{1}$ will defeat
$N_{2}$ and {\em vice versa}.
However, as $N_{2}<N_{1}$, we would expect that rule $N_{2}$ is preferred
to apply first and then defeat rule $N_{1}$ so that
the desired solution $\neg Fly(Tweety)$ can be derived.

\begin{definition}
Let $\Pi$ be a ground extended logic program and $r$ a ground rule of the
form (\ref{eq2})
($r$ does not necessarily belong to $\Pi$).
Rule $r$ is {\em defeated}
by $\Pi$ iff $\Pi$ has an answer set and
for any answer set $S$ of $\Pi$,
there exists some $L_{i}\in S$, where $m+1\leq i\leq n$.
\end{definition}

Now our idea of evaluating a PLP is as follows.
Let ${\cal P}=(\Pi,{\cal N},<)$. If there are two rules
$r$ and $r'$ in $\Pi$ and ${\cal N}(r)<{\cal N}(r')$,
$r'$ will be ignored in the evaluation of ${\cal P}$, {\em only if}
keeping $r$ in $\Pi$ and deleting $r'$
from $\Pi$ will result in a defeat of $r'$.
By eliminating all such potential rules from $\Pi$,
$\cal P$ is eventually reduced to an extended logic program in which
the partial ordering $<$ has been removed. Our evaluation for
${\cal P}$ is then based on this {\em reduced} extended logic program.

Similarly to the case of extended logic programs,
the evaluation of a PLP will be based on its ground form.
We say that a PLP ${\cal P'}=(\Pi',{\cal N'},<')$ is the
{\em ground instantiation} of ${\cal P}=(\Pi,{\cal N},<)$ if
(1) $\Pi'$ is the ground instantiation of $\Pi$; and (2)
${\cal N'}(r_1')<' {\cal N'}(r_2')\in {\cal P'}(<')$ if and only
if there exist rules $r_1$ and $r_2$ in $\Pi$ such that
$r_1'$ and $r_2'$ are ground instances of $r_1$ and $r_2$ respectively and
${\cal N}(r_1) < {\cal N}(r_2)\in {\cal P}(<)$.
Under this definition,
however, we require a restriction on a PLP since
not every PLP's ground instantiation presents a consistent
information with respect to the original PLP.
Consider a PLP as follows:
\begin{quote}
\noindent
$N_{1}: P(f(x))\leftarrow not P(x)$,\\
$N_{2}: P(f(f(x)))\leftarrow not P(f(x))$,\\
$N_{2}<N_{1}$.
\end{quote}
If the only constant in the language is $0$, then the set of ground
instances of $N_{1}$ and $N_{2}$ includes rules like:
\begin{quote}
$N_{1}': P(f(0))\leftarrow not P(0)$,\\
$N_{2}': P(f(f(0)))\leftarrow not P(f(0))$,\\
$N_{3}': P(f(f(f(0))))\leftarrow not P(f(f(0)))$,\\
$\cdots$,
\end{quote}
It is easy to see that $N_{2}'$ can be viewed as an
instance for both $N_{1}$ and $N_{2}$. Therefore, the ordering $<'$
among rules $N_1', N_2',N_3',\cdots$
is no longer a strict partial ordering because of
$N_{2}'<'N_{2}'$.
Obviously, we need to exclude this kind of programs in our context.
On the other hand,
we also want to avoid a situation like
$\cdots <'N_{3}'<'N_{2}'<'N_{1}'$ in the ground
prioritized logic program because this $<'$ indicates that
there is no most preferred rule in the program.

Given a PLP ${\cal P}=(\Pi, {\cal N}, <)$. We say that
${\cal P}$ is {\em well formed} if
there is no rule $r'$ that is an instance of two different rules
$r_{1}$ and $r_{2}$ in $\Pi$ and
${\cal N}(r_{1})<{\cal N}(r_{2}) \in {\cal P}(<)$.
Then it is not difficult to observe that
the following fact holds.

\begin{quote}
{\bf Fact}: If a PLP ${\cal P}=(\Pi, {\cal N}, <)$ is well formed, then
in its ground instantiation ${\cal P'}=(\Pi', {\cal N'}, <')$, $<'$ is
a partial ordering and
every non-empty subset of $\Pi'$ has a least element with respect to $<'$.
\end{quote}

Due to the above fact,
in the rest of this paper, we will only consider well formed PLP programs
in our
discussions, and consequently, the evaluation for an arbitrary
PLP ${\cal P}=(\Pi,{\cal N},<)$  will be based on its ground
instantiation ${\cal P'}=(\Pi', {\cal N'}, <')$. Therefore, in our context
a ground prioritized (or extended) logic program may contain
infinite number of rules. In this case, we will assume that
this ground program is the ground instantiation of
some program that only contains
finite number of rules. In the rest of the paper, whenever
there is no confusion, we will only consider
ground prioritized (extended) logic programs without explicit
declaration.

\begin{definition}
\cite{zf97:logic2}
Let ${\cal P}=(\Pi,{\cal N},<)$ be a prioritized logic program.
${\cal P}^{<}$ is a {\em reduct} of $\cal P$ with respect to $<$
if and only if there exists a sequence of sets $\Pi_{i}$
($i=0, 1, \cdots$) such that:
\begin{enumerate}
\item $\Pi_{0}=\Pi$;
\item $\Pi_{i}=\Pi_{i-1}-\{r_{1},r_{2},\cdots \mid$ (a)
there exists $r\in \Pi_{i-1}$ such that \\
\hspace*{.2in} for every $j$ ($j=1,2, \cdots$),
${\cal N}(r)<{\cal N}(r_{j})\in {\cal P}(<)$ and \\
\hspace*{.2in} $r_{1}, r_2, \cdots$
are defeated by $\Pi_{i-1}-\{r_{1}, r_{2}, \cdots\}$, and (b) there \\
\hspace*{.2in} are no rules $r',r'', \cdots \in\Pi_{i-1}$
such that $N(r_{j})<N(r')$, \\
\hspace*{.2in} $N(r_{j})<N(r''),\cdots$
for some $j$ ($j=1, 2, \cdots$)
and $r', r'', \cdots$ \\
\hspace*{.2in} are defeated by $\Pi_{i-1}-\{r', r'', \cdots\} \}$;
\item ${\cal P}^{<}=\bigcap_{i=0}^{\infty} \Pi_{i}$.
\end{enumerate}
\end{definition}

In Definition 2,
${\cal P}^{<}$ is an extended logic program
obtained from
$\Pi$ by eliminating some rules from $\Pi$.  In particular,
if ${\cal N}(r)<{\cal N}(r_{1})$, ${\cal N}(r)<{\cal N}(r_{2})$,
$\cdots$,
and $\Pi_{i-1}-\{r_{1}, r_{2},\cdots\}$ defeats $\{r_{1}, r_{2},\cdots\}$, then rules
$r_{1},r_{2},\cdots$
will be eliminated from $\Pi_{i-1}$ if no {\em less preferred rule} can be
eliminated (i.e. conditions (a) and (b)).
This procedure is continued until a fixed point is reached.
It should be noted that condition (b) in the above definition is
necessary because without it some unintuitive
results may be derived. For instance, consider ${\cal P}_{1}$ again,
if we add additional preference $N_3<N_2$ in ${\cal P}_{1}$, then using
a modified version of Definition 2 without condition (b),
\begin{quote}
$\{Fly(Tweety)\leftarrow Bird(Tweety), not \neg Fly(Tweety)$, \\
\hspace*{.2in} $Bird(Tweety)\leftarrow$,\\
\hspace*{.2in} $Penguin(Tweety)\leftarrow\}$
\end{quote}
is a reduct of ${\cal P}_1$, from which we will
conclude that Tweety can fly.

\begin{theorem}
Every PLP has a reduct.
\end{theorem}

\begin{definition}
\cite{zf97:logic2}
Let ${\cal P}=(\Pi,{\cal N},<)$ be a PLP and $Lit$ the set of all
ground literals in the language of $\cal P$. For any subset $S$ of $Lit$, $S$
is an {\em answer set} of $\cal P$ iff
$S$ is an answer set for some reduct ${\cal P}^{<}$ of ${\cal P}$.
\end{definition}

\begin{example}
Using Definitions 2 and 3, it is easy to conclude that ${\cal P}_{1}$
has a unique reduct as follows:
\begin{quote}
${\cal P}_{1}^{<}=\{\neg Fly(Tweety)\leftarrow Penguin(Tweety)$,
{\em not} $Fly(Tweety)$,\\
\hspace*{.5in} $Bird(Tweety)\leftarrow$, \\
\hspace*{.5in} $Penguin(Tweety)\leftarrow\}$,
\end{quote}
from which we obtain the following answer set of ${\cal P}_{1}$:
\begin{quote}
$S=\{Bird(Tweety)$, $Penguin(Tweety)$, $\neg Fly(Tweety)\}$.
\end{quote}

Now we consider another program ${\cal P}_{2}$:

\begin{quote}
$N_{1}: A\leftarrow$,\\
\hspace*{.1in} $N_{2}: B\leftarrow$ {\em not} $C$,\\
\hspace*{.1in} $N_{3}: D\leftarrow$,\\
\hspace*{.1in} $N_{4}: C\leftarrow$ {\em not} $B$,\\
\hspace*{.1in} $N_{1}<N_{2}, N_{3}<N_{4}$.
\end{quote}

According to Definition 2, it is easy to see that
${\cal P}_{2}$ has two reducts:
\begin{quote}
$\{A\leftarrow$, \hspace*{.05in}
$D\leftarrow$,
\hspace*{.05in} $C\leftarrow$ {\em not} $B\}$, and\\
\hspace*{.1in}  $\{A\leftarrow$,
\hspace*{.05in} $B\leftarrow$ {\em not} $C$,\
\hspace*{.05in} $D\leftarrow\}$.
\end{quote}
From Definition 3, it follows that ${\cal P}_{2}$ has two answer sets:
$\{A,C,D\}$ and $\{A,B,D\}$.
\end{example}

\comment{

We first introduce the
extended logic program and its answer set semantics defined by
Gelfond and Lifschitz \cite{gl:elp}.
A language $\cal L$ of extended
logic programs is determined by its object constants, function
symbols and predicate symbols. {\em Terms} are
built as in the corresponding first order language; {\em atoms}
have the form $P(t_{1},\cdots,t_{n})$, where
$t_{i}$ ($1\leq i\leq n$) is a term and $P$ is a predicate symbol of arity $n$;
a {\em literal} is either an atom $P(t_{1},\cdots,t_{n})$ or
a negative atom $\neg P(t_{1},\cdots,t_{n})$.
A {\em rule} is an expression of the form:
\begin{equation}
L_{0}\leftarrow L_{1},\cdots,L_{m}, not L_{m+1},\cdots, not L_{n},
\label{eq2}
\end{equation}
where each $L_{i}$ ($0\leq i\leq n$) is a literal.
$L_{0}$ is called the {\em head} of the rule,
while $\{L_{1},\cdots,L_{m}$,{\em not} $L_{m+1},\cdots$,
{\em not} $L_{n}\}$ is called the {\em body} of the rule.
Obviously, the body of a
rule could be empty.
A term, atom, literal, or rule is {\em ground} if no variable
occurs in it.
An {\em extended logic program} $\Pi$ is a collection of rules.

\comment{

Gelfond and
Lifschitz proposed the answer set semantics for extended logic
programs. For simplicity,
we treat a rule $r$ in program $\Pi$ with variables as
the set of all ground instances
of $r$ formed from the set of ground literals of the language of $\Pi$.
In the rest of the paper, we will not explicitly declare this assumption whenever
there is no ambiguity in our discussion.

Let $\Pi$ be an extended logic program not containing {\em not}
and {\em Lit} be the set of all ground literals in the language of $\Pi$.
The {\em answer set} of $\Pi$, denoted
by $Ans(\Pi)$, is the smallest subset $S$ of {\em Lit} such that
\begin{enumerate}
\item[(i)] for any rule $L_{0}\leftarrow L_{1},\cdots,L_{m}$ from
$\Pi$, if $L_{1},\cdots,L_{m}\in S$, then $L_{0}\in S$;
\item[(ii)] if $S$ contains a pair of complementary literals, then
$S=Lit$.
\end{enumerate}
Now consider an arbitrary extended logic program $\Pi$.
For any subset $S$
of $Lit$, let $\Pi^{S}$ be the logic program obtained
from $\Pi$ by deleting
\begin{enumerate}
\item[(i)] each rule that has a formula
{\em not} $L$ in its body while $L\in S$, and
\item[(ii)] all formulas of the form {\em not} $L$ in the bodies
of the remaining rules.
\end{enumerate}
We define that $S$ is an {\em answer set} of $\Pi$,
denoted by $Ans(\Pi)$,
iff $S$ is an answer set of $\Pi^{S}$, i.e. $S=Ans(\Pi^{S})$.

It is easy to see that an extended logic program may have one, more than one, or
no answer set at all.
We note that a rule of the form (\ref{eq2}) is a kind of default rule
\cite{gl:elp}. Therefore, a conflict between two rules in a program $\Pi$
may occur
and this conflict may lead $\Pi$ to have different answer sets.
For instance, the program $\Pi=\{A\leftarrow$ {\em not} $B$,
$B\leftarrow$ {\em not} $A\}$ has two answer sets $\{A\}$ and
$\{B\}$ respectively. In many applications, however,
we need to solve such a conflict.
A common idea of solving conflicts in logic programming is to add priority
in the underlying reasoning mechanism of the logic program. Now we
introduce prioritized logic programs (PLPs) which
extend Gelfond and Lifschitz's extended logic programs \cite{gl:elp}
by adding priority into programs.

The language ${\cal L}^{P}$ of PLPs is a language
${\cal L}$ of extended logic programs with the following augments:\\
- {\em Names:} $N,N_{1},N_{2},\cdots$.\\
- A strict partial ordering (i.e. antireflexive,
antisymmetric and transitive) $<$ on names.\\
- A naming function ${\cal N}$, which maps a rule to a name.

Terms, atoms, literals and rules in PLPs are defined as the same in
extended logic programs.
For the naming function $\cal N$, we require that for any rules
$r$ and $r'$ in a PLP,
${\cal N}(r)={\cal N}(r')$ iff $r$ and $r'$ indicate the same
rule.
A {\em prioritized logic program} ${\cal P}$ is a triplet
$(\Pi,{\cal N},<)$, where
$\Pi$ is an extended logic program, $\cal N$ is a
naming function
mapping each rule in $\Pi$ to a name, and $<$ is a
strict partial ordering on names. 
The following is an example of a prioritized logic program.
\begin{quote}
\hspace*{-.1in} ${\cal P}_{0}$:\\
$N_{1}: P\leftarrow$ {\em not} $Q$, {\em not} $R$,\\
$N_{2}: Q\leftarrow$ {\em not} $P$,\\
$N_{3}: R\leftarrow$ {\em not} $P$,\\
$N_{1}<N_{2}, N_{2}<N_{3}$\footnote{Since $<$ is transitive,
$N_{1}<N_{3}$ should be also included in this part. However, 
for simplicity, whenever there is
no confusion in the context, we usually omit this extra $<$-relation from
our presentation.}.
\end{quote}
We also use notations ${\cal P}_{0}(\Pi)$, ${\cal P}_{0}({\cal N})$, and
${\cal P}_{0}(<)$ to denote the sets of rules, naming function's values and
$<$-relations of ${\cal P}_{0}$ respectively.

The partial ordering $<$ in a PLP ${\cal P}$ plays an essential role in
the evaluation of ${\cal P}$.
Intuitively $<$ represents a preference
of applying rules during the evaluation of the program.
In particular, if
${\cal N}(r)<{\cal N}(r')$ holds in ${\cal P}$, rule $r$ would be
preferred to apply over rule $r'$ during the evaluation of ${\cal P}$
(i.e. rule $r$ is more preferred than rule $r'$).
It is also important to note that
for the case that $r$ and $r'$ include variables,
${\cal N}(r)<{\cal N}(r')$ is viewed as a set of all $<$-relations
${\cal N}(r_{i})<{\cal N}(r'_{i})$ where $r_{i}$ and $r'_{i}$ are ground
instances of $r$ and $r'$ respectively.
Consider the following
classical example represented in our formalism:
\begin{quote}
\hspace*{-.1in} ${\cal P}_{1}$:\\
$N_{1}: Fly(x)\leftarrow Bird(x)$, {\em not} $\neg Fly(x)$,\\
$N_{2}: \neg Fly(x)\leftarrow Penguin(x)$, {\em not} $Fly(x)$, \\
$N_{3}: Bird(Tweety)\leftarrow$,\\
$N_{4}: Penguin(Tweety)\leftarrow$,\\
$N_{2}<N_{1}$.
\end{quote}
Obviously, rules $N_{1}$ and $N_{2}$ conflict with each other
as their heads are complementary literals,
and applying $N_{1}$ will defeat
$N_{2}$ and {\em vice versa}.
However, as $N_{2}<N_{1}$, we would expect that rule $N_{2}$ is preferred
to apply first and then defeat rule $N_{1}$ after applying $N_{2}$ so that
the desired solution $\neg Fly(Tweety)$ can be derived.
This idea is formalized by the following definitions.

\begin{definition}
Let $\Pi$ be a ground extended logic program and $r$ a ground rule of the
form $L_{0}\leftarrow L_{1},\cdots,L_{m}$, {\em\em not} $L_{m+1}, \cdots$,
{\em\em not} $L_{n}$ ($r$ does not necessarily belong to $\Pi$).
Rule $r$ is {\em defeated}
by $\Pi$ iff $\Pi$ has an answer set and
for every answer set $S$ of $\Pi$,
there exists some $L_{i}\in S$, where $m+1\leq i\leq n$.
\end{definition}

Sometimes, it is also convenient to say that a set $S$ of ground literals
{\em defeats} a ground 
rule $r$ if there is some literal $L$ in $S$ and $r$ has a form
$L_{0}\leftarrow\cdots, not L, \cdots$.
Now our idea of evaluating a PLP is as follows.
Let ${\cal P}=(\Pi,{\cal N},<)$. If there are two rules
$r$ and $r'$ in $\Pi$ and ${\cal N}(r)<{\cal N}(r')$,
$r'$ will be ignored in the evaluation of ${\cal P}$, {\em only if}
keeping $r$ in $\Pi$ and deleting $r'$
from $\Pi$ will result in a defeat of $r'$.
By eliminating all such rules from $\Pi$,
$\cal P$ is eventually reduced to an extended logic program in which
the partial ordering $<$ has been removed. Then the answer
set ${\cal P}$ is defined to be the answer set of
this {\em reduced} extended logic program.

The evaluation of a PLP will be based on its ground form. That is,
for any PLP ${\cal P}=(\Pi,{\cal N},<)$, we consider its
{\em ground instantiation} ${\cal P'}=(\Pi',{\cal N'},<')$, where
$\Pi'$, ${\cal N'}$ and $<'$ are ground instantiations of
$\Pi$, ${\cal N}$ and $<$ respectively\footnote{Note that if ${\cal P'}$
is a ground instantiation of ${\cal P}$, then
${\cal N}(r_{1})<{\cal N}(r_{2}) \in {\cal P}(<)$ implies
${\cal N'}(r_{1}')<'{\cal N}(r_{2}') \in {\cal P'}(<')$, where
$r_{1}'$ and $r_{2}'$ are ground instances of $r_{1}$ and $r_{2}$
respectively.}.
However, this requires some restriction on a PLP since
not every PLP's ground instantiation presents a consistent
information with respect to the original PLP.
Consider a PLP as follows:
\begin{quote}
$N_{1}: P(f(x))\leftarrow not P(x)$,\\
$N_{2}: P(f(f(x)))\leftarrow not P(f(x))$,\\
$N_{2}<N_{1}$.
\end{quote}
If the only constant in the language is $0$, then the set of ground
instantiations of $N_{1}$ and $N_{2}$ includes rules as follows:
\begin{quote}
$N_{1}': P(f(0))\leftarrow not P(0)$,\\
$N_{2}': P(f(f(0)))\leftarrow not P(f(0))$,\\
$N_{3}': P(f(f(f(0))))\leftarrow not P(f(f(0)))$,\\
\hspace*{.2in} $\cdots$,
\end{quote}
It is easy to see that $N_{2}'$ can be reviewed as an
instance for both $N_{1}$ and $N_{2}$. Therefore, the ground instantiation $<'$
of $<$ is no longer a partial ordering because of
$N_{2}'<'N_{2}'$.
Obviously, we need to exclude this kind of programs in our context.
On the other hand,
we also want to avoid a situation like
$\cdots <'N_{3}'<'N_{2}'<'N_{1}'$ in the ground
prioritized logic program because this $<'$ indicates that
there does not exist most preferred rule in the program.

Given a PLP ${\cal P}=(\Pi, {\cal N}, <)$. We say
${\cal P}$ is {\em well formed} if
there does not exist a rule $r'$ that is an instance of two different rules
$r_{1}$ and $r_{2}$ in $\Pi$ and
${\cal N}(r_{1})<{\cal N}(r_{2}) \in {\cal P}(<)$.
Then it is not difficult to observe that
the following fact holds.

\begin{quote}
{\bf Fact}: If a PLP ${\cal P}=(\Pi, {\cal N}, <)$ is well formed, then
in its ground instantiation ${\cal P'}=(\Pi', {\cal N'}, <')$, $<'$ is
a partial ordering and
every non-empty subset of $\Pi'$ has a least element with respect to $<'$.
\end{quote}

Due to the above fact,
in the rest of this paper, we will only consider well formed PLPs in our
discussions, and consequently, the evaluation for an arbitrary
program ${\cal P}=(\Pi,{\cal N},<)$  will be based on its ground
instantiation ${\cal P'}=(\Pi', {\cal N'}, <')$. Therefore, in our context
a ground prioritized (or extended) logic program may contain
infinite number of rules. In this case, we will assume that
this ground program is the ground instantiation of
some program that only contains
finite number of rules.

\begin{definition}
Let ${\cal P}=(\Pi,{\cal N},<)$ be a ground prioritized extended logic program.
${\cal P}^{<}$ is a {\em reduct} of $\cal P$ with respect to $<$
if and only if there exists a sequence of sets $\Pi_{i}$
($i=0, 1, \cdots$) such that:
\begin{enumerate}
\item $\Pi_{0}=\Pi$;
\item $\Pi_{i}=\Pi_{i-1}-\{r_{1},r_{2},\cdots \mid$ (a)
there exists $r\in \Pi_{i-1}$ such that for every $j$\\
\hspace*{1.6in} ($j=1,2, \cdots$),
${\cal N}(r)<{\cal N}(r_{j})\in {\cal P}(<)$ and \\
\hspace*{1.6in} $r_{1}, r_{2}, \cdots$
are defeated by $\Pi_{i-1}-\{r_{1}, r_{2}, \cdots\}$, and \\
\hspace*{1.6in} (b) there
does not exist rules $r',r'', \cdots \in\Pi_{i-1}$\\
\hspace*{1.6in} such that $N(r_{j})<N(r')$,
$N(r_{j})<N(r''),\cdots$ \\
\hspace*{1.6in} for some $j$ ($j=1, 2, \cdots$)
and $r', r'', \cdots$ are \\
\hspace*{1.6in} defeated by $\Pi_{i-1}-\{r', r'', \cdots\} \}$;
\item ${\cal P}^{<}=\bigcap_{i=0}^{\infty} \Pi_{i}$.
\end{enumerate}
\end{definition}

\begin{theorem}
Every PLP has a reduct.
\end{theorem}


\begin{definition}
Let ${\cal P}=(\Pi,{\cal N},<)$ be a PLP and $Lit$ the set of all
ground literals in the language of $\cal P$. For any subset $S$ of $Lit$, $S$
is an {\em answer set} of $\cal P$ iff
$S$ is an answer set of some reduct ${\cal P}^{<}$ of ${\cal P}$.
A ground literal $L$ is {\em entailed} from
${\cal P}$, denoted as ${\cal P}\models L$, if $L$ belongs to every answer
set of ${\cal P}$. ${\cal P}$ is called {\em well defined} if it has
a consistent answer set.
\end{definition}

}


To evaluate an extended logic program, Gelfond and
Lifschitz proposed an answer set semantics for extended logic
programs.
Let $\Pi$ be a ground extended logic program not containing {\em not}
and {\em Lit} the set of all ground literals in the language of $\Pi$.
An {\em answer set} of $\Pi$, denoted
as $Ans(\Pi)$, is the smallest subset $S$ of {\em Lit} such that
(i) for any rule $L_{0}\leftarrow L_{1},\cdots,L_{m}$ from
$\Pi$, if $L_{1},\cdots,L_{m}\in S$, then $L_{0}\in S$; and (ii)
if $S$ contains a pair of complementary literals, then
$S=Lit$. 
Now let $\Pi$ be a ground arbitrary extended logic program. For any subset $S$
of $Lit$, let $\Pi^{S}$ be the logic program obtained
from $\Pi$ by deleting
(i) each rule that has a formula
{\em not} $L$ in its body with $L\in S$, and
(ii) all formulas of the form {\em not} $L$ in the bodies
of the remaining rules\footnote{We also
call $\Pi^{S}$ is the Gelfond-Lifschitz transformation of
$\Pi$ in terms of $S$.}.
We define that $S$ is an {\em answer set} of $\Pi$,
denoted by $Ans(\Pi)$,
iff $S$ is an answer set of $\Pi^{S}$, i.e. $S=Ans(\Pi^{S})$.
         
For a non-ground extended logic program $\Pi$, we usually view a rule
in $\Pi$ containing variables to be the set of all ground instances of this
rule formed from the set of ground literals in the language. The
collection of all these ground rules forms the
{\em ground instantiation} $\Pi'$ of $\Pi$.
Then a set of ground literals is an answer set of
$\Pi$ if and only if it is an answer set of $\Pi'$.
It is easy to see that an
extended logic program may have one, more than one, or
no answer set at all.
         
The language ${\cal L}^{P}$ of PLPs is a language ${\cal L}$
of extended logic programs \cite{gl:elp}
with the following augments:\\
- Names: $N,N_{1},N_{2},\cdots$.\\
- A strict partial ordering $<$ on names.\\
- A naming function ${\cal N}$, which maps a rule to a name.\\
A {\em prioritized logic program} (PLP) ${\cal P}$ is a triple
$(\Pi,{\cal N},<)$, where
$\Pi$ is an extended logic program, $\cal N$ is a
naming function
mapping each rule in $\Pi$ to a name, and $<$ is a strict
partial ordering on names.
The partial ordering $<$ in ${\cal P}$ plays an essential role in
the evaluation of ${\cal P}$.
We also use ${\cal P}(<)$ to denote the set of $<$-relations of ${\cal P}$.
Intuitively $<$ represents a preference
of applying rules during the evaluation of the program.
In particular, if
${\cal N}(r)<{\cal N}(r')$ holds in ${\cal P}$, rule $r$ would be
preferred to apply over rule $r'$ during the evaluation of ${\cal P}$
(i.e. rule $r$ is more preferred than rule $r'$).
Consider the following
classical example represented in our formalism:
\begin{quote}
\hspace*{-.1in} ${\cal P}_{1}=(\Pi,{\cal N},<)$:\\
$N_{1}: Fly(x)\leftarrow Bird(x)$, {\em not} $\neg Fly(x)$,\\
$N_{2}: \neg Fly(x)\leftarrow Penguin(x)$, {\em not} $Fly(x)$, \\
$N_{3}: Bird(Tweety)\leftarrow$,\\
$N_{4}: Penguin(Tweety)\leftarrow$,\\
$N_{2}<N_{1}$.
\end{quote}
Obviously, rules $N_{1}$ and $N_{2}$ conflict with each other
as their heads are complementary literals,
and applying $N_{1}$ will defeat
$N_{2}$ and {\em vice versa}.
However, as $N_{2}<N_{1}$, we would expect that rule $N_{2}$ is preferred
to apply first and then defeat rule $N_{1}$ so that
the desired solution $\neg Fly(Tweety)$ can be derived.
         
\begin{definition}
Let $\Pi$ be a ground extended logic program and $r$ a ground rule of the
form $L_{0}\leftarrow L_{1},\cdots,L_{m}$, {\em\em not} $L_{m+1}, \cdots$,
{\em\em not} $L_{n}$ ($r$ does not necessarily belong to $\Pi$).
Rule $r$ is {\em defeated}
by $\Pi$ iff $\Pi$ has an answer set and
for any answer set $Ans(\Pi)$ of $\Pi$,
there exists some $L_{i}\in Ans(\Pi)$, where $m+1\leq i\leq n$.
\end{definition}
         
Sometimes, it is also convenient to say that a set $S$ of ground literals
{\em defeats} a ground rule $r$ if there is some literal $L$ in $S$ and $r$ has a form
$L_{0}\leftarrow\cdots, not L, \cdots$.
Now our idea of evaluating a PLP is as follows.
Let ${\cal P}=(\Pi,{\cal N},<)$. If there are two rules
$r$ and $r'$ in $\Pi$ and ${\cal N}(r)<{\cal N}(r')$,
$r'$ will be ignored in the evaluation of ${\cal P}$, {\em only if}
keeping $r$ in $\Pi$ and deleting $r'$
from $\Pi$ will result in a defeat of $r'$.
By eliminating all such potential rules from $\Pi$,
$\cal P$ is eventually reduced to an extended logic program in which
the partial ordering $<$ has been removed. Our evaluation for
${\cal P}$ is then based on this {\em reduced} extended logic program.
         
Similarly to the case of extended logic programs,
the evaluation of a PLP will be based on its ground form.
We say a PLP ${\cal P'}=(\Pi',{\cal N'},<')$ is the
{\em ground instantiation} of ${\cal P}=(\Pi,{\cal N},<)$ if
(1) $\Pi'$ is the ground instantiation of $\Pi$; and (2)
${\cal N'}(r_1')<' {\cal N'}(r_2')\in {\cal P'}(<')$ if and only
if there exist rules $r_1$ and $r_2$ in $\Pi$ such that
$r_1'$ and $r_2'$ are ground instances of $r_1$ and $r_2$ respectively and
${\cal N}(r_1) < {\cal N}(r_2)\in {\cal P}(<)$.
Under this definition,
However, we require a restriction on a PLP since
not every PLP's ground instantiation presents a consistent
information with respect to the original PLP.
Consider a PLP as follows:
\begin{quote}
$N_{1}: P(f(x))\leftarrow not P(x)$,\\
$N_{2}: P(f(f(x)))\leftarrow not P(f(x))$,\\
$N_{2}<N_{1}$.
\end{quote}
If the only constant in the language is $0$, then the set of ground
instances of $N_{1}$ and $N_{2}$ includes rules like:
\begin{quote}
$N_{1}': P(f(0))\leftarrow not P(0)$,\\
$N_{2}': P(f(f(0)))\leftarrow not P(f(0))$,\\
$N_{3}': P(f(f(f(0))))\leftarrow not P(f(f(0)))$,\\
\hspace*{.2in} $\cdots$,
\end{quote}
It is easy to see that $N_{2}'$ can be reviewed as an
instance for both $N_{1}$ and $N_{2}$. Therefore, the ordering $<'$
among rules $N_1', N_2',N_3',\cdots$
is no longer a partial ordering because of
$N_{2}'<'N_{2}'$.
Obviously, we need to exclude this kind of programs in our context.
On the other hand,
we also want to avoid a situation like
$\cdots <'N_{3}'<'N_{2}'<'N_{1}'$ in the ground
prioritized logic program because this $<'$ indicates that
there does not exist most preferred rule in the program.
         
Given a PLP ${\cal P}=(\Pi, {\cal N}, <)$. We say
${\cal P}$ is {\em well formed} if
there does not exist a rule $r'$ that is an instance of two different rules
$r_{1}$ and $r_{2}$ in $\Pi$ and
${\cal N}(r_{1})<{\cal N}(r_{2}) \in {\cal P}(<)$.
Then it is not difficult to observe that
the following fact holds.
         
\begin{quote}
{\bf Fact}: If a PLP ${\cal P}=(\Pi, {\cal N}, <)$ is well formed, then
in its ground instantiation ${\cal P'}=(\Pi', {\cal N'}, <')$, $<'$ is
a partial ordering and
every non-empty subset of $\Pi'$ has a least element with respect to $<'$.
\end{quote}
         
Due to the above fact,
in the rest of this paper, we will only consider well formed PLPs in our
discussions, and consequently, the evaluation for an arbitrary
a program ${\cal P}=(\Pi,{\cal N},<)$  will be based on its ground
instantiation ${\cal P'}=(\Pi', {\cal N'}, <')$. Therefore, in our context
a ground prioritized (or extended) logic program may contain
infinite number of rules. In this case, we will assume that
this ground program is the ground instantiation of
some program that only contains
finite number of rules. In the rest of the paper, whenever
there is no confusion, we will only consider
ground prioritized (extended) logic programs without explicit
declaration.
         
\begin{definition}
Let ${\cal P}=(\Pi,{\cal N},<)$ be a prioritized logic program.
${\cal P}^{<}$ is a {\em reduct} of $\cal P$ with respect to $<$
if and only if there exists a sequence of sets $\Pi_{i}$
($i=0, 1, \cdots$) such that:
\begin{enumerate}
\item $\Pi_{0}=\Pi$;
\item $\Pi_{i}=\Pi_{i-1}-\{r_{1},r_{2},\cdots \mid$ (a)
there exists $r\in \Pi_{i-1}$ such that \\
\hspace*{.2in} for every $j$ ($j=1,2, \cdots$),
${\cal N}(r)<{\cal N}(r_{j})\in {\cal P}(<)$ and \\
\hspace*{.2in} $r_{1}, r_2, \cdots$
are defeated by $\Pi_{i-1}-\{r_{1}, r_{2}, \cdots\}$, and (b) there \\
\hspace*{.2in} does not exist rules $r',r'', \cdots \in\Pi_{i-1}$
such that $N(r_{j})<N(r')$, \\
\hspace*{.2in} $N(r_{j})<N(r''),\cdots$
for some $j$ ($j=1, 2, \cdots$)
and $r', r'', \cdots$ \\
\hspace*{.2in} are defeated by $\Pi_{i-1}-\{r', r'', \cdots\} \}$;
\item ${\cal P}^{<}=\bigcap_{i=0}^{\infty} \Pi_{i}$.
\end{enumerate}
\end{definition}

\begin{theorem}
Every PLP has a reduct.
\end{theorem}

\begin{definition}
Let ${\cal P}=(\Pi,{\cal N},<)$ be a PLP and $Lit$ the set of all
ground literals in the language of $\cal P$. For any subset $S$ of $Lit$, $S$
is an {\em answer set} of $\cal P$, denoted as $Ans^{P}({\cal P})$, iff
$S=Ans({\cal P}^{<})$ for some reduct ${\cal P}^{<}$ of ${\cal P}$.
\end{definition}
         
\begin{example}
Using Definitions 1 and 2, it is not
difficult to conclude that ${\cal P}_{1}$
has a unique reduct:
\begin{quote}
${\cal P}_{1}^{<}$:\\
\hspace*{.2in} $\neg Fly(Tweety)\leftarrow Penguin(Tweety)$, {\em not} $Fly(Tweety)$,\\
\hspace*{.2in} $Bird(Tweety)\leftarrow$,\\
\hspace*{.2in} $Penguin(Tweety)\leftarrow$,
\end{quote}
and from Definition 3, it has a unique answer set
\begin{quote}
$Ans^{P}({\cal P}_{1})$ $=$ $\{Bird(Tweety)$, $Penguin(Tweety)$,
$\neg Fly(Tweety)\}$.
\end{quote}
Now we consider another PLP ${\cal P}_{2}$ as follows:
\begin{quote}
\hspace*{-.2in} ${\cal P}_{2}$:\\
$N_{1}: P\leftarrow$,\\
$N_{2}: Q\leftarrow$ {\em not} $R$,\\
$N_{3}: T\leftarrow$,\\
$N_{4}: R\leftarrow$ {\em not} $Q$,\\
$N_{1}<N_{2}, N_{3}<N_{4}$.
\end{quote}
According to Definition 2, it is easy to see that
${\cal P}_{2}$ has two reducts:
\begin{quote}
$P\leftarrow$,\\
$T\leftarrow$, \\
$R\leftarrow$ {\em not} $Q$
\end{quote}
and
\begin{quote}
$P\leftarrow$, \\
$Q\leftarrow$ {\em not} $R$,\\
$T\leftarrow$. 
\end{quote}
From Definition 3, it follows that
${\cal P}_{2}$ has two answer sets $\{P,R,T\}$ and $\{P,Q,T\}$.
\end{example}

}

\section{${\cal AT}^{0}$: Representing Actions in 
             Domains with Defeasible Constraints}

In this section, we develop an action language ${\cal AT}^{0}$ which 
is able to handle domains with defeasible constraints. 
The syntax of language ${\cal AT}^{0}$ is inspired by 
${\cal A}$ family languages, and a transition system
will be developed to provide the semantics of ${\cal AT}^{0}$ where
a corresponding prioritized logic program is employed to 
define the transition function.

\subsection{Syntax of ${\cal AT}^{0}$}

The language ${\cal AT}^{0}$ has two disjoint sets of names called
{\em actions} and {\em fluents}. We will use $A$, $A_{1}$, $A_{2}$, $\cdots$
to denote action names , and $F$, $F_{1}$, $F_{2}$, $\cdots$ to denote 
fluent names. 
We define a {\em fluent expression} to be a fluent name possibly 
preceded by a negation sign $\neg$.

A {\em value proposition} is an expression of the form:
\begin{eqnarray}
L \hspace*{.05in} \mbox{\bf after} \hspace*{.05in} A_{1},\cdots,A_{l},
\label{s1.0}
\end{eqnarray}
where $L$ is a fluent expression and $A_{1},\cdots, A_{l}$ are
action names. 
A value proposition is also called an
{\em initial proposition} if no action name occurs in it:
\begin{eqnarray}
\mbox{\bf initially} \hspace*{.05in} L.
\label{s1.1}
\end{eqnarray}

A {\em causal proposition} is an expression of the form:
\begin{eqnarray}
L \hspace*{.05in} \mbox{\bf is caused if} \hspace*{.05in} 
L_{1},\cdots, L_{m} \hspace*{.05in} \mbox{\bf with absence} \hspace*{.05in}
L_{m+1},\cdots, L_{n},
\label{s1.2}
\end{eqnarray}
where $L$, $L_{1},\cdots,L_{n}$ are fluent expressions. 
This is so-called {\em defeasible constraint} whose intuitive meaning 
is that $L$ is caused to be true
if $L_{1},\cdots,L_{m}$ are true and $L_{m+1},\cdots,L_{n}$ 
are not present. As a special case, (\ref{s1.2}) is reduced to a non-defeasible
causal rule if no absent fluent expression is mentioned:
\begin{eqnarray}
L \hspace*{.05in} \mbox{\bf is caused if} \hspace*{.05in} 
L_{1},\cdots, L_{m}
\label{s1.3}
\end{eqnarray}

An {\em action effect proposition} is an expression of the form:
\begin{eqnarray}
A \hspace*{.05in} \mbox{\bf causes} \hspace*{.05in} L
\hspace*{.05in} \mbox{\bf if} \hspace*{.05in} 
L_{1},\cdots, L_{k},
\label{s1.4}
\end{eqnarray}
where $A$ is an action name and $L_{1},\cdots,L_{k}$ are fluent expressions.
(\ref{s1.4}) means that if preconditions $L_{1},\cdots,L_{k}$ of $A$ are true,
then action $A$ causes $L$ to be true. Note the difference between
(\ref{s1.3}) and (\ref{s1.4}) while no action is involved in the former.

Now we define a {\em domain description} ${\cal D}$ of ${\cal AT}^{0}$ 
to be a finite
set of initial propositions, causal propositions and
action effect propositions. 
It should be noted that here we do not include 
value propositions of the form (\ref{s1.0}) into a domain description 
since at the moment we restrict our formulation only to deal with
prediction reasoning while a value proposition (\ref{s1.0}) is only
used as a {\em query statement} in the language\footnote{This
restriction will be released in language ${\cal AT}^{1}$.}.
The following example shows how language ${\cal AT}^{0}$ is used to 
describe an action domain.

\begin{example}
Let us consider the Switch-Power domain mentioned in section 1 again.
The domain includes
two constraints:
(a) if the switch is on, then the light is usually on; (b) if there is no
power, then the light is not on. We treat the first constraint as a
defeasible causal rule.
We also suppose that initially the light is on, there is power and
the switch is on. An action $Cut$-$Power$ is then performed. It has been
shown that the previous approaches have difficulties to deal with this example
due to a lack of expressibility of defeasible constraints
\cite{z:comm96}. This action scenario can be described by
specifying a domain description ${\cal D}(Switch\mbox{-}Power)$ 
of ${\cal AT}^{0}$ as follows. Firstly,
${\cal D}(Switch\mbox{-}Power)$
contains the following three initial propositions:
\begin{quote}
\noindent
{\bf initially} $On$,\\
{\bf initially} $Power$, \\
{\bf initially} $Switch$.
\end{quote}
${\cal D}(Switch\mbox{-}Power)$ also includes the following two causal
propositions to capture the domain constraints presented above:
\begin{quote}
\noindent
$On$ {\bf is caused if} $Switch$ {\bf with absence} $\neg On$, \\
$\neg On$ {\bf is caused if} $\neg Power$.
\end{quote}
Finally, ${\cal D}(Switch\mbox{-}Power)$ has one action effect proposition:
\begin{quote}
$Cut\mbox{-}Power$ {\bf causes} $\neg Power$.
\end{quote}
\end{example}

\subsection{Semantics of ${\cal AT}^{0}$}

Similarly to the idea presented in \cite{gl:action},
we will define a transition system to provide a formal semantics for
${\cal AT}^{0}$. However, instead of developing an independent transition system for
the language, our transition function is defined based on the PLP. This
is because the PLP has a powerful mechanism of solving conflicts between
defeasible information, which, from our observation, is
difficult to handle in the traditional transition system approach.

\subsubsection{Translating ${\cal AT}^{0}$ into PLP}

We first propose a translation from a domain description ${\cal D}$ of
${\cal AT}^{0}$ into a PLP, and our transition function
will be defined based on
this translated PLP.
To implement this translation, we consider a language
${\cal L}^{P}_{{\cal AT}^{0}}$ of PLPs including the following vocabulary:
\begin{enumerate}
\item[-] {\em Situation sort}: one situation constant $S_{0}$,
and situation variables $s, s_{1},s_{2},\cdots$.
\item[-] {\em Action sort}: action constants $A, A_{1}, A_{2}, \cdots$, 
and action variables $a, a_{1}, a_{2}, \cdots$.
\item[-] {\em Propositional fluent sort}: fluent constants 
$F, F_{1}, F_{2}, \cdots$,
and fluent variables $f, f_{1}, f_{2}, \cdots$.
\item[-] {\em Function symbol}: a binary function symbol {\em Result}
which takes arguments of {\em action} and {\em situation} respectively, and
returns a {\em situation}.
\item[-] {\em Predicate symbols}: five binary predicate symbols {\em Holds},
$Caused^{+}$, $Caused^{-}$, $Effect^{+}$ and $Effect^{-}$, all of
which take arguments of {\em fluent} and {\em situation} respectively.
\end{enumerate}

In ${\cal L}^{P}_{{\cal AT}^{0}}$ 
situation term $Result(a,s)$ indicates the resulting situation after performing
action $a$ in $s$.
Atom $Holds(f,s)$ (or literal $\neg Holds(f,s)$) indicates
the fact that fluent $f$ is true (or false, resp.) in situation $s$.
Atom $Caused^{+}(f,s)$ (or $Caused^{-}(f,s)$ resp.) indicates
that fluent $f$ is {\em caused} to be true (or false, resp.)
in situation $s$.  
{\em Causal rules} in ${\cal L}^{P}_{{\cal AT}^{0}}$ have the 
following forms\footnote{Notation
$[\neg ]$ means that the negation sign $\neg$ may or may not occur.}:

\begin{eqnarray}
Caused^{+}(F,s) & \leftarrow &
[\neg ] Holds(F_{1},s),\cdots, [\neg ] Holds(F_{m},s), \nonumber \\
 & & not \hspace*{.05in} [\neg ] Holds(F_{m+1},s),\cdots, 
not \hspace*{.05in} [\neg ] Holds(F_{n},s), 
\label{3.1}
\end{eqnarray}
\begin{eqnarray}
Caused^{-}(F,s) & \leftarrow &
 [\neg ] Holds(F_{1},s),\cdots, [\neg ] Holds(F_{m},s), \nonumber \\
& & not \hspace*{.05in} [\neg ] Holds(F_{m+1},s),\cdots, 
not \hspace*{.05in} [\neg ] Holds(F_{n},s),  
\label{3.2}
\end{eqnarray}
\begin{eqnarray}
Holds(f,s) & \leftarrow & Caused^{+}(f,s),
\label{3.3}
\end{eqnarray}
\begin{eqnarray}
\neg Holds(f,s) & \leftarrow & Caused^{-}(f,s). 
\label{3.4}
\end{eqnarray}

Basically,
rule (\ref{3.1}) together with rule
(\ref{3.3}) (or  (\ref{3.2}) together with (\ref{3.4}) resp.)
says that if literals $[\neg ]Holds(F_{1},s)$,
$\cdots$, $[\neg ]Holds(F_{m},s)$ are true, and
there is no explicit statement saying that
$[\neg ]Holds(F_{m+1},s)$, $\cdots$, $[\neg ]Holds(F_{n},s)$ are true, then
fluent $F$ is caused to be true (or false resp.)
in situation $s$. As it will be seen, in our following translation,
rules (\ref{3.1}) and (\ref{3.2}) are actually related to 
the domain description ${\cal D}$ and hence are {\em domain specific}, while
rules (\ref{3.3}) and (\ref{3.4}) act as generic rule schemas that
are 
irrelevant to ${\cal D}$ and hence are {\em domain independent}.
For simplicity, we denote 
\begin{eqnarray*}
\Pi_{ind}^{c}=\{(\ref{3.3}),(\ref{3.4})\}.
\end{eqnarray*}

Atoms $Effect^{+}(f,s)$ and $Effect^{-}(f,s)$ are used to represent 
direct effects of actions. Generally, {\em action effect rules} 
have the following forms:
\begin{eqnarray}
Effect^{+}(F,Result(A,s)) & \leftarrow &
[\neg ] Holds(F_{1},s),\cdots, [\neg ] Holds(F_{k},s), 
\label{3.5}
\end{eqnarray}
\begin{eqnarray}
Effect^{-}(F,Result(A,s)) & \leftarrow &
[\neg ] Holds(F_{1},s),\cdots, [\neg ] Holds(F_{k},s), 
\label{3.6}
\end{eqnarray}
\begin{eqnarray}
Holds(f,s)  & \leftarrow & Effect^{+}(f,s),
\label{3.7}
\end{eqnarray}
\begin{eqnarray}
\neg Holds(f,s) & \leftarrow & Effect^{-}(f,s).
\label{3.8}
\end{eqnarray}

Intuitively, rule (\ref{3.5}) together with rule
(\ref{3.7})  (or (\ref{3.6}) together with
(\ref{3.8}) resp.) says
that if action $A$'s preconditions
$[\neg ] Holds(F_{1},s),\cdots$,
$[\neg ] Holds(F_{k},s)$ are true, then
fluent $F$ becomes true (or false resp.) after performing action $A$.
Again, rules (\ref{3.5}) and (\ref{3.6}) are domain specific, while
rules (\ref{3.7}) and (\ref{3.8}) represent domain independent schemas.
Similarly, we denote
\begin{eqnarray*}
\Pi_{ind}^{eff}=\{(\ref{3.7}), (\ref{3.8})\}.
\end{eqnarray*}

\begin{definition}
A PLP is called a {\em translation} of domain description ${\cal D}$ of
${\cal AT}^{0}$, denoted
by ${\cal P}^{{\cal AT}^{0}}({\cal D})=(\Pi,{\cal N},<)$, 
if it is obtained as follows:
\begin{enumerate}
\item $\Pi$ consists of the following rules:\\
\underline{\em Initial fact rules}:
For each initial proposition (\ref{s1.1}) in ${\cal D}$, there is a rule of the form
$[\neg ]Holds(F,S_{0})\leftarrow$\footnote{Here $Holds(F,S_{0})$ or
$\neg Holds(F,S_{0})$ is corresponding to whether $L$ occurring in 
{\bf initially} $L$ is $F$ or $\neg F$ respectively. This 
assumption is also adopted in the rest of this paper.}.\\
\underline{\em Causal rules}:
For each causal proposition (\ref{s1.2}) in ${\cal D}$, there is a causal rule
of the form 
(\ref{3.1}) or (\ref{3.2}). Two domain independent causal rules
(\ref{3.3}) and (\ref{3.4}) are also included in this set.\\
\underline{\em Action effect rules}: 
For each action effect proposition (\ref{s1.4}), there is an action effect rule of
the form (\ref{3.5}) or (\ref{3.6}). 
Two domain independent action effect rules (\ref{3.7}) and (\ref{3.8}) are
also included in this set.\\
\underline{\em Inertia rules}\footnote{Note that 
these two inertia rules actually represent a set of inertia rules by 
substituting fluent and action variables $f$ and $a$ with every 
fluent and action constants occurring in the domain respectively.}:
\begin{eqnarray}
Holds(f,Result(a,s)) & \leftarrow & Holds(f,s),
not \hspace*{.05in} \neg Holds(f,Result(a,s)), 
\label{3.9}
\end{eqnarray}
\begin{eqnarray}
\neg Holds(f,Result(a,s))& \leftarrow & \neg Holds(f,s), 
not \hspace*{.05in} Holds(f,Result(a,s)).
\label{3.10}
\end{eqnarray}

\item Naming function ${\cal N}$ assigns a unique name to each rule in $\Pi$.

\item For each causal rule $N_{c}$ and each inertia rule 
$N_{i}$, $<$-relation $N_{c}<N_{i}$ holds.
\end{enumerate}
\end{definition}

In ${\cal P}^{{\cal AT}^{0}}({\cal D})$ specified above,
$\Pi$ represents
initial facts, domain constraints (causal rules) and action effects corresponding to
${\cal D}$, and inertia rules are used to capture things that do not change 
with respect to actions. 
Since we allow to represent defeasible
causal rules while inertia rules are also defeasible, 
possible conflicts may occur between these two types of rules.
To solve such conflicts, we specify that a causal rule is 
more preferred than an inertia rule.
The intuition behind this is clear: generally causal rules 
are used to derive indirect effects of actions, and whenever
there is no explicit condition to block a defeasible causal rule, 
this rule should be triggered to derive necessary indirect effects.
This point is illustrated in Example 3 next. 

It is also obvious that to translate a 
specific domain description ${\cal D}$ into
${\cal P}^{{\cal AT}^{0}}({\cal D})$, 
we only need to translate domain specific information such as
initial propositions, causal and action effect propositions into
logic program rules, while
other domain independent schema rules such as $\Pi^{c}_{ind}$, $\Pi^{eff}_{ind}$
and rules (\ref{3.9}) and (\ref{3.10}) are automatically
embedded in every translated PLP. Formally,  in a given
${\cal P}^{{\cal AT}^{0}}({\cal D})$,
we denote a set of domain specific rules as $\Pi_{spec}^{0}$, and
specify the set of domain independent rules as
\begin{eqnarray}
\Pi_{ind}^{0}=\Pi_{ind}^{c}\cup\Pi_{ind}^{eff}\cup\Pi_{ind}^{i},
\label{3.ex}
\end{eqnarray}
where $\Pi_{ind}^{i}=\{(\ref{3.9}),(\ref{3.10})\}$.


\begin{example}
Example 2 continued.
According to Definition 4, the domain description ${\cal D}(Switch\mbox{-}Power)$ 
presented in Example 2 can be translated into a PLP, denoted by
${\cal P}^{{\cal AT}^{0}}(Switch\mbox{-}Power)$
$=(\Pi^{0}_{spec}\cup\Pi^{0}_{ind},{\cal N},<)$, 
where $\Pi^{0}_{spec}$ consists of the following rules:\\
\underline{Initial fact rules}:\\
$N_{1}: Holds(On,S_{0})\leftarrow$,\\
$N_{2}: Holds(Power,S_{0})\leftarrow$,\\
$N_{3}: Holds(Switch,S_{0})\leftarrow$,\\
\underline{Causal rules}:\\
$N_{4}: Caused^{+}(On,s)\leftarrow Holds(Switch,s)$, {\em not} $\neg Holds(On,s)$,\\
$N_{5}: Caused^{-}(On,s)\leftarrow \neg Holds(Power,s)$,\\
\underline{Action effect rule}:\\
$N_{6}: Effect^{-}(Power,Result(Cut$-$Power, s))\leftarrow$.\\
\underline{Naming rules in $\Pi_{ind}^{0}$}:\\
Assigning a unique name
to each rule in $\Pi_{ind}^{0}$. That is, we assign names $N_{7}$, $N_{8}$, 
$N_{9}, N_{10}, N_{11}$ and $N_{12}$ to rules (\ref{3.3}), (\ref{3.4}),
(\ref{3.7}) (\ref{3.8}), (\ref{3.9}) and (\ref{3.10}) respectively.\\
\underline{$<$-relations}: \\
$N_{c}<N_{i}$, while $N_{c}$ and $N_{i}$
are names of any causal rule and inertia rule in $\Pi$ respectively.
That is, we have
$\{N_{4},N_{5},N_{7},N_{8}\}<\{N_{11},N_{12}\}$\footnote{This is an abbreviation of 
a set of $<$-relations of the form $N_{i}<N_{j}$, where $i=4,5,7,8$ and $j=11,12$.}.
\end{example}

\subsubsection{Transition function, models and entailment}

To define the transition function,
we first introduce the concept of state. 
A {\em state} is a collection of fluent expressions. A state is {\em consistent} if it does
not contain a fluent $F$ and its negative correspondent $\neg F$. 
We use symbols $\hat{S}_{0}$,
$\hat{S}_{1}$, $\hat{S}_{2}$, $\cdots$ to denote states.
Then transition function ${\cal R}$
maps a state to a power set of states by some action.

\begin{definition}
Given a domain description
${\cal D}$ and its translation ${\cal P}^{{\cal AT}^{0}}({\cal D})$,
let ${\mathbb A}$
be the set of all answer sets of ${\cal P}^{{\cal AT}^{0}}({\cal D})$.
The {\em transition function} ${\cal R}(A,\hat{S})$
of ${\cal D}$ with respect to action $A$ and
state $\hat{S}$ is defined as follows:
\begin{enumerate}
\item If ${\mathbb A}$ is empty or includes an inconsistent answer set of 
${\cal P}^{{\cal AT}^{0}}({\cal D})$, then ${\cal R}(A,\hat{S})$ is undefined; 
\item $\hat{S}_{0}$
$=\{[\neg ] F\mid [\neg ] Holds(F,S_{0})\in Ans\}$, where
$Ans\in {\mathbb A}$;
\item ${\cal R}(A,\hat{S})=\{\{[\neg ]F\mid [\neg ]Holds(F,Result(A,S'))\in Ans$,
and for any $F'$\\
\hspace*{1.3in} 
$[\neg ] F'\in \hat{S}$ iff $[\neg ]Holds(F',S')\in Ans\}\mid Ans\in {\mathbb A} \}$.
\end{enumerate}
\end{definition}

It should be noted that we define a state to be a collection of
fluent expressions, that is very different from the state
defined in standard ${\cal A}$-style action languages where 
states correspond to possible {\em physical} worlds and
every fluent is either true or false in a state \cite{gl:action}. 
In our context, a state may not present a complete information for
fluents. If a fluent is not present in a state, 
then this fluent's truth value is viewed as {\em unknown}. Defining
states in this way will bring us a flexibility to develop a formal
semantics for our action theories where incomplete
information related to defeasibility is admitted. 

In Definition 5, $\hat{S}_{0}$ is called the {\em initial state} of ${\cal D}$, and
${\cal R}(A,\hat{S})$ represents the set 
of all possible states resulting from 
the execution of $A$ on state $\hat{S}$. 
From the feature of ${\cal P}^{{\cal AT}^{0}}({\cal D})$, it is quite obvious that
the initial state $\hat{S}_{0}$ is always unique. On the other hand, 
${\cal R}(A,\hat{S})$ may include more than one state. 
To see how transition function ${\cal R}$ works, we consider a domain ${\cal D}$
consisting of the following propositions:
\begin{quote}
\noindent
{\bf initially} $F_1$,\\
$F_2$ {\bf is caused if} $\neg F_1$ {\bf with absence} $\neg F_2$,\\
$\neg F_2$ {\bf is caused if} $\neg F_1$ {\bf with absence} $F_2$,\\
$A_1$ {\bf causes} $\neg F_1$,\\
$A_2$ {\bf causes} $\neg F_2$ {\bf if} $\neg F_1,F_2$.
\end{quote}

Since two causal propositions conflict with each other and action
$A_1$ is executable in the initial situation, it is not difficult to 
see that ${\cal D}$'s PLP translation ${\cal P}^{{\cal AT}^{0}}({\cal D})$
has two different answer sets such that $Holds(F_2,Result(A_1,S_0))$ is in
one and\\
$\neg Holds(F_2,Result(A_1,S_0))$ is in the other. Then from Definition 5,
state transitions of ${\cal D}$ 
specified by transition function ${\cal R}$
can be described by the following diagram, where
$\{F_1\}$ is the initial state:

\newpage

\begin{figure}[tbhp]
\centerline{\includegraphics[height=8cm]{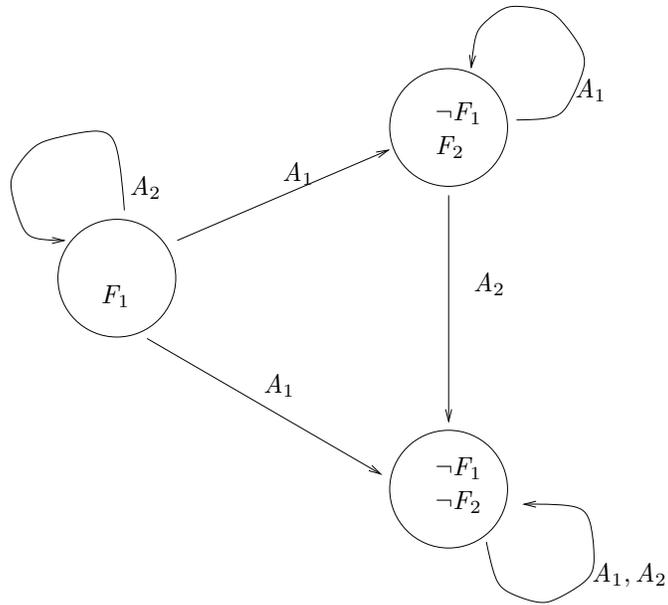}}
\caption{State transitions.}
\label{fig1}
\end{figure}


\vspace*{-2.2in}
\hspace*{1.25in}
$F_1$

\vspace*{-.75in}
\hspace*{1.4in}
$A_2$

\vspace*{-.25in}
\hspace*{2.2in}
$A_1$


\vspace*{-.5in}
\hspace*{3.in}
$\neg F_1$

\hspace*{3.in}
$F_2$

\vspace*{-.48in}
\hspace*{3.73in}
$A_1$


\vspace*{1.8in}
\hspace*{3.in}
$\neg F_1$

\hspace*{3.in}
$\neg F_2$

\vspace*{.2in}
\hspace*{3.82in}
$A_1,A_2$

\vspace*{-1.7in}
\hspace*{3.2in}
$A_2$

\vspace*{.35in}
\hspace*{2.1in}
$A_1$

\newpage

%

Let $\overline{A}$ denote an action
string $A_{1} \cdots A_{l}$ (as a special case, an empty action string
is denoted as $\epsilon$).
A {\em structure} $\Psi$ is a partial function from strings of actions to states
whose domain is prefix closed. We refer
$\Psi(\epsilon)=\hat{S}_{0}$, i.e. the initial state of ${\cal D}$.
The following definition describes possible trajectories
of the dynamic system (domain description) defined in ${\cal AT}^{0}$ 
under structure $\Psi$.

\begin{definition}
Given a structure $\Psi$. 
\begin{enumerate}
\item An initial proposition of the form (\ref{s1.1}) is {\em satisfied} in
$\Psi$ if $L\in \Psi(\epsilon)$;
\item A causal proposition of the form (\ref{s1.2}) or an action effect
proposition of the form (\ref{s1.4}) is {\em satisfied} 
in $\Psi$ if the following conditions hold:
\begin{enumerate}
\item[-]
for any action string $\overline{A}$ and action constant
$A$, if $\Psi(\overline{A})$ and ${\cal R}(A,\Psi(\overline{A}))$ are
defined,
then $\Psi(\overline{A}\cdot A)\in  {\cal R}(A,\Psi(\overline{A}))$; 
\item[-] otherwise
$\Psi(\overline{A}\cdot A)$ is undefined.
\end{enumerate}
\end{enumerate}
A fluent expression $L$ is {\em true} in a state $\Psi(\overline{A})$ if
$L\in \Psi(\overline{A})$.
\end{definition}

%

\begin{definition}
Given a domain description ${\cal D}$, a structure $\Psi$ is a {\em model} of
${\cal D}$ if all initial, causal and action effect propositions in
${\cal D}$ are satisfied in $\Psi$, and for any action string $\overline{A}$ and 
fluent $F$, $F$ and $\neg F$ are not both true in $\Psi(\overline{A})$.
We say a value proposition of the form (\ref{s1.0}):
$L$ {\bf after} $A_{1},\cdots, A_{l}$
is {\em satisfied} in $\Psi$ if $L\in \Psi(\overline{A})$, 
where $\overline{A}=A_{1}\cdots A_{l}$.
We say ${\cal D}$ {\em entails} value proposition (\ref{s1.0}), denoted as
${\cal D}\models_{{\cal AT}^{0}} L$ {\bf after} $A_{1},\cdots, A_{l}$, if 
(\ref{s1.0}) is satisfied in all models of ${\cal D}$.
\end{definition}

\begin{example}
Example 3 continued.
From ${\cal D}(Switch\mbox{-}Power')$'s  PLP 
translation\\
${\cal P}^{{\cal AT}^{0}}(Switch\mbox{-}Power)$ 
as shown in Example 3, it can be verified that
${\cal P}^{{\cal AT}^{0}}(Switch\mbox{-}Power)$ has a unique answer set that includes
the following ground literals\footnote{Obviously
the answer set also includes many other ground literals that
we are not interested in listing here.}:
\begin{quote}
\noindent
$Holds(On,S_{0})$, \\
$Holds(Power,S_{0})$, \\
$Holds(Switch,S_{0})$,\\
$\neg Holds(Power,Result(Cut\mbox{-}Power,S_{0}))$,\\
$\neg Holds(On, Result(Cut\mbox{-}Power,S_{0}))$ and\\
$Holds(Switch, Result(Cut\mbox{-}Power,S_{0}))$.
\end{quote}

Since $\hat{S}_{0}=\{On, Power, Switch\}$, we have
${\cal R}(Cut\mbox{-}Power,\hat{S}_{0})=\{\hat{S}_{1}\}$, where\\
$\hat{S}_{1}=\{\neg On, \neg Power, Switch\}$. Now it is easy to see that 
structure $\Psi$ is a model of ${\cal D}(Switch\mbox{-}Power)$, where
$\Psi(\epsilon)=\hat{S}_{0}$ and $\Psi(Cut\mbox{-}Power)=\hat{S}_{1}$.
Furthermore, according to Definition 7, we have 
\begin{quote}
\noindent
${\cal D}(Switch\mbox{-}Power)\models_{{\cal AT}^{0}} \neg Power$ {\bf after} $Cut\mbox{-}Power$,\\
${\cal D}(Switch\mbox{-}Power)\models_{{\cal AT}^{0}} \neg On$ {\bf after} $Cut\mbox{-}Power$,\\
${\cal D}(Switch\mbox{-}Power)\models_{{\cal AT}^{0}} Switch$ {\bf after} $Cut\mbox{-}Power$.
\end{quote}

Now we slightly modify the domain of Switch-Power as stated
in Example 2. Suppose initially the light is not on and the switch is off,
and another action $Turn\mbox{-}On$ is also 
available. Then the modified domain description 
${\cal D}(Switch\mbox{-}Power')$ includes the following initial propositions:
\begin{quote}
\noindent
{\bf initially} $\neg On$,\\
{\bf initially} $\neg Switch$,
\end{quote}
and the action effect proposition
\begin{quote}
$Turn\mbox{-}On$ {\bf causes} $Switch$,
\end{quote}
together with the effect proposition of action $Cut\mbox{-}Power$ and
two causal propositions as given in Example 2.
Ignoring the detail, we can derive the following results:
\begin{quote}
\noindent
${\cal D}(Switch\mbox{-}Power')\models_{{\cal AT}^{0}} On$ {\bf after} $Turn\mbox{-}On$,\\
${\cal D}(Switch\mbox{-}Power')\models_{{\cal AT}^{0}} \neg On$
{\bf after} $Turn\mbox{-}On, Cut\mbox{-}Power$,\\
${\cal D}(Switch\mbox{-}Power')\models_{{\cal AT}^{0}} Switch$
{\bf after} $Turn\mbox{-}On, Cut\mbox{-}Power$.
\end{quote}
\end{example}

\section{${\cal AT}^{1}$: Combining Defeasible Observations into Action Domains}

We have shown that language  ${\cal AT}^{0}$ 
handles temporal prediction  
where defeasible constraints are admitted.
It, however, cannot deal with temporal postdiction, e.g. 
within the framework of ${\cal AT}^{0}$ we cannot reason
from the current state to the past under some observations. 
It has been realized that observations on any intermidate
states (including the final state)
play an important role in temporal postdiction \cite{jlm:action96}. Here,
an observation is viewed as an agent's beliefs about the domain that 
is either obtained from the outside world or from the agent's own assumption.
In the case that
an agent makes an observation under some
assumption, such observation becomes defeasible because
once the assumption is proved not to be true, the agent's observation
should be defeated.

In this section, 
we extend ${\cal AT}^{0}$ to ${\cal AT}^{1}$ such that the extended language
can handle temporal prediction and postdiction where 
both defeasible constraints and observations are admitted. 

\subsection{Syntax of ${\cal AT}^{1}$}

The syntax of ${\cal AT}^{1}$ is the same as  
${\cal AT}^{0}$'s except that  
${\cal AT}^{1}$ also has an {\em 
observation proposition} of the form:
\begin{eqnarray}
L \hspace*{.05in} \mbox{\bf is observed if} \hspace*{.05in} 
L_{1},\cdots,L_{m} \hspace*{.05in} \mbox{\bf with absence}
\hspace*{.05in} L_{m+1},\cdots, L_{n} \hspace*{.05in} \mbox{\bf after}
\hspace*{.05in} A_{1},\cdots, A_{l},
\label{s2.1}
\end{eqnarray}
where $L, L_{1},\cdots, L_{n}$ are fluent expressions, and
$A_{1},\cdots,A_{l}$ are actions.
Intuitively, (\ref{s2.1}) says after actions $A_{1},\cdots, A_{l}$ are 
performed sequentially, $L$ is observed to be true if $L_{1},\cdots,L_{m}$
are true while $L_{m+1},\cdots,L_{n}$ are absent. Obviously, (\ref{s2.1})
represents a kind of defeasible information.
In the case that no action occurs in (\ref{s2.1}), (\ref{s2.1}) can be written
as the following form:
\begin{eqnarray}
\mbox{\bf initially} \hspace*{.05in}
L \hspace*{.05in} \mbox{\bf is observed if} \hspace*{.05in}
L_{1},\cdots,L_{m} \hspace*{.05in} \mbox{\bf with absence}
\hspace*{.05in} L_{m+1},\cdots, L_{n}.
\label{s2.2}
\end{eqnarray}

Under the language ${\cal AT}^{1}$, we define a {\em domain description} ${\cal D}$
to be a finite set of observation propositions,
causal propositions and action effect propositions. 
${\cal AT}^{1}$ will still have the value proposition (\ref{s1.0}) and its special
case the initial proposition (\ref{s1.1}), 
but are only used as query statements in ${\cal AT}^{1}$.

\begin{example}
Let us consider a modified shooting action scenario
which we name $Shooting$-$1$.
Suppose the turkey is observed alive in the initial situation, and as
there is no explicit information about whether the gun is loaded in the
initial situation, the agent would assume that the gun is
initially not loaded by default. After actions {\em Shoot} and {\em Wait} are
successively performed, it is observed that the turkey is dead (not alive).
This scenario can be naturally described by language 
${\cal AT}^{1}$. In particular, we specify a domain description
${\cal D}(Shooting\mbox{-}1)$ which has the following observation propositions:
\begin{quote}
\noindent
{\bf initially} $Alive$ {\bf is observed},\\
{\bf initially} $\neg Loaded$ {\bf is observed with absence} $Loaded$, \\
$\neg Alive$ {\bf is observed after} $Shoot, Wait$,
\end{quote}
and an action effect proposition:
\begin{quote}
$Shoot$ {\bf causes} $\neg Alive$ {\bf if} $Loaded$.
\end{quote}
\end{example}

\subsection{Semantics of ${\cal AT}^{1}$}

We will use a  similar way as described in section 3.2 to develop
a formal semantics of ${\cal AT}^{1}$ based on a transition system that
is defined on the basis of the translation from a ${\cal AT}^{1}$
domain description into a PLP.

\subsubsection{Translating ${\cal AT}^{1}$ into PLP}

As we have mentioned earlier, the major improvement from ${\cal AT}^{0}$ to
${\cal AT}^{1}$ is that we allow defeasible observations to be presented in
a domain description so that temporal postdiction becomes 
possible. It is quite straightforward to translate an 
observation proposition of the form (\ref{s2.1}) into the following logic rule:
\begin{eqnarray}
[\neg] Holds(F,S) & \leftarrow &
[\neg ] Holds(F_{1},S),\cdots, [\neg ] Holds(F_{m},S), \nonumber \\
 & & not \hspace*{.05in} [\neg ] Holds(F_{m+1},S),\cdots,
not \hspace*{.05in} [\neg ] Holds(F_{n},S),
\label{n4.1}
\end{eqnarray}
where $S=Result(A_{l},Result(\cdots, Result(A_{1},S_{0})\cdots))$.

To do postdiction reasoning, for each action effect proposition in ${\cal D}$, we
need to have some action explanation rules which will be used to derive 
action preconditions based on proper observations. First, if there is an 
action effect rule (\ref{3.5}), the following rule
explains that the fact $Holds(F,Result(A,s))$ is caused by performing
action $A$:
\begin{eqnarray}
Effect^{+}(F,Result(A,s)) & \leftarrow & Holds(F,Result(A,s)),
not \hspace*{.05in} Holds(F,s), \nonumber \\
& & not \hspace*{.05in} Caused^{+}(F,Result(A,s)),
\label{4.1}
\end{eqnarray}
Clearly, the function of rule (\ref{4.1}) is to identify action $A$'s actual execution. 
The intuition is that if fluent $F$ is true (or false, resp.)
in situation $Result(A,s)$, and
there is no explicit information saying that
$F$ is true in the previous situation $s$ or
$F$ is caused to be true by some causal
rule, then it derives that $F$'s truth value in
$Result(A,s)$ is a direct effect of action $A$.

Furthermore, if a fluent $F$ is a direct effect
of some action $A$, i.e. $Effect^{+}(F,Result(A,s))$ holds, then
each precondition of $A$ must
also hold in the previous situation. That is, we should have 
rules like:
\begin{eqnarray}
[\neg ]Holds(F_{i},s) & \leftarrow & Effect^{+}(F,Result(A,s)), 
\label{4.2}
\end{eqnarray}
where $i=1,\cdots,k$ and 
$A$ {\bf causes} $F$ {\bf if} $[\neg ] F_{1},\cdots, [\neg] F_{k}$ is an 
action effect proposition in domain ${\cal D}$.
However, it should be noted that sometimes one action may cause the same
effect under different preconditions. In this case,
deriving all possible action preconditions may cause contradictions.
For instance, consider the following domain description ${\cal D}(Door)$: 
\begin{quote}
\noindent
{\bf initially} $\neg$ {\em HasKey},\\ 
$DoorOpened$ {\bf is observed after} {\em OpenDoor},\\
{\em OpenDoor} {\bf causes} {\em DoorOpened} {\bf if} {\em HasCard},\\
{\em OpenDoor} {\bf causes} {\em DoorOpened} {\bf if} {\em HasKey}.
\end{quote}
In this domain, action {\em OpenDoor} has two independent
preconditions {\em HasCard} and {\em HasKey}.
If we translate this domain according to our proposal above, we will have the following
logic rules:
\begin{quote}
\noindent
$Holds(DoorOpened,Result(OpenDoor,S_{0}))\leftarrow$,
\end{quote} 
\begin{quote}
$Effect^{+}(DoorOpened,Result(OpenDoor,s))  \leftarrow$  \\
\hspace*{.8in} $Holds(DoorOpened,Result(OpenDoor,s))$,  \\
\hspace*{.8in}  $not \hspace*{.05in} Holds(DoorOpened,s)$,  \\
\hspace*{.8in}  $not \hspace*{.05in} Caused^{+}(DoorOpened,Result(OpenDoor,s))$,
\end{quote}
\begin{quote}
$Holds(HasCard,s)\leftarrow Effect^{+}(DoorOpened,Result(OpenDoor,s))$, and\\
$Holds(HasKey,s)\leftarrow Effect^{+}(DoorOpened,Result(OpenDoor,s))$.
\end{quote}

From the above logic rules, we will deduce both $Holds(HasCard,S_{0})$ and \\
$Holds(HasKey,S_{0})$.
But from ${\cal D}(Door)$, we know that
$\neg HasKey$ initially holds.
To avoid this kind of contradiction, instead of 
using rule (\ref{4.2}), 
we should have a weaker rule to 
derive action preconditions: {\em whenever there is no conflict, we only
deduce a minimal number of preconditions to explain an action}.
Under this principle, we will change rule (\ref{4.2}) to the following form:
\begin{eqnarray}
[\neg ]Holds(F_{i},s) & \leftarrow & Effect^{+}(F,Result(A,s)), \nonumber \\
& & not \overline{[\neg ]Holds(F_{i},s)}, \nonumber \\
& & not [\neg ]Holds(F_{1}',s), \cdots, not [\neg ]Holds(F_{l}',s),
\label{n4.2}
\end{eqnarray}
where $i=1,\cdots,k$, and
fluents $[\neg ] F_{1}',\cdots [\neg ]F_{l}'$ occur as preconditions in all other action 
effect propositions of $A$ that have the same effect\footnote{$\overline{[\neg ]Holds(F_{i},s)}$
denotes the complementary literal of $[\neg ]Holds(F_{i},s)$.}.

The following rules represent the dual case of rules (\ref{4.1}) and
(\ref{n4.2}) corresponding to action effect rule (\ref{3.6}):
\begin{eqnarray}
Effect^{-}(F,Result(A,s)) & \leftarrow & Holds(F,Result(A,s)),
not \hspace*{.05in} Holds(F,s), \nonumber \\
& & not \hspace*{.05in} Caused^{+}(F,Result(A,s)),
\label{n4.3}   
\end{eqnarray} 
\begin{eqnarray}
[\neg ]Holds(F_{i},s) & \leftarrow & Effect^{-}(F,Result(A,s)), \nonumber \\
& & not \overline{[\neg ]Holds(F_{i},s)}, \nonumber \\
& & not [\neg ]Holds(F_{1}',s), \cdots, not [\neg ]Holds(F_{l}',s).
\label{n4.4}
\end{eqnarray}

Now the following definition describes the formal translation from a domain description
${\cal D}$ of ${\cal AT}^{1}$ into a PLP .

\begin{definition}
A PLP is called a {\em translation} of domain description
${\cal D}$ of ${\cal AT}^{1}$, denoted by
${\cal P}^{{\cal AT}^{1}}({\cal D}) = (\Pi, {\cal N},<)$, if it is obtained 
as follows:
\begin{enumerate}
\item $\Pi$ consists of the following rules:\\
\underline{\em Observation rules}: for each observation proposition of (\ref{s2.1}), 
there is a rule of the form (\ref{n4.1}),\\
\underline{\em Causal rules}: the same as in Definition 4,\\
\underline{\em Action effect rules}: the same as in Definition 4,\\
\underline{\em Action explanation rules}: 
for each action effect rule (\ref{3.5}), there are 
rules (\ref{4.1}) and (\ref{n4.2}),
and for each action effect rule (\ref{3.6}), there are rules (\ref{n4.3})
and (\ref{n4.4}),
\underline{\em Inertia rules}: (\ref{3.9}), (\ref{3.10}) and:
\begin{eqnarray}
Holds(f,s) & \leftarrow & 
Holds(f,Result(a,s)), not \hspace*{.05in} \neg Holds(f,s), \nonumber \\
& &  not \hspace*{.05in} Caused^{+}(f,Result(a,s)), \nonumber \\
& & not \hspace*{.05in} Effect^{+}(f,Result(a,s)),  
\label{4.5}
\end{eqnarray}
\begin{eqnarray}
\neg Holds(f,s)& \leftarrow &
\neg Holds(f,Result(a,s)), not \hspace*{.05in} Holds(f,s), \nonumber \\
& &  not \hspace*{.05in} Caused^{-}(f,Result(a,s)), \nonumber \\
& & not \hspace*{.05in} Effect^{-}(f,Result(a,s)). 
\label{4.6}
\end{eqnarray}
\item Naming function ${\cal N}$ assigns a unique name to each rule in $\Pi$;

\item For each observation rule $N_{o}$, causal rule 
$N_{c}$, action explanation rule $N_{ex}$ and inertia rule
$N_{i}$, the following $<$-relations hold: 
\begin{eqnarray}
N_{ex}<N_{c}<N_{i}<N_{o}. 
\label{4.7}
\end{eqnarray}
\end{enumerate}
\end{definition}

Compared with  
Definition 4, the PLP translation specified in Definition 8
presents several
new features. First, ${\cal P}^{{\cal AT}^{1}}({\cal D})$
allows to represent 
defeasible observations not only at the initial situation but 
also at any other situations. 
Second, ${\cal P}^{{\cal AT}^{1}}({\cal D})$
includes action explanation rules (\ref{4.1}), 
(\ref{n4.2}), (\ref{n4.3}), and (\ref{n4.4}).
Finally, 
the extra inertia rules (\ref{4.5}) and (\ref{4.6}) allow us to reason 
about fluents' truth values from the current situation to the past.
That is, if a fluent $f$ is true (or false, resp.)
currently, and there is no explicit
information saying that 
$f$ is not true (or not false, resp.) in the
previous situation, or $f$ is caused to be true by some causal rule, 
or $f$ is true (or false, resp.) as a direct effect of some action, then
it derives that $f$ is true (or false, resp.) in the 
previous situation.

Since both observation and action explanation rules may be defeasible, 
more possible conflicts may occur in ${\cal P}^{{\cal AT}^{1}}({\cal D})$.
For instance, conflicts may not only
occur between causal rules and inertia rules, but also between
action explanation rules and inertia rules, observation rules and causal 
rules, etc..
To solve these possible conflicts, the underlying
$<$-relation is specified as (\ref{4.7}).
(\ref{4.7}) presents that action explanation rules are most preferred because
the execution of an action usually override
defeasible causal and inertia rules, while observation rules are 
less preferred than inertia rules due to the intuition that a fluent's truth value 
normally persists if there is no explicit 
action or causal rule to change it.

Note that in ${\cal P}^{{\cal AT}^{1}}({\cal D})$, 
action explanation rules (\ref{4.1}),  (\ref{n4.2}),
(\ref{n4.3}) and (\ref{n4.4})
are domain specific because they are specified based on 
action effect rules (\ref{3.5}) and (\ref{3.6}).
On the other hand, the new inertia rules (\ref{4.5}) and
(\ref{4.6}) are domain independent. Therefore, we can denote domain 
independent rules in ${\cal P}^{{\cal AT}^{1}}({\cal D})$ as follows:
\begin{eqnarray}
\Pi_{ind}^{1}=\Pi_{ind}^{c}\cup\Pi_{ind}^{eff}\cup\Pi_{ind}^{i'}, 
\label{4.ex}
\end{eqnarray}
where $\Pi_{ind}^{i'}=\Pi_{ind}^{i}\cup\{(\ref{4.5}),(\ref{4.6})\}$.
We also denote the set of domain specific rules
of ${\cal P}^{{\cal AT}^{1}}({\cal D})$ as $\Pi_{spec}^{1}$.

\begin{example} 
Example 5 continued. According to Definition 8, it is not
difficult to obtain
the translation of domain description ${\cal D}(Shooting\mbox{-}1)$, 
${\cal P}^{{\cal AT}^{1}}(Shooting$-$1)=(\Pi_{spec}^{1}\cup\Pi_{ind}^{1},{\cal N},<)$, 
where $\Pi_{spec}^{1}$ consists of the
following rules:\\
\underline{Observation rules}:\\
$N_{1}: Holds(Alive,S_{0})\leftarrow$,\\
$N_{2}: \neg Holds(Loaded,S_{0})\leftarrow$ {\em not} $Holds(Loaded,S_{0})$,\\
$N_{3}: \neg Holds(Alive,Result(Wait,Result(Shoot,S_{0})))\leftarrow$, \\
\underline{Action effect rules}:\\
$N_{4}:Effect^{-}(Alive,Result(Shoot,s))\leftarrow$
$Holds(Loaded,s)$,\\
%
\underline{Action explanation rules}:
\begin{eqnarray*}
N_{5}: Effect^{-}(Alive,Result(Shoot,s)) & \leftarrow &
\neg Holds(Alive,Result(Shoot,s)), \nonumber \\
& & not \hspace*{.05in} \neg Holds(Alive,s), \nonumber \\
& & not \hspace*{.05in} Caused^{-}(Alive,Result(Shoot,s)),
\end{eqnarray*}
$N_{6}: Holds(Loaded,s)\leftarrow$
$Effect^{-}(Alive,Result(Shoot,s)), not \neg Holds(Loaded,s)$.\\
\underline{Naming rules in $\Pi_{ind}^{1}$}:\\
Assigning a unique name to each rule in $\Pi_{ind}^{1}$.
Therefore, we have names $N_{7}, N_{8}, N_{9}, N_{10}$, $N_{11}$,
$N_{12}$, $N_{13}$, and $N_{14}$ for 
rules (\ref{3.3}), (\ref{3.4}),
(\ref{3.7}), (\ref{3.8}), (\ref{3.9}), (\ref{3.10}), (\ref{4.5}) and
(\ref{4.6}) respectively.\\
\underline{$<$-relations}:\\
$N_{ex}<N_{i}<N_{o}$. That is, we have:\\   
$\{N_{5},N_{6}\}<\{N_{11},N_{12},N_{13},N_{14}\}<\{N_{1},N_{2},N_{3}\}$.
\end{example}

\subsubsection{Transition function,  models and entailment}

Transition function ${\cal R}$,
structures and models $\Psi$
are defined in the same way as in ${\cal AT}^{0}$ (see section 3.2.2).
We denote the entailment relation under $\Psi$ in ${\cal AT}^{1}$
as $\models_{{\cal AT}^{1}}$.
The only thing we should emphasize is that
since we allow a domain description to include defeasible initial
observation propositions, it is possible that one initial observation
proposition conflicts with the other. Therefore, different initial states
$\hat{S}_{0}$ may be deduced from different answer sets of the
corresponding translated PLP ${\cal P}^{{\cal AT}^{1}}({\cal D})$
of ${\cal D}$. 

\begin{example} 
Example 5 continued.
In the shooting action scenario as described in Example 5, the question
we are interested in is when the turkey died and 
whether the gun was actually loaded initially. This is a 
question about postdiction that we need to reason from the current 
situation to the past. After translating the domain description
${\cal D}(Shooting\mbox{-}1)$ into ${\cal P}^{{\cal AT}^{1}}(Shooting\mbox{-}1)$
as illustrated in Example 6, we obtain the following results:
\begin{quote}
\noindent
${\cal D}(Shooting\mbox{-}1)\models_{{\cal AT}^{1}} \neg Alive$ {\bf after} $Shoot$,\\
${\cal D}(Shooting\mbox{-}1)\models_{{\cal AT}^{1}}$ {\bf initially} $Loaded$,
\end{quote}
where the first solution says that the turkey was dead after
the execution of action $Shoot$, and the second indicates that initially the
gun was actually loaded, which defeats the original observation.
\end{example}

\section{${\cal AT}^{2}$: Representing Actions 
          with Defeasible and Abnormal Effects}

It is common that in temporal reasoning under some circumstances,
an action might be abnormally executed and the 
original expected action effect is defeated. Sometimes, an abnormal
effect associated with this action may be also produced.
Consider the classic shooting scenario, e.g. \cite{book:shoot},
in which it is usually assumed that if the gun is loaded, then the {\em shoot} action
causes a direct effect that the turkey is not alive. However, it is probably 
more natural to treat {\em shoot} as a defeasible action. For instance,
if the bullet is dumb, the turkey would be still alive
after executing action {\em shoot}, or
it could be an abnormal effect of {\em shoot} if after shooting the turkey is
still alive but the pigeon is dead.
In this section, we try to further generalize our action
language ${\cal AT}^{1}$ to ${\cal AT}^{2}$ in order to capture 
actions with defeasible and/or abnormal effects as described above.

\subsection{Syntax of ${\cal AT}^{2}$}

${\cal AT}^{2}$ includes the same forms of 
observation propositions, causal propositions, and
value propositions of ${\cal AT}^{1}$, but has 
different forms of action effect propositions.
First, an {\em action effect proposition} of ${\cal AT}^{2}$ is of the
following form:
\begin{eqnarray}
A \hspace*{.05in} \mbox{\bf normally causes} \hspace*{.05in}
L \hspace*{.05in} \mbox{\bf if} \hspace*{.05in} L_{1},\cdots,L_{k},
\label{s3.1}
\end{eqnarray}
where $A$ is an action and $L, L_{1},\cdots, L_{k}$ are
fluent expressions. Intuitively,
this action effect proposition is defeasible since
we consider that if an action is {\em abnormally} executed, its normal 
effect then cannot be produced.

Therefore, the following 
{\em action abnormal effect proposition} represents the abnormal effect of
an action:
\begin{eqnarray}
A \hspace*{.05in} \mbox{\bf abnormally causes} \hspace*{.05in}
L \hspace*{.05in} \mbox{\bf if} \hspace*{.05in} L_{1},\cdots,L_{k}.
\label{s3.2}
\end{eqnarray}

Finally, an {\em abnormal condition proposition} represents the condition under which 
an action can be considered to be abnormal:
\begin{eqnarray}
A \hspace*{.05in} \mbox{\bf is abnormal if before} 
\hspace*{.05in} L_{1},\cdots, L_{h} \hspace*{.05in} \mbox{\bf after}
\hspace*{.05in} L_{h+1},\cdots, L_{p}.
\label{s3.3}
\end{eqnarray}
Usually, the abnormality of an action can be 
identified from observations on the changes of
some particular fluents' truth values before and after the 
action execution. Hence, (\ref{s3.3}) says that if $L_{1},\cdots, L_{h}$ are true
{\em before} action $A$ is executed, and $L_{h+1},\cdots, L_{p}$ are true
{\em after} action $A$ is executed, then $A$ is identified to be abnormal.

A {\em domain description} ${\cal D}$ of ${\cal AT}^{2}$ is a finite set of
observation propositions, causal propositions, action effect propositions, 
abnormal action effect propositions, and abnormal condition propositions.
The following example shows how we can use ${\cal AT}^{2}$ to represent
domains where actions may have abnormal or/and defeasible effects.

\begin{example}
Let us consider a different shooting scenario named 
$Shooting\mbox{-}2$ in which action $Shoot$ has a defeasible effect and
it is abnormally executed if initially the gun is
loaded and after performing the action, the turkey is observed still alive. 
Initially the gun is loaded and turkey is alive.
This scenario is easy to formalize by using ${\cal AT}^{2}$. 
We specify a domain description ${\cal D}(Shooting\mbox{-}2)$
that has the following observation propositions:
\begin{quote}
\noindent
{\bf initially} $Loaded$ {\bf is observed},\\
{\bf initially} $Alive$ {\bf is observed},
\end{quote}
a defeasible action effect proposition:
\begin{quote}
$Shoot$ {\bf normallly causes} $\neg Alive$ {\bf if} $Loaded$,
\end{quote}
and an abnormal condition proposition:
\begin{quote}
$Shoot$ {\bf is abnormal if before} $Loaded$ {\bf after} $Alive$.
\end{quote}
\end{example} 

\subsection{Semantics of ${\cal AT}^{2}$}

Similarly to previous languages ${\cal AT}^{0}$ and ${\cal AT}^{1}$,
we will propose a transition system to provide a formal semantics
of ${\cal AT}^{2}$. Again, this transition system is defined based on
a translation from a domain description of ${\cal AT}^{2}$ into a PLP. 

\subsubsection{Translating  ${\cal AT}^{2}$ into PLP}

To translate an action domain of ${\cal AT}^{2}$, we need to
extend the language ${\cal L}^{P}_{{\cal AT}^{0}}$ of PLPs introduced in section 3.2
to a new language ${\cal L}^{P}_{{\cal AT}^{2}}$ of PLPs by adding following 
symbols:
\begin{enumerate}
\item[-] $Ab$: a binary predicate symbol taking 
arguments {\em action} and {\em situation} respectively.
\item[-] $AbEffect^{+}$ and $AbEffect^{-}$: binary predicate symbols 
taking arguments {\em fluent} and {\em situation} respectively.
\end{enumerate}
Intuitively, atom $Ab(a,s)$ expresses that action $a$ is abnormally 
executed at situation $s$, while atoms $AbEffect^{+}(f,s)$ 
and $AbEffect^{-}(f,s)$ are used to represent
abnormal effects of actions (see the following for detail).

Considering the defeasibility of action executions, 
we need to modify our original action effect rules (\ref{3.5}) and 
(\ref{3.6}) presented in section 3.2 to the following forms respectively:
\begin{eqnarray}
Effect^{+}(F,Result(A,s)) & \leftarrow &
[\neg ]Holds(F_{1},s),\cdots,[\neg ]Holds(F_{k},s), \nonumber \\
& & not Ab(A,s),
\label{5.1}
\end{eqnarray}
\begin{eqnarray}
Effect^{-}(F,Result(A,s)) & \leftarrow &
[\neg ]Holds(F_{1},s),\cdots,[\neg ]Holds(F_{k},s), \nonumber \\
& & not Ab(A,s),
\label{5.2}
\end{eqnarray}

Rule (\ref{5.1}) (or (\ref{5.2}) resp.) says that if $A$'s
preconditions $[\neg ]Holds(F_{1},s), \cdots, [\neg Holds(F_{k},s)$ hold, and
there is no explicit information stating that $A$ is abnormally executed at 
situation $s$, then fluent $F$ will be true (or false, resp.) 
in situation $Result(A,s)$ as a direct effect of $A$.
Additionally we also need a generic schema for any action $a$:
\begin{eqnarray}
\neg Ab(a,s) & \leftarrow & not \hspace*{.05in} Ab(a,s),
\label{5.3}
\end{eqnarray}
which simply expresses that if there is no explicit information 
saying that action $A$ is abnormally executed at situation $s$, then it
is assume that $A$ is {\em not} abnormally executed at situation $s$.
To simplify our following presentation, we denote

\begin{eqnarray*}
\Pi_{ind}^{eff'}=\Pi_{ind}^{eff}\cup \{(\ref{5.3})\}\footnote{Note that
$\Pi_{ind}^{eff}=\{(\ref{3.7}), (\ref{3.8})\}$ 
(see section 3.2).}.
\end{eqnarray*}

Consequently, action explanation rules (\ref{4.1}) and (\ref{n4.3})
in ${\cal AT}^{1}$ are also modified as follows respectively:
\begin{eqnarray}
Effect^{+}(F,Result(A,s)) & \leftarrow & Holds(F,Result(A,s)), 
not \hspace*{.05in} Holds(F,s), \nonumber \\
& & not \hspace*{.05in} Caused^{+}(F,Result(A,s)), \nonumber \\
& & not \hspace*{.05in} Ab(A,s),
\label{5.4}
\end{eqnarray}
\begin{eqnarray}
Effect^{-}(F,Result(A,s)) & \leftarrow & \neg Holds(F,Result(A,s)), 
not \hspace*{.05in} \neg Holds(F,s), \nonumber \\
& & not \hspace*{.05in} Caused^{-}(F,Result(A,s)), \nonumber \\
& & not \hspace*{.05in} Ab(A,s).
\label{5.5}
\end{eqnarray}
(\ref{5.4}) states that if fluent $F$ is true in situation $Result(A,s)$, and
there is no evidence to show that (a) 
$F$ is true in the previous situation $s$; (b) $F$ is caused to be
true in situation $Result(A,s)$; and (c) action $A$ is abnormal at
situation $s$, then it is derived that $F$ must be a direct effect of 
action $A$ in situation $Result(A,s)$. (\ref{5.5}) has a similar interpretation.

As we mentioned earlier, some actions with defeasible effects may also
produce abnormal effects. 
Hence, we also specify action abnormal effect rules of the 
following forms:
\begin{eqnarray}
AbEffect^{+}(F,Result(A,s)) & \leftarrow &
[\neg ]Holds(F_{1},s),\cdots,[\neg ]Holds(F_{l},s), \nonumber \\
& & Ab(A,s),
\label{5.6}
\end{eqnarray}
\begin{eqnarray}
AbEffect^{-}(F,Result(A,s)) & \leftarrow &
[\neg ]Holds(F_{1},s),\cdots,[\neg ]Holds(F_{l},s), \nonumber \\
& & Ab(A,s),
\label{5.7}
\end{eqnarray}
\begin{eqnarray}
Holds(f,s) & \leftarrow & AbEffect^{+}(f,s),
\label{5.8}
\end{eqnarray}
\begin{eqnarray}
\neg Holds(f,s) & \leftarrow & AbEffect^{-}(f,s),
\label{5.9}
\end{eqnarray}
\begin{eqnarray}
Ab(A,s) & \leftarrow & [\neg ]Holds(F_{1},s),
\cdots,[\neg ]Holds(F_{h},s), \nonumber \\
& & [\neg ]Holds(F_{h+1},Result(A,s)), \cdots, [\neg ]Holds(F_{p},Result(A,s)),
\label{5.10}
\end{eqnarray}

Basicly, rule (\ref{5.6}) (or (\ref{5.7}) resp.) says that 
if conditions $[\neg ]Holds(F_{1},s),\cdots$,\\
$[\neg ]Holds(F_{l},s)$ hold and
$A$ is abnormally executed, then fluent $F$ will be true
(or false resp.) as an abnormal effect of $A$ in situation
$Result(A,s)$. 
Rule (\ref{5.10}), on the other hand,  is a direct translation of 
abnormal condition proposition (\ref{s3.3}).
Clearly, rules (\ref{5.8}) and (\ref{5.9}) are domain independent while
rules (\ref{5.6}), (\ref{5.7}) and (\ref{5.10}) are domain specific. 
Again for simplicity, we denote 
\begin{eqnarray*}
\Pi_{ind}^{ab}=\{(\ref{5.8}), (\ref{5.9})\}.
\end{eqnarray*}

Now we are able to describe our translation from a
domain description of ${\cal AT}^{2}$ into a PLP
as follows.

\begin{definition}
A PLP is called a {\em translation} of domain description 
${\cal D}$ of ${\cal AT}^{2}$, denoted by
${\cal P}^{{\cal AT}^{2}}({\cal D})=(\Pi,{\cal N},<)$, if it obtained as follows:
\begin{enumerate}
\item $\Pi$ consists of following rules:\\
\underline{\em Observation rules}: the same as in 
Definition 8. \\
\underline{\em Causal rules}: the same as in Definition 8.\\
\underline{\em Action effect rules}: for each action effect proposition
(\ref{s3.1}), there is a rule of the form (\ref{5.1}) or
(\ref{5.2}). Three domain independent action effect rules
(\ref{3.7}), (\ref{3.8}) and (\ref{5.3}) are also included in this set.\\
\underline{\em Action explanation rules}: 
for each action effect rule of the form (\ref{5.1}), there are 
rules (\ref{5.4}) and (\ref{n4.2}), and 
for each action effect rule of the form (\ref{5.2}), there are rules
(\ref{5.5}) and (\ref{n4.4}),\\
\underline{\em Action abnormal effect rules}: for each action abnormal effect 
proposition (\ref{s3.2}), there are rules 
(\ref{5.6}) - (\ref{5.9}), and for each abnormal condition proposition
(\ref{s3.3}), there is a rule (\ref{5.10}).\\
\underline{\em Inertia rules}: (\ref{3.9}), (\ref{3.10}) and:
\begin{eqnarray}
Holds(f,s) & \leftarrow & Holds(f,Result(a,s)), 
not \hspace*{.05in} \neg Holds(f,s), \nonumber \\
& & not \hspace*{.05in} Caused^{+}(f,Result(a,s)), \nonumber \\
& & not \hspace*{.05in} Effect^{+}(f,Result(a,s)), \nonumber \\
& & not \hspace*{.05in} AbEffect^{+}(f,Result(a,s)),
\label{5.11}
\end{eqnarray}
\begin{eqnarray}
\neg Holds(f,s) & \leftarrow & \neg Holds(f,Result(a,s)), 
not \hspace*{.05in} Holds(f,s), \nonumber \\
& & not \hspace*{.05in} Caused^{-}(f,Result(a,s)), \nonumber \\
& & not \hspace*{.05in} Effect^{-}(f,Result(a,s)), \nonumber \\
& & not \hspace*{.05in} AbEffect^{-}(f,Result(a,s));
\label{5.12}
\end{eqnarray}

\item Naming function ${\cal N}$ assigns a unique name to each rule
in $\Pi$;

\item For each observation rule $N_{o}$, causal rule 
$N_{c}$, action effect rule $N_{eff}$, action explanation rule
$N_{ex}$, and
inertia rule $N_{i}$, there are $<$-relations (\ref{4.7}):
\begin{eqnarray*}
N_{ex}<N_{c}<N_{i}<N_{o}
\end{eqnarray*}
and
\begin{eqnarray}
N_{eff}<N_{c}<N_{i}<N_{o}.
\label{5.13}
\end{eqnarray}
\end{enumerate}
\end{definition}

Note that inertia rules (\ref{5.11}) and (\ref{5.12}) are a natural extension of
inertia rules (\ref{4.5}) and (\ref{4.6}) in ${\cal P}^{{\cal AT}^{1}}({\cal D})$
respectively.
The $<$-relations in ${\cal AT}^{2}$ are  specified in a similar
way as in ${\cal P}^{{\cal AT}^{1}}({\cal D})$ 
except one more schema (\ref{5.13})
is added. This is because in 
${\cal P}^{{\cal AT}^{2}}({\cal D})$ action effect rules
(\ref{5.1}) and (\ref{5.2}) are defeasible, possible conflicts between these rules 
and other defeasible rules (e.g. causal rules, inertia rules and
observation rules) may occur indirectly
through the action abnormal effect rule
(\ref{5.10}). Therefore, $<$-relations (\ref{4.7}) 
(see section 4.2) and (\ref{5.13}) are needed as
we always assume that an action's
successful execution should have the highest priority.

We denote domain independent
rules in ${\cal P}^{{\cal AT}^{2}}({\cal D})$ as follows:
\begin{eqnarray}
\Pi_{ind}^{2}=\Pi_{ind}^{c}\cup\Pi_{ind}^{eff'}\cup\Pi_{ind}^{ab}\cup\Pi_{ind}^{i''},
\label{5.14}
\end{eqnarray}
where $\Pi_{ind}^{i''}=\Pi_{ind}^{i}\cup\{(\ref{5.11}), (\ref{5.12})\}$, and
denote the set of domain specific rules by $\Pi_{spec}^{2}$.

\subsubsection{Transition function, models and entailment}

Transition function ${\cal R}$,
structures and models $\Psi$
of ${\cal AT}^{2}$ are defined exactly the same as 
in section 3.2.2. The entailment relation under $\Psi$ in ${\cal AT}^{2}$ 
is denoted as $\models_{{\cal AT}^{2}}$.
Again, it is observed that 
the initial state of a domain description ${\cal D}$ of ${\cal AT}^{2}$ may not
be unique due to a possible conflict occurring between two
defeasible initial observation propositions in ${\cal D}$.

\begin{example} 
Example 8 continued.
Given the domain description ${\cal D}(Shooting\mbox{-}2)$ as presented in 
Example 8, the translated PLP ${\cal P}^{{\cal AT}^{2}}(Shooting\mbox{-}2)$
is easy to be obtained according to Definition 9.
Let ${\cal P}^{{\cal AT}^{2}}(Shooting\mbox{-}2)$
$=(\Pi_{spec}^{2}\cup\Pi_{ind}^{2},{\cal N}, <)$, where
$\Pi_{spec}^{2}$ consists of the following rules:\\
\underline{Observation rules}:\\
$N_{1}: Holds(Loaded,S_{0})\leftarrow$,\\
$N_{2}: Holds(Alive,S_{0})\leftarrow$,\\
\underline{Action effect rule}:
\begin{eqnarray*}
N_{3}: Effect^{-}(Alive,Result(Shoot,s)) & \leftarrow & 
Holds(Loaded,s), not \hspace*{.05in} Ab(Shoot,s),
\end{eqnarray*}
\underline{Action explanation rule}:
\begin{eqnarray*}
N_{4}: Effect^{-}(Alive,Result(Shoot,s)) & \leftarrow & 
\neg Holds(Alive,Result(Shoot,s)), \\
& & not \hspace*{.05in} \neg Holds(Alive,s), \\
& & not \hspace*{.05in} Caused^{-}(Alive,Result(Shoot,s)),\\
& & not \hspace*{.05in} Ab(Shoot,s),
\end{eqnarray*}
\underline{Action abnormal effect rule}:
\begin{eqnarray*}
N_{5}: Ab(Shoot,s) & \leftarrow & Holds(Loaded,s), Holds(Alive,Result(Shoot,s)),
\end{eqnarray*}
\underline{Naming rules in $\Pi_{ind}^{2}$}:
Assigning a unique name to each rule in $\Pi_{ind}^{2}$.\\
\underline{$<$-relations}: (\ref{4.7}) and (\ref{5.13}).\\

Since the action effect rule $N_{3}$
is defeasible,
it is not difficult to see that a conflict on the truth value of
$Holds(Alive,Result(Shoot,S_{0}))$ occurs
between rule $N_{3}$ and an inertia rule  
\begin{quote}
$N':$
$Holds(Alive,Result(Shoot,s))\leftarrow Holds(Alive,s)$, \\
\hspace*{2.6in} $not \hspace*{.05in} \neg Holds(Alive,Result(Shoot,s))$
\end{quote}
which is an instance of the generic inertia
rule (\ref{3.9}) included in $\Pi_{ind}^{2}$\footnote{Note that the conflict
is introduced through rule $N_{5}$.}.
However, this conflict is solved by 
$N_{3}<N'$. Therefore, we have the final result
\begin{quote}
${\cal D}(Shooting$-$2)\models_{{\cal AT}^{2}} \neg Alive$ {\bf after} $Shoot$, 
\end{quote}
from which it is concluded
that action $Shoot$ is not abnormally executed.
\end{example}

\section{Characterizations of Action Domains}

Among all action domains specified by languages ${\cal AT}^{0}$, 
${\cal AT}^{1}$ and ${\cal AT}^{2}$, there are
some classes of action domains that may have more 
desirable properties than other classes of domains.
In this section, we investigate these desirable 
properties and characterize different
action domains within languages ${\cal AT}^{0}$,
${\cal AT}^{1}$ and ${\cal AT}^{2}$ respectively.

In particular, 
we will explore the following questions that
are important for evaluating an action formulation: (a)
How can we decide whether an action domain description is consistent
(has a model)? (b) Given an action domain description, 
how is a fluent's truth value affected by executing some acion(s)? (c)
Under what conditions does the reasoning within an action domain become
monotonic? and (d) Is it possible to characterize a set of
fluents that are {\em temporally definite} with respect to the underlying
action domain description? 
For instance, if 
fluent $F$'s truth value is known currently, will its truth value 
be also known after some action or action sequence is executed?
Furthermore, we will also discuss how to improve
our action formulation to handle domain
dependent preferences so that they can be suited for more 
general cases in reasoning about action.

\comment{

As we declared earlier, in our action formulation, ${\cal AT}^{0}$,
${\cal AT}^{1}$, and ${\cal AT}^{2}$ are only viewed as high level 
action description languages whose semantics is defined based
on PLPs. Hence, our investigation on an action domain will
depend on its PLP translation.
}

\subsection{Consistency of action domains} 

In this subsection, we consider the problem of how we can decide if a domain
description is consistent (has a model).
In our semantics development, the transition function
${\cal R}$ is defined based on a translation from the underlying 
domain description ${\cal D}$ to a PLP. 
Hence, it is not difficult obtain a general PLP characterization for
consistent domain descriptions.

\begin{proposition}
Let ${\cal D}$ be a domain description of ${\cal AT}^{i}$
and ${\cal P}^{{\cal AT}^{i}}({\cal D})$ ($i=0,1,2$) the corresponding PLP
translation of ${\cal D}$ specified previously.
${\cal D}$ is consistent if and only if
${\cal P}^{{\cal AT}^{i}}({\cal D})$ has a consistent answer set.
\end{proposition}

Proposition 1, however, can not always be used as a feasible way to 
decide the consistency of a domain description because in general deciding whether
a PLP has an answer set is NP-complete \cite{yan:00}\footnote{Note that 
deciding whether an extended logic program has an answer set is also 
NP-complete \cite{mt:default}.}.
So it is important to study
syntactic characterizations on different cases.
Our investigation on this issue starts from 
language ${\cal AT}^{0}$.

\subsubsection{Characterizing consistent action domains of ${\cal AT}^{0}$}

Given a domain description ${\cal D}$ of ${\cal AT}^{0}$,
we first introduce the following notions:
\begin{quote}
\noindent
${\cal F}^{+}_{Initial}=\{F\mid \mbox{\bf initially} F\in {\cal D}\}$,\\
${\cal F}^{-}_{Initial}=\{F\mid \mbox{\bf initially} \neg F\in {\cal D}\}$,\\
${\cal F}^{+}_{Effect}=\{F\mid A \hspace*{.05in}\mbox{\bf causes}\hspace*{.05in} F$
$\mbox{\bf if}\hspace*{.05in} L_{1},\cdots,L_{m}$
$\in {\cal D}\}$,\\
${\cal F}^{-}_{Effect}=\{F\mid A \hspace*{.05in}\mbox{\bf causes} \neg F$
$\mbox{\bf if} \hspace*{.05in}L_{1},\cdots,L_{m}$
$\in {\cal D}\}$,\\
${\cal F}^{+}_{Caused}=\{F\mid F \hspace*{.05in}\mbox{\bf is caused}$
$\mbox{\bf if} \cdots$
$\in {\cal D}\}$,\\
${\cal F}^{-}_{Caused}=\{F\mid \neg F \hspace*{.05in}\mbox{\bf is caused}$
$\mbox{\bf if} \cdots \in {\cal D}\}$.
\end{quote} 
For convenience, we use $\overline{{\cal F}^{-}_{Initial}}$ to denote the set containing
those complementary elements of ${\cal F}^{-}_{Initial}$. That is,
\begin{quote}
$\overline{{\cal F}^{-}_{Initial}}=\{\neg F\mid F\in {\cal F}^{-}_{Initial}\}$. 
\end{quote}
Similar notations may be 
used for other sets, e.g.
$\overline{{\cal F}^{-}_{Effect}}$, $\overline{{\cal F}^{-}_{Caused}}$, etc..

\begin{definition}
Given a domain description ${\cal D}$ of ${\cal AT}^{0}$, two
fluent expressions $L$ and $L'$ are {\em mutually exclusive}
in ${\cal D}$ if:
\begin{quote}
\noindent
$L\in ({\cal F}^{+}_{Initial}\cup \overline{{\cal F}^{-}_{Initial}}$
$\cup {\cal F}^{+}_{Effect} \cup \overline{{\cal F}^{-}_{Effect}}$ 
$\cup {\cal F}^{+}_{Caused} \cup \overline{{\cal F}^{-}_{Caused}})$ implies\\
$L'\not\in ({\cal F}^{+}_{Initial}\cup \overline{{\cal F}^{-}_{Initial}}$
$\cup {\cal F}^{+}_{Effect} \cup \overline{{\cal F}^{-}_{Effect}}$
$\cup {\cal F}^{+}_{Caused} \cup \overline{{\cal F}^{-}_{Caused}})$.
\end{quote}
\end{definition}

Intuitively, if two fluent expressions are mutually exclusive, it means that
these two fluent expressions cannot be both true in any state. 
%
%
Based on the concept of
mutual exclusion, we will provide a sufficient condition to decide the
consistency of a domain description.
Before we present the result, we need to introduce further notions.
For a domain description ${\cal D}$, we 
assign a unique label $l$ to each proposition in ${\cal D}$ so that 
we can use $l$ to refer a proposition in ${\cal D}$. Let
$l$ be a causal or action effect proposition
in ${\cal D}$. That is, $l$ has one of the following forms:
\begin{quote}
\noindent
$L$ {\bf is caused if} $L_{1},\cdots,L_{m}$ {\bf with absence} $L_{m+1},\cdots, L_{n}$, or\\
$A$ {\bf causes} $L$ {\bf if} $L_1,\cdots,L_m$.
\end{quote}
We use $pre(l)$, $default(l)$ and $eff(l)$ 
to denote the set $\{L_{1},\cdots,L_{m}\}$, 
$\{L_{m+1},\cdots, L_{n}\}$ and $\{L\}$ respectively.
Clearly, 
$default(l)=\emptyset$ if $l$ is an action effect proposition or
the causal proposition does not include absent fluent expressions.
For the case that $l$ is an initial proposition
{\bf initially} $L$, 
$pre(l)=default(l)=\emptyset$ and $eff(l)=\{L\}$.

\begin{definition}
Given a domain description ${\cal D}$ of ${\cal AT}^{0}$. Two propositions 
$l$ and $l'$ in ${\cal D}$ are {\em complementary} if 
one of the following conditions holds:
\begin{enumerate}
\item[(i)] both $l$ and $l'$ are causal propositions, and $eff(l)$ is a complement of
$eff(l')$; 
\item[(ii)] $l$ is a causal proposition,  $l'$ is an action effect proposition, and
$eff(l)$ is a complement of $eff(l')$, i.e.
\begin{quote}
$l$: $F$ {\bf is caused if} $L_{1},\cdots, L_{m}$ {\bf with absence} $L_{m+1},\cdots, L_{n}$, \\
$l'$: $A$ {\bf causes} $\neg F$ {\bf if} $L_{1}',\cdots, L_{k}'$;
\end{quote}
\item[(iii)] both $l$ and $l'$ are action effect propositions of the same action, and
$eff(l)$ is a complement of $eff(l')$, i.e. 
\begin{quote}
$l$: $A$ {\bf causes} $F$ {\bf if} $L_{1},\cdots, L_{h}$, \\
$l'$: $A$ {\bf causes} $\neg F$ {\bf if} $L_{1}',\cdots, L_{k}'$.
\end{quote}
\end{enumerate}
\end{definition}

\begin{definition}
Given a domain description ${\cal D}$ of ${\cal AT}^{0}$. ${\cal D}$ is 
{\em normal} if 
${\cal D}$ satisfies all of the following conditions.
\begin{enumerate}
\item[(i)] ${\cal F}^{+}_{Initial}\cap {\cal F}^{-}_{Initial}=\emptyset$;
\item[(ii)] For any two causal propositions $l_1$ and $l_2$ in ${\cal D}$, \\
$\overline{eff(l_i)}\cap pre(l_i)=\emptyset$ and \\
$default(l_i)\cap eff(l_j)=\emptyset$ ($i,j=1,2$)\footnote{Note that
this condition includes $default(l_i)\cap eff(l_i)=\emptyset$ ($i=1,2$).};
\item[(iii)] For any pair $(l,l')$ of complementary propositions in ${\cal D}$,
there is a pair of
fluent expressions $(L,L')$ in ${\cal D}$ such that $L$ and $L'$ are
mutually exclusive, where $L\in pre(l)$ and $L'\in pre(l')$.
\end{enumerate}
\end{definition}

Let us explain the intuition behind a normal domain description
in some details. Condition (i) ensures a consistent initial state deduced from
the domain description ${\cal D}$.
Condition (ii), on the other hand,  says that 
for each causal proposition in ${\cal D}$, the complement of its effect should
not occur in its preconditions, and furthermore, the effect of this causal 
proposition does not occur in the absence component (i.e. the default part) 
of all other (including itself) causal propositions in ${\cal D}$. 
Finally, Condition (iii) represents a non-trivial restriction for
complementary propositions in ${\cal D}$. 
Since two complementary propositions may cause two complementary fluents to be
true in some state, this condition actually indicates that 
if there are two complementary propositions in the domain description, then
the effects of these
two propositions cannot be both true in any state. 
The following theorem gives a sufficient condition to guarantee a 
domain description to be consistent.

\begin{theorem}
Every normal domain description of ${\cal AT}^{0}$ is consistent.
\end{theorem}

\subsubsection{Characterizing consistent action 
domains of ${\cal AT}^{1}$ and ${\cal AT}^{2}$}

Now we try to investigate an analogue of Theorem 2 for ${\cal AT}^{1}$
and ${\cal AT}^{2}$.
As ${\cal AT}^{2}$ is viewed as an extension of ${\cal AT}^{1}$, 
here we only need to 
consider domain descriptions of ${\cal AT}^{2}$.
To achieve our purpose, we must modify the concept of
mutual exclusion of fluent expressions in order to
cover observation and
abnormal action effect propositions in a domain description that are not allowed in
${\cal AT}^{0}$.
In particular, we define 
\begin{quote}
\noindent
${\cal F}^{+}_{\overline{A}}=\{F\mid L$ {\bf is observed if} $\cdots$ {\bf after} $\overline{A}\}$,\\
${\cal F}^{-}_{\overline{A}}=\{\neg F\mid L$ {\bf is observed if} 
$\cdots$ {\bf after} $\overline{A}\}$.
\end{quote}
As a special case, 
${\cal F}^{+}_{\epsilon}$ is formed based on initial observation propositions of ${\cal D}$.
Let 
\begin{quote}
\noindent
${\cal F}^{+}_{Observe}=\bigcup {\cal F}^{+}_{\overline{A}}$, and\\
${\cal F}^{-}_{Observe}=\bigcup {\cal F}^{-}_{\overline{A}}$,
\end{quote}
where each action string $\overline{A}$ occurs in some observation proposition of ${\cal D}$.
Under the context of ${\cal AT}^{2}$,
we also redefine the following notions:
\begin{quote}
\noindent
${\cal F}^{+}_{Effect} =\{F\mid A$ {\bf normally causes} $F$ {\bf if} $\cdots\}$ $\cup$\\
\hspace*{.65in} $\{F\mid A$ {\bf abnormally causes} $F$ {\bf if} $\cdots\}$,\\
${\cal F}^{-}_{Effect} =\{F\mid A$ {\bf normally causes} $\neg F$ {\bf if} $\cdots\}$ $\cup $\\
\hspace*{.65in} $\{F\mid A$ {\bf abnormally causes} $\neg F$ {\bf if} $\cdots\}$.
\end{quote}
Given domain description ${\cal D}$, we use label $l$ to 
(uniquely) refer to an observation proposition, causal proposition, action effect
proposition, or action abnormal effect proposition. Then 
notions $pre(l)$, $default(l)$ and $eff(l)$ are defined in an obvious way.
Two fluent expressions $L$ and $L'$ are {\em mutually exclusive} in ${\cal D}$ if
\begin{quote}
\noindent
$L\in ({\cal F}^{+}_{Observe}\cup \overline{{\cal F}^{-}_{Observe}}$
$\cup {\cal F}^{+}_{Effect} \cup \overline{{\cal F}^{-}_{Effect}}$
$\cup {\cal F}^{+}_{Caused} \cup \overline{{\cal F}^{-}_{Caused}})$ implies\\
$L'\not\in ({\cal F}^{+}_{Observe}\cup \overline{{\cal F}^{-}_{Observe}}$
$\cup {\cal F}^{+}_{Effect} \cup \overline{{\cal F}^{-}_{Effect}}$
$\cup {\cal F}^{+}_{Caused} \cup \overline{{\cal F}^{-}_{Caused}})$.
\end{quote}         

Finally, we should also modify the definition of 
complementary propositions as follows.

\begin{definition}
Given a domain description ${\cal D}$ of ${\cal AT}^{1}$ or
${\cal AT}^{2}$. Two propositions
$l$ and $l'$ in ${\cal D}$ are {\em complementary} if one of the following conditions
holds:      
\begin{enumerate}
\item[(i)] both $l$ and $l'$ are causal propositions and $eff(l)$ is a complement of
$eff(l')$;  
\item[(ii)] $l$ is a causal proposition and $l'$ is an action effect or abnormal 
effect proposition and $eff(l)$ is a complement of $eff(l')$;
\item[(iii)] both $l$ and $l'$ are action effect propositions 
of the same action where
$eff(l)$ is a complement of $eff(l')$, i.e.
\begin{quote}
$l$: $A$ {\bf normally causes} $F$ {\bf if} $L_{1},\cdots, L_{h}$, \\
$l'$: $A$ {\bf normally causes} $\neg F$ {\bf if} $L_{1}',\cdots, L_{k}'$; 
\end{quote} 
\item[(iv)] both $l$ and $l'$ are action abnormal effect propositions of the
same action and $eff(l)$ is a complementary of
$eff(l')$, i.e.
\begin{quote}
$l$: $A$ {\bf abnormally causes} $F$ {\bf if} $L_{1},\cdots, L_{h}$, \\
$l'$: $A$ {\bf abnormally causes} $\neg F$ {\bf if} $L_{1}',\cdots, L_{k}'$.
\end{quote}
\end{enumerate}
\end{definition}

The following definition then extends the concept of normal domain description to
${\cal AT}^{1}$ and ${\cal AT}^{2}$.

\begin{definition}
Given a domain description ${\cal D}$ of ${\cal AT}^{1}$ or ${\cal AT}^{2}$. 
${\cal D}$ is {\em normal} if
${\cal D}$ satisfies all of the following conditions.
\begin{enumerate}
\item[(i)] For any action string $\overline{A}$
occurring in observation propositions of ${\cal D}$, 
${\cal F}^{+}_{\overline{A}}\cap {\cal F}^{-}_{\overline{A}}=\emptyset$;
\item[(ii)] For any two observation or causal propositions $l_1$ and $l_2$ in ${\cal D}$,
$\overline{eff(l_i)}\cap pre(l_i)=\emptyset$ and
$default(l_i)\cap eff(l_j)=\emptyset$ ($i,j=1,2$);
\item[(iii)] For any pair $(l,l')$ of complementary propositions in ${\cal D}$,
there is a pair of
fluent expressions $(L,L')$ in ${\cal D}$ such that $L$ and $L'$ are
mutually exclusive, where $L\in pre(l)$ and $L'\in pre(l')$.
\end{enumerate}
\end{definition}

\begin{theorem}
Every normal domain description of ${\cal AT}^{1}$ or ${\cal AT}^{2}$ is
consistent.
\end{theorem}

\subsection{Cause of change on fluents' truth values}

In the rest of the paper, our discussion will focus on consistent action 
domains. First,
the following theorem illustrates a basic property of
any (consistent) action domain of ${\cal AT}^{0}$ showing that 
a fluent's truth value can only be affected
by some action effect proposition or causal proposition.

\begin{theorem}
Let ${\cal D}$ be a consistent domain description and 
${\cal P}^{{\cal AT}^{0}}({\cal D})$  the corresponding
PLP translation of ${\cal D}$. Then the following results hold.
\begin{enumerate}
\item[(i)] If ${\cal D} \models_{{\cal AT}^{0}} F$ {\bf after} $\overline{A}\cdot A$ and
${\cal D} \not\models_{{\cal AT}^{0}} F$ {\bf after} $\overline{A}$, then \\
${\cal P}^{{\cal AT}^{0}}({\cal D}) \models Effect^{+}(F,Result(A,S))$ or
${\cal P}^{{\cal AT}^{0}}({\cal D}) \models Caused^{+}(F,Result(A,S))$, where
$S=Result(A_l,\cdots,Result(A_1,S_{0})\cdots)$ and $\overline{A}=A_1\cdots A_l$\footnote{Without
further explanation, this 
notion is also used in our other statements presented in this section.};
\item[(ii)]
If ${\cal D} \models_{{\cal AT}^{0}} \neg F$ {\bf after} $\overline{A}\cdot A$ and
${\cal D} \not\models_{{\cal AT}^{0}} \neg F$ {\bf after} $\overline{A}$, then\\
${\cal P}^{{\cal AT}^{0}}({\cal D}) \models Effect^{-}(F,Result(A,S))$ or
${\cal P}^{{\cal AT}^{0}}({\cal D}) \models  Caused^{-}(F,Result(A,S))$.
\end{enumerate}
\end{theorem}

While the intuition of Theorem 4 is quite clear, it, however, 
does not hold for
action domains of ${\cal AT}^{1}$ and ${\cal AT}^{2}$ 
since observation propositions
of the form (\ref{s2.1}) is allowed
in a domain description of ${\cal AT}^{1}$ or ${\cal AT}^{2}$ that 
may override an inertia rule in the corresponding
PLP translation and
present a change of a fluent's truth value even if there is
no action or causal rule to cause such a change. In this case,
we may think that either the fluent's truth value
is changed by some external event that
is not described in the domain description or the 
domain description is not properly specified.
Weaker results may be obtained for domains of
${\cal AT}^{1}$ and ${\cal AT}^{2}$ under some restrictions.

\begin{theorem}
Let ${\cal D}$ be a consistent domain description of ${\cal AT}^{1}$ and
${\cal P}^{{\cal AT}^{1}}({\cal D})$ the corresponding PLP translation of ${\cal D}$.
Suppose each observation proposition in ${\cal D}$ has the form
\begin{quote}
$L$ {\bf is observed if} $L_{1},\cdots, L_{m}$ {\bf with absence} 
$\overline{L}, L_{m+1},\cdots, L_{n}$ 
{\bf after} $\overline{A}$,
\end{quote}
where $\overline{A}$ is not an empty string of actions.
Then the following results hold.
\begin{enumerate}
\item[(i)]
If ${\cal D}\models_{{\cal AT}^{1}} \neg F$ {\bf after} $\overline{A}$, and
${\cal D} \models_{{\cal AT}^{1}} F$ {\bf after} $\overline{A}\cdot A$,  then \\
${\cal P}^{{\cal AT}^{1}}({\cal D})\models Effect^{+}(F,Result(A,S))$ or
${\cal P}^{{\cal AT}^{1}}({\cal D}) \models Caused^{+}(F,Result(A,S))$;
\item[(ii)]
If ${\cal D}\models_{{\cal AT}^{1}} F$ {\bf after} $\overline{A}$ and
${\cal D} \models_{{\cal AT}^{1}} \neg F$ {\bf after} $\overline{A}\cdot A$, then \\
${\cal P}^{{\cal AT}^{1}}({\cal D}) \models Effect^{-}(F,Result(A,S))$ or
${\cal P}^{{\cal AT}^{1}}({\cal D}) \models Caused^{-}(F,Result(A,S))$.
\end{enumerate}
\end{theorem}

\begin{theorem}
Let ${\cal D}$ be a consistent domain description of ${\cal AT}^{2}$ and
${\cal P}^{{\cal AT}^{2}}({\cal D})$ the corresponding PLP translation of ${\cal D}$.
Suppose each observation proposition in ${\cal D}$ has the form
\begin{quote}
$L$ {\bf is observed if} $L_{1},\cdots, L_{m}$ {\bf with absence}
$\overline{L}, L_{m+1},\cdots, L_{n}$ {\bf after} $\overline{A}$,
\end{quote}
where $\overline{A}$ is not an empty string of actions.
Then the following results hold.
\begin{enumerate}
\item[(i)]
If ${\cal D} \models_{{\cal AT}^{2}} \neg F$ {\bf after} $\overline{A}$, and
${\cal D} \models_{{\cal AT}^{2}} F$ {\bf after} $\overline{A}\cdot A$, 
then one of the following
results holds:\\
${\cal P}^{{\cal AT}^{2}}({\cal D}) \models Effect^{+}(F,Result(A,S))$; \\
${\cal P}^{{\cal AT}^{2}}({\cal D}) \models  AbEffect^{+}(F,Result(A,S))$; or\\
${\cal P}^{{\cal AT}^{2}}({\cal D}) \models Caused^{+}(F,Result(A,S))$;
\item[(ii)]
If ${\cal D} \models_{{\cal AT}^{2}} F$ {\bf after} $\overline{A}$, and
${\cal D} \models_{{\cal AT}^{2}} \neg F$ {\bf after} $\overline{A}\cdot A$, 
then one of following results holds:\\
${\cal P}^{{\cal AT}^{2}}({\cal D}) \models Effect^{-}(F,Result(A,S))$;\\
${\cal P}^{{\cal AT}^{2}}({\cal D}) \models AbEffect^{-}(F,Result(A,S))$; or\\
${\cal P}^{{\cal AT}^{2}}({\cal D}) \models Caused^{-}(F,Result(A,S))$.
\end{enumerate}
\end{theorem}

\subsection{Restricted monotonicity}

Monotonicity is a desirable property for reasoning about action in the 
sense that whenever new domain specific information is added to 
a domain description, no previous conclusion will be retracted. However, it
is well known that most current action formulations are nonmonotonic 
in general. In this subsection, we investigate some restricted 
monotonicity for action domains.
Formally, let ${\cal D}$ be a domain description of 
${\cal AT}^{0}$, ${\cal AT}^{1}$, or ${\cal AT}^{2}$. A domain 
description ${\cal D'}$ is called an {\em augment} of ${\cal D}$ if
${\cal D}\subseteq {\cal D'}$ and the only extra propositions in
${\cal D'}$ are observation propositions (or initial propositions
in the case that
${\cal D}$ and ${\cal D'}$ are domain descriptions of ${\cal AT}^{0}$). 

\begin{definition}
A domain description ${\cal D}$ of ${\cal AT}^{i}$ ($i=0,1,2$)
is {\em monotonic with respect to
observations} (we also simply call 
{\em O-monotonic}) if
for each augment ${\cal D'}$ of ${\cal D}$,
${\cal D}\models_{{\cal AT}^{i}} L$ {\bf after} $\overline{A}$ implies
${\cal D'} \models_{{\cal AT}^{i}} L$ {\bf after} $\overline{A}$
($i=0,1,2$).
\end{definition}

It is clear that in general 
O-monotonicity does not hold for any domain description ${\cal D}$ due to
a possibility that in the PLP translation of ${\cal D'}$,
some new added observations
may defeat previous conclusions derived through 
defeasible causal rules, inertial rules, action effect rules or action
explanation rules. 
As an alternative, we can investigate proper 
restricted conditions under which O-monotonicity holds. 
\comment{
We first introduce the following notions for domain descriptions of 
${\cal AT}^{0}$.
\begin{quote}
${\cal F}^{+}_{Initial}=\{F\mid \mbox{\bf initially} F\in {\cal D}\}$,\\
${\cal F}^{-}_{Initial}=\{F\mid \mbox{\bf initially} \neg F\in {\cal D}\}$,\\
${\cal F}^{+}_{Effect}=\{F\mid A \hspace*{.05in}\mbox{\bf causes}\hspace*{.05in} F$
$\mbox{\bf if}\hspace*{.05in} L_{1},\cdots,L_{m}$
$\in {\cal D}\}$,\\
${\cal F}^{-}_{Effect}=\{F\mid A \hspace*{.05in}\mbox{\bf causes} \neg F$
$\mbox{\bf if} \hspace*{.05in}L_{1},\cdots,L_{m}$
$\in {\cal D}\}$,\\
${\cal F}^{+}_{Caused}=\{F\mid F \hspace*{.05in}\mbox{\bf is caused}$
$\mbox{\bf if} \cdots$
$\in {\cal D}\}$,\\
${\cal F}^{-}_{Caused}=\{F\mid \neg F \hspace*{.05in}\mbox{\bf is caused}$
$\mbox{\bf if} \cdots \in {\cal D}\}$.
\end{quote}
}

\begin{theorem}
Let ${\cal D}$ be a domain description ${\cal AT}^{0}$.
${\cal D}$ is O-monotonic if  
\begin{enumerate}
\item[(i)] each causal proposition in ${\cal D}$ is of the form
\begin{quote}
$L$ {\bf is caused if} $L_{1},\cdots, L_{m}$, and
\end{quote}
\item[(ii)] ${\cal F}^{+}_{Initial}\cap ({\cal F}^{-}_{Effect}$
$\cup {\cal F}^{-}_{Caused})=\emptyset$, \\
${\cal F}^{-}_{Initial}\cap ({\cal F}^{+}_{Effect}$
$\cup {\cal F}^{+}_{Caused})=\emptyset$, and\\
$({\cal F}^{+}_{Effect}\cup {\cal F}^{+}_{Caused})$ $\cap$
$({\cal F}^{-}_{Effect}\cup {\cal F}^{-}_{Caused})=\emptyset$.
\end{enumerate}
\end{theorem}

Intuitively, Theorem 7 says that to guarantee 
a domain description ${\cal D}$ of ${\cal AT}^{0}$ to be
O-monotonic, (i) all causal propositions in ${\cal D}$ should be non-defeasible,
and (ii) all fluents involved in initial propositions, action effect propositions and
causal propositions should be {\em irrelevant} in such a way:
fluents involved in positive (or negative, resp.)
initial propositions should be disjoint with fluents involved in
negative (or positive, resp.) action effect and causal propositions, and
fluents involved in positive action effect and causal propositions should be
disjoint with fluents involved in negative action effect and causal propositions.
Let ${\cal P}^{{\cal AT}^{0}}({\cal D})$ be the PLP translation of ${\cal D}$.
Condition (i) is necessary since this follows that adding any new
initial fact rules in ${\cal P}^{{\cal AT}^{0}}({\cal D})$
will not defeat any fact $Holds(F,S)$ or
$\neg Holds(F,S)$ that is drived through some 
causal rules at the initial situation
$S_{0}$. Condition (ii), on the other hand, guarantees that 
initiating any action effect rules or causal rules by adding new
initial fact rules into 
${\cal P}^{{\cal AT}^{0}}({\cal D})$
will not affect any previous facts drived through
old initial fact rules, action effect rules, or causal rules. 

An analogous result of Theorem 7, however, does not hold for 
domain descriptions of ${\cal AT}^{1}$ and ${\cal AT}^{2}$. In fact, 
since both ${\cal AT}^{1}$ and ${\cal AT}^{2}$
allow domain descriptions to have observations 
not only at the initial state but also at any 
other intermediate states, the property
of O-monotonicity is hard to
be achieved.
For instance, consider the PLP
translation ${\cal P}^{{\cal AT}^{1}}({\cal D})$
of a domain description ${\cal D}$ of 
${\cal AT}^{1}$, if ${\cal P}^{{\cal AT}^{1}}({\cal D})\models Holds(F,S)$ and $Holds(F,S)$
is derived through instances of action
explanation rules (\ref{4.1}) and (\ref{n4.2}) in ${\cal P}^{{\cal AT}^{1}}({\cal D})$:
\begin{eqnarray}
Effect^{+}(F',Result(A,S)) & \leftarrow & Holds(F',Result(A,S)),
not \hspace*{.05in} Holds(F',S), \nonumber \\
& & not \hspace*{.05in} Caused^{+}(F',Result(A,S)),
\label{pp.1}
\end{eqnarray}
\begin{eqnarray}
Holds(F,S) & \leftarrow & Effect^{+}(F',Result(A,S)), not \neg Holds(F,S),
\label{pp.2}
\end{eqnarray}
then adding a new observation rule $Holds(F',S)\leftarrow$ 
into ${\cal P}^{{\cal AT}^{1}}({\cal D})$, rule
$Holds(F',S)\leftarrow$ will always override rule (\ref{pp.1}) and the fact
$Holds(F,S)$ cannot be derived from the new PLP obtained by adding rule
$Holds(F',S)\leftarrow$ into ${\cal P}^{{\cal AT}^{1}}({\cal D})$.
A similar example can be given for a domain description of ${\cal AT}^{2}$ as well.
Therefore, in general, domain descriptions of 
${\cal AT}^{1}$ and ${\cal AT}^{2}$ are not O-monotonic under 
the condition of Theorem 7.

\subsection{Temporal definiteness}

Besides O-monotonicity, there is also
a class of action domains that satisfies a so-called temporal
definiteness property in temporal reasoning. 
Consider a domain description ${\cal D}$. We say that ${\cal D}$ is 
{\em temporally definite} if for any value proposition of the form
(\ref{s1.0}), ${\cal D}\models_{{\cal AT}^{i}} L$ {\bf after} $\overline{A}$
implies ${\cal D}\models_{{\cal AT}^{i}} L$ {\bf after} $\overline{A'}$ or
${\cal D}\models_{{\cal AT}^{i}} \overline{L}$ {\bf after} $\overline{A'}$ 
($i=0,1,2$),
where $\overline{A}$ is a substring of $\overline{A'}$, i.e. 
$\overline{A'}=\overline{A}\cdot A_{1}\cdots A_{k}$.
Intuitively, temporal definiteness expresses a kind of definite information on
fluents' truth values with respect to actions. For instance, 
if the switch is on initially, then we would expect that
no matter what actions are executed afterward, 
the switch should be either on or off. It would be 
undesirable if after executing some actions, the status of switch 
becomes {\em unknown}.

As only deterministic actions are considered in our context, 
the temporal definiteness seems a reasonable requirement for
our temporal reasoning. 
It is easy to verify that domain descriptions ${\cal D}(Switch$-$Power)$
and ${\cal D}(Switch$-$Power')$ described in section 3 are
temporally definite.
However, as defeasible information is allowed in domain descriptions,
this property does not always hold.

\begin{example}
Consider a scenario where there are constraints: (1) birds
normally can fly; (2) a wounded bird normally cannot fly.
Suppose we initially know that a specific bird Tweety is not wounded.  Then
after being shot, Tweety is wounded.
What we are interested in is whether Tweety can fly after she is shot.
We name this scenario $Shooting$-$3$ which
can be described by our action language ${\cal AT}^{0}$.
Let ${\cal D}(Shooting\mbox{-}3)$ be a domain description of ${\cal AT}^{0}$ including
the following propositions:
\begin{quote}
{\bf initially} $\neg Wounded$,\\
$Fly$ {\bf is caused if with absence} $\neg Fly$,\\
$\neg Fly$ {\bf is caused if} $Wounded$ {\bf with absence} $Fly$,\\
$Shoot$ {\bf causes} $Wounded$.
\end{quote}

Now we translate ${\cal D}(Shooting\mbox{-}3)$ into the corresponding PLP
${\cal P}^{{\cal AT}^{0}}(Shooting$-$3)=(\Pi_{spec}^{0}\cup\Pi_{ind}^{0},{\cal N},<)$,
where $\Pi_{spec}^{0}$ consists of the following rules\footnote{For simplicity,
here we omit the explicit description of
naming function ${\cal N}$ and $<$-relations.}:\\
\underline{Initial fact rule}:\\
$N_{1}: \neg Holds(Wounded,S_{0})\leftarrow$,\\
\underline{Causal rules}:\\
$N_{2}: Caused^{+}(Fly,s)\leftarrow$ {\em not} $\neg Holds(Fly,s)$,\\
$N_{3}: Caused^{-}(Fly,s)\leftarrow Holds(Wounded,s)$, {\em not} $Holds(Fly,s)$,\\
\underline{Action effect rule}:\\
$N_{4}: Effect^{+}(Wounded,Result(Shoot,s))\leftarrow$.

Then
it is easy to see
${\cal P}^{{\cal AT}^{0}}(Shoot$-$3)\models Holds(Fly,S_{0})$ (e.g. Tweety can fly
initially). Furthermore,
it is also not difficult to conclude that there exist two answer sets for
${\cal P}^{{\cal AT}^{0}}(Shoot$-$3)$ such that
$Holds(Fly,Result(Shoot,S_{0}))$ is in one answer set and
$\neg Holds(Fly,Result(Shoot,S_{0}))$ is in another.
So we have 
\begin{quote}
\noindent
${\cal D}(Shooting\mbox{-}3)\models_{{\cal AT}^{0}}$ {\bf initially} $Fly$,\\
${\cal D}(Shooting\mbox{-}3)\not\models_{{\cal AT}^{0}} Fly$ {\bf after} $Shoot$,\\
${\cal D}(Shooting\mbox{-}3)\not\models_{{\cal AT}^{0}} \neg Fly$ {\bf after} $Shoot$.
\end{quote}
So ${\cal D}(Shooting\mbox{-}3)$ is not temporally definite.
But intuitively,
we would prefer that Tweety cannot fly after being shot because
causal rule $N_{3}$ seems to be more specific than $N_{2}$. 
Solving this problem
involves the issue of representing domain-dependent preference which
will be discussed in section 6.5.
\end{example}

\begin{lemma}
A domain description ${\cal D}$ of
${\cal AT}^{i}$ ($i=0,1,2$) is temporally definite
if its PLP translation 
${\cal P}^{{\cal AT}^{i}}({\cal D})$ has a unique answer set.
\end{lemma}

The converse of Lemma 1, however, does not hold. That is,
for a temporally definite domain description,
its PLP translation may have more than one answer set.
For instance, in domain description ${\cal D}(Shooting$-$3)$ described above,
if we
initially know that Tweety is already wounded, then 
the modified domain description becomes temporally definite but its PLP
translation will still have 
more than one answer sets, i.e. one answer set
includes $Holds(Fly,S_{0})$ while the other includes $\neg Holds(Fly,S_{0})$.

Lemma 1 actually presents a sufficient condition to ensure
a domain description to be temporally definite.
Observing Example 10, we can see that ${\cal P}^{{\cal AT}^{0}}(Shooting$-$3)$ has
more than one answer set 
because two causal rules $N_{2}$ and $N_{3}$ conflict with each other
on fluent $Fly$'s truth value in situation $Result(Shoot,S_{0})$, while
$Fly$'s truth value is initially true, i.e.
${\cal P}^{{\cal AT}^{0}}(Shooting$-$3)\models Holds(Fly,S_{0})$.
This observation motivates our examination on the structure
of an action domain.

Consider an extended logic program $\Pi$. Using a procedure proposed
by Gelfond and Lifschitz (see Appendix A), we can actually
transform $\Pi$ into a general logic program\footnote{A
{\em general logic program} is a set of rules of the form
$A\leftarrow B_{1},\cdots,B_{m}, not B_{m+1},\cdots, not B_{n}$, 
where $A, B_{1},\cdots, B_{n}$ are atoms. Also see Appendix A.}, denoted
by $Trans(\Pi)$. It has been showed that
a sufficient condition to ensure that
$Trans(\Pi)$ has a unique stable model (or answer set under the context of 
extended logic program) is that $Trans(\Pi)$ is
{\em locally stratified}. That means, there does not exist any potential
conflict among any rules in $Trans(\Pi)$
(see Appendix A for a technical description on local
stratification). Therefore, we have the following result.

\begin{theorem}
A domain description ${\cal D}$ of ${\cal AT}^{i}$ ($i=0, 1, 2$) is 
temporally definite if 
its PLP translation ${\cal P}^{{\cal AT}^{i}}({\cal D})$
has a unique reduct $\Delta^{i}$ and
$Trans(\Delta^{i})$ is locally stratified.
\end{theorem}

Theorem 8 implies that to guarantee a domain description ${\cal D}$
to be temporally definite, 
no conflict should occur among the same type of
defeasible rules after reducing 
${\cal P}^{{\cal AT}^{i}}({\cal D})$ to
its reduct $\Delta^{i}$. In Example 10, since two causal rules
$N_{2}$ and $N_{3}$ contain a potential conflict 
with each other, it causes ${\cal D}(Shooting$-$3)$ to be 
not temporally definite. However, conflicts between
different types of defeasible rules will not affect the  temporal definiteness
for a domain description because such a conflict can be resolved during
the process of generating a reduct 
of the PLP translation of the domain description.

\subsection{Indefiniteness and domain-dependent preferences}

As we mentioned before,  temporal definiteness
is a desirable property in temporal reasoning.
However, it is also the fact that 
sometimes a domain description which is not
temporally definite may still present right results from our intuition.
For instance, in the domain of Switch-Power presented in section 3, 
if we add one more causal proposition into ${\cal D}(Switch\mbox{-}Power)$:
\begin{quote}
$\neg On$ {\bf is caused if with absence} $Power$.
\end{quote}
which says that if there is no explicit information stating
that there is power, then it is assumed that the light is not on, 
the circumstance will then change.
Suppose that initially we know that the light is not on, the switch is off, and
there is no any information
about if there is power. Then after turning on the switch, we would like to 
know whether the light is on.
It is not difficult to show that the modified domain description, 
say ${\cal D}(Switch$-$Power'')$, is not temporally definite. Specifically, 
we have 
\begin{quote}
\noindent
${\cal D}(Switch$-$Power'')\models_{{\cal AT}^{0}}$ {\bf initially} $\neg On$,\\
${\cal D}(Switch$-$Power'')\not\models_{{\cal AT}^{0}} On$ {\bf after} $Turn\mbox{-}On$,\\
${\cal D}(Switch$-$Power'')\not\models_{{\cal AT}^{0}}\neg On$ {\bf after} $Turn\mbox{-}On$.
\end{quote}

Although action $Turn$-$On$ is deterministic (see its effect proposition
in Example 4 in section 3.2.2), the above indefinite result
seems reasonable from our intuition because without having 
definite information about power, 
it is impossible to decide whether
the light is on after performing action 
$Turn$-$On$ due to a conflict between two causal propositions in 
${\cal D}(Switch$-$Power'')$.
%

This example reveals that although temporal definiteness
sometimes indeed describes
a desired property, it should not become
a particular restriction on action domains.
So far, in our domain descriptions, preferences are used as
built-in mechanisms of their PLP translations 
to handle conflicts among different
types of propositions. It is observed that 
domain-dependent preferences also
play important roles in temporal reasoning.
For instance, in some domains, it is the case that
within the same type of defeasible propositions,
one proposition is more preferred than the other. 
Consider Example 10 presented in section 6.4 once again.
We have mentioned that two causal propositions 
\begin{quote}
\noindent
$Fly$ {\bf is caused if with absence} $\neg Fly$,\\
$\neg Fly$ {\bf is caused if} $Wounded$ {\bf with absence} $Fly$,
\end{quote}
contain a conflict under the
circumstance by knowing that Tweety is wounded.
This conflict leads 
${\cal D}(Shooting$-$3)$ to be
temporally indefinite. But from our intuition,
the second causal proposition seems to
represent more specific information than the first 
causal proposition. 
Therefore, during the temporal reasoning, once conflict occurs
between these two  causal rules, we would prefer the second
causal proposition to defeat the first one
(e.g. the wounded bird Tweety normally cannot fly if we do not know she can fly).

This problem may be handled by including
domain-dependent preferences on causal  and 
observation propositions into the corresponding
PLP translations of domain descriptions. For instance, in Example 10,
we may add preference $N_{3}<N_{2}$ into 
${\cal P}^{{\cal AT}^{0}}(Shooting$-$3)$, and then
${\cal P}^{{\cal AT}^{0}}(Shooting$-$3)$ becomes temporally definite and 
the fact $\neg Fly$ {\bf after} $Shoot$ is entailed from the
modified domain description.

In general, to represent domain-dependent preferences in a domain
description, we need to extend the language so that 
preference between two propositions can be explicitly expressed.
One way of doing this is to introduce {\em labels} in the
language and each proposition in the domain description is assigned a unique
label. A  
{\em preference proposition} can be proposed as follows: 
\begin{eqnarray}
l_{1} \hspace*{.05in}\mbox{\bf is more preferred than} \hspace*{.05in} l_{2},
\label{6.0}
\end{eqnarray}
where $l_{1}$ and $l_{2}$ are labels for causal or
observation propositions in the domain description.
Then we define the PLP translation of the extended domain description
as $(\Pi,{\cal N},<\cup <_{C}\cup <_{O})$ ($i=0,1,2$),
where $\Pi$, ${\cal N}$ and 
$<$ are the same as before, and
$<_{C}$ and $<_{O}$ are the preference orderings on
causal and observation rules respectively that correspond to the specified 
preference propositions of the
form (\ref{6.0}) in the domain description.

\section{Related Work}

In this section, we discuss some related work.
In the research of reasoning about action, 
it is difficult to evaluate various action theories
from a systematic standard though some studies on this 
topic have been developed, e.g. \cite{book:shoot}. To compare with competing
approaches, people usually have to demonstrate their methods with a small
number of typical examples. It is still not clear yet 
what should be the unified standard for an action theory to satisfy.
We feel that it would be rather weak to compare our approach with other
action theories just through a small number of examples. As 
defeasibility handling is the central issue in our
action formulation proposed in this paper, we will
focus on this point as a major criterion to compare our approach 
with other methods.

An early effort on handling defeasible causal rules in 
reasoning about action was due to the author's previous work
\cite{z:comm96}, in which the author identified the 
restriction of McCain and Turner's causal theory of actions \cite{mt:ijcai95}
and claimed that in general a causal rule should be treated as 
a defeasible rule in order to solve the ramification problem 
properly. In \cite{z:comm96},
constraints (\ref{1.1}) and (\ref{1.2}) simply correspond to
defaults $Switch: On/On$ and $\neg Power:/\neg On$ respectively.
By combining Reiter's default theory \cite{r:default}
and Winslett's PMA \cite{w:pma}
the author developed a causality-based minimal change principle
for reasoning about action and change which subsumes McCain and Turner's 
causal theory.

Although the work presented in \cite{z:comm96} provided a natural way
to represent causality in reasoning about action,
there were several restrictions in this action theory. 
First, due to technical restrictions, only normal defaults or
defaults without justifications are the suitable forms to represent
causal rules in problem domains. Second, this action theory 
did not handle the other two major defeasibilities - 
defeasible observations and actions with defeasible and abnormal effects.

Probably Jab{\l}onowski, {\L}ukaszewicz and Madali\'{n}ska-Bugaj's
work \cite{jlm:action96} was one of the early efforts on handling the
problem of defeasible observations and actions with abnormal effects. 
Following Dijkstra's semantics on programming
languages \cite{lm:ijcai95}, they proposed an action theory in which 
both defeasible observations and actions with abnormal effects 
were expressible. Their work actually presented a few 
new features. For instance,
by employing Dijkstra's semantics in action theory, 
their method reduced the computational cost in action reasoning;
it also dealt with both temporal prediction and postdiction reasoning while
incomplete information is allowable in problem domains.

However, the major limitation of this approach is that it did not solve
the ramification problem properly. To deal with domain constraints
in action scenarios, the action theory has to be extended by  
adding statements like
\begin{quote}
$A;release(F_{1});\cdots;release(F_{n})$, 
\end{quote}
which means that fluents
$F_{1},\cdots,F_{n}$ involved in domain constraints may not obey the inertia
rule with respect to the performance of action $A$ \cite{lm:ijcai95}.
For example, if we combine a constraint like ``{\em the fact that
the turkey is not alive implies that the turkey is not walking}''
into the previous shooting scenario, in order to derive an indirect
effect $\neg Walk$ of action {\em Shoot}, a statement 
like $Shoot; release(Walk)$ has to be added into the action theory.
But to specify such statements, we have to know how each action 
exactly affects fluents involved in the domain constraint. Obviously
for a complex problem domain
this usually is not practicable without taking causality into account.
Not surprisingly, due to such 
restriction, this approach is also hard to be extended to handle 
defeasible constraints in reasoning about action.

Baral and Lobo recently also proposed an action formulation to address
the issue of defeasible constraints and actions with defeasible 
effects \cite{cl:action97}.
Following a similar spirit of Gelfond and Lifschitz's action
language ${\cal A}$ \cite{gl:action}, Baral and Lobo proposed an action 
language named ${\cal ADC}$ to describe action domains in which
both defeasible constraints and actions with defeasible 
effects are admitted. In their language ${\cal ADC}$ 
a defeasible constraint like (\ref{1.1}) is represented as 
\begin{quote}
$Switch$ {\bf normally suffices for} $On$
\end{quote}
and the defeasible {\em Shoot} action illustrated in Example 5 is 
represented as 
\begin{quote}
$Shoot$ {\bf normally causes} $\neg Alive$ {\bf if} $Loaded$.
\end{quote} 
As showed in \cite{cl:action97}, 
${\cal ADC}$ has a simple syntax. 
Based on an extended logic program translation, a
transition system is defined to provide a formal semantics of ${\cal ADC}$.

It is worth to mention that our idea of defining semantics for 
${\cal AT}^{0}$, ${\cal AT}^{1}$ and
${\cal AT}^{2}$ is similar to Baral and Lobo's proposal for 
${\cal ADC}$.  
Both of these two approaches directly use logic programs to 
define a transition system for the action language, instead of developing
a separate semantics like ${\cal A}$ language.
Also, both approaches define states in a different way from the 
standard ${\cal A}$ language, that is, instead of defining 
a state to be a truth value assignment on fluents, these two approaches
define a state to be a collection of fluent expressions so that 
incomplete information about fluents becomes allowable.

Nevertheless,
some restrictions exist in  
action language ${\cal ADC}$: it can only reason about forward, 
i.e. temporal prediction, and observations on 
intermediate situations  
and final situation are not expressible. Therefore, their
approach cannot deal with temporal postdiction. On the other hand, 
although actions with defeasible effects are allowed in the
domain description, it seems that the issue of solving conflicts
between defeasible action effect propositions 
and defeasible constraints was not addressed in detail. 

Finally, we briefly mention Geffner's recent work 
on causal theory of action \cite{gef:ijcai97} which
is closely related to models of causal 
reasoning based on Bayesian networks and structural equation models
\cite{gp:network}. To provide a well-founded solution to
the ramification problem,
Geffner claimed that causal rules of the domain should be
defeasible in general. Although with a very
different language and methodology, 
Geffner's system actually addressed the same 
problem discussed in \cite{z:comm96} and \cite{cl:action97}.
However, from the viewpoint of defeasibility handling, 
this system is restricted because defeasible observations and
actions with defeasible and abnormal effects were not considered.

\section{Conclusions}

We have developed a unified action formulation to handle
three types of defeasibilities in reasoning about action. 
Our formulation consists of three action languages
named ${\cal AT}^{0}$, ${\cal AT}^{1}$ and ${\cal AT}^{2}$ respectively.
We have showed that 
our action formulation is applicable to
both temporal prediction and postdiction with incomplete
information while defeasible constraints, defeasible 
observations and actions with defeasible and abnormal effects
are admitted. As discussed in the previous section, although
the issue of defeasibility in reasoning about action has been addressed by 
some researchers recently, our work presented here is the first effort
to handle various defeasible information in temporal reasoning by using
a prioritized 
logic programming approach. It enhances the viewpoint that the logic 
programming languages can be employed as efficient 
low level formal languages for reasoning about action.

Besides the author's
work \cite{zf97:logic2}, different prioritized logic programming 
formalisms have been proposed recently, e.g. \cite{b:logic,b2:logic,g:logic}. 
The reason why we choose our PLPs to develop our action formulation is 
as follows. First we think that
the answer set semantics for PLPs provides an intuitive and natural
interpretation for conflict resolution in logic programs, and hence it is easy
to use not only in reasoning about action, but also in other 
aspects of modeling system dynamics \cite{zf97:logic1,yan:ecai98}.
Second, a propositional prioritized logic 
programming system (PLPS) has been implemented recently
by the author and his students \cite{yan:aic01}.
We believe that our PLPS can finally provide a practical programming
language prototype for 
representing actions with the capability of
the defeasibility handling within the
framework we proposed in this paper.

The computational issue of prioritized logic programs has been addressed
in the author's another work \cite{yan:00}. Briefly, the author has proved that 
for a propositional prioritized logic program, deciding whether it has an
answer set is NP-complete, and deciding whether a given ground literal is entailed
from this prioritized logic program is $\Pi^{P}_{2}$-complete. 

It is also easy to observe that since a rule containing variables in a PLP is
viewed as a set of ground instances of this rule by replacing variables with
all possible constants occurring in the PLP, under the case that
a PLP does not have function symbols,
the number of defeated rules eliminated from this PLP as described in
Definition 2 is always finite. Hence, we can always compute a finite
reduct of such PLP\footnote{Note that Theorem 1 shows that every PLP has 
a reduct, but such a reduct may contain infinite rules.}.

In the case that there are function symbols occurring in a PLP,
the situation is different. Basically, the set of ground instances of
a rule, that includes variables and function symbols, may be infinite 
and therefore
it might be possible that there are infinite number of defeated rules
which should be eliminated from the original PLP. Under this situation,
a reduct containing infinite rules may be produced 
according to Definition 2.
From a practical viewpoint, we are only able to deal with
finite reducts. To overcome this problem, 
we can set a proper restriction
on the variable substitution. For instance, in the modified
Switch-Power domain discussed in Example 4 (see section 3.2.2),
if all we are interested is to know what are the effect after actions
{\em Cut-Power} and {\em Turn-On} are executed, then in the computation of
the answer set of ${\cal P}^{{\cal AT}^{0}}(Switch\mbox{-}Power')$,
we only need to consider situations $S_{0}$, $Result(Turn\mbox{-}On,S_{0})$,
$Result(Cut\mbox{-}Power,S_{0})$, and
$Result(Turn\mbox{-}On, Result(Cut\mbox{-}Power,S_{0}))$.
This implies that the ground form of PLP
${\cal P}^{{\cal AT}^{0}}(Switch\mbox{-}Power')$ only has finite rules and hence
it always has a finite reduct.

Finally, we should mention that currently our action formulation
cannot represent nondeterministic actions and disjunctive domain 
information. That is, we only consider 
deterministic problem domains in this paper. This is due to
the limit of prioritized logic programs inherited from extended logic programs.
But we would argue that our prioritized logic programs are extendedable 
to represent disjunctive information by using a similar method 
described in \cite{gl:elp} for extended logic
programs, and our action formulation can then be extended
to represent nondeterministic actions.

\section*{Acknowledgement}

The author thanks anonymous referees for many valuable comments on 
the early version of this paper.

\newpage

\section*{Appendix A: General Logic Programs and Stratification}

A {\em general logic program} is a finite set of rules of the form
\begin{eqnarray}
A & \leftarrow & B_{1},\cdots,B_{m}, not \hspace*{.05in} B_{m+1},
\cdots, not \hspace*{.05in} B_{n},
\label{append.1}
\end{eqnarray}
where $A, B_{1},\cdots, B_{m},\cdots, B_{n}$ are atoms.

Gelfond and Lifschitz developed a transformation to reduce an extended
logic program to a general logic program \cite{gl:elp}.  
Consider an extended logic program $\Pi$. For any predicate $P$ occurring in 
$\Pi$, let $P'$ be a new predicate of the same arity. The atom $P'(x)$
is called the {\em positive form} of the negative literal
$\neg P(x)$. Every positive literal is, by definition, its own positive form. The
positive form of a literal $L$ will be denoted by $L^{+}$.
$\Pi^{+}$ stands for the general program obtained from $\Pi$ by replacing each 
rule $L_{0}\leftarrow L_{1},\cdots,L_{m}$, {\em not} $L_{m+1},\cdots$, 
{\em not} $L_{n}$ in $\Pi$ by rule
\begin{eqnarray*}
L_{0}^{+}\leftarrow L_{1}^{+},\cdots,L_{m}^{+}, not L_{m+1}^{+},\cdots, not L_{n}^{+}.
\end{eqnarray*}

\begin{proposition} \cite{gl:elp}
A consistent set $S\subset Lit$ is an answer set of $\Pi$ iff 
$S^{+}$ is an answer set of $\Pi^{+}$.
\end{proposition}

\begin{definition} (Local stratification \cite{a:logic})\\
Let $\Pi$ be a general logic program. 
\begin{itemize}
\item A {\em local stratification} for 
$\Pi$ is a function $\psi$ from the Herbrand base of
$\Pi$, $B_{\Pi}$, to the countable ordinals.
\item Given a local stratification $\psi$, we extend it to 
ground negative literals\footnote{Note that
here we mean weak negation {\em not}.}
by setting $\psi(not$ $A)=\psi(A)+1$.
\item A rule with form (\ref{append.1})
of $\Pi$ is called {\em locally stratified with respect to a local
stratification} $\psi$ if for every ground instance of
(\ref{append.1}), 
\begin{eqnarray*}
A'\leftarrow B_{1}', \cdots, B_{m}', not \hspace*{.05in} B_{m+1}',\cdots, 
not \hspace*{.05in} B_{n}',
\end{eqnarray*}
\begin{quote}
$\psi(A')\geq \psi(B'_{i})$, where $1\leq i\leq m$, and\\
$\psi(A')\geq \psi(not B'_{j})$, where $m+1\leq i\leq n$.
\end{quote}
\item $\Pi$ is called {\em locally stratified with respect to a local
stratification} $\psi$ if all its rules are. $\Pi$ is called
{\em locally stratified} if it is locally stratified with 
respect to some local stratification.
\end{itemize}
\end{definition}

\comment{

\begin{definition} (Acyclic program \cite{a:logic})\\
Let $\Pi$ be a general logic program. 
\begin{itemize}
\item A {\em level mapping} for $\Pi$ is a function $l: B_{\Pi}\rightarrow N$ of
Herbrand base $B_{\Pi}$ of $\Pi$ to natural numbers. For $A\in B_{\Pi}$, $l(A)$
is the level of $A$.
\item Given a level mapping $l$, we extend it to ground negative literals
by putting $l(not$ $A)=l(A)$.
\item A rule of $\Pi$ with the form (\ref{append.1}) is called {\em acyclic 
with respect to a level mapping l} if for every ground instance
of (\ref{append.1}):
\begin{eqnarray*}
A'\leftarrow B_{1}', \cdots, B_{m}', not \hspace*{.05in} B_{m+1}',\cdots,
not \hspace*{.05in} B_{n}',
\end{eqnarray*}
there exists some
$L\in \{B_{1}',\cdots,B_{m}'$, not $B_{m+1}',\cdots$, not $B_{n}'\}$,
such that $l(L)< l(A')$.
\item $\Pi$ is called {\em acyclic with respect to 
a level mapping f} if all its rules are. $\Pi$ is called {\em acyclic} if
it is acyclic with respect to some level mapping.
\end{itemize}
\end{definition}

\begin{proposition} \cite{a:logic}
An acyclic general logic program is locally stratified.
\end{proposition}

}

\begin{proposition} \cite{gl:glp}
If a general logic program $\Pi$
is locally stratified, then by treating
$\Pi$ as an extended logic program where each rule does not contain
classical negation, it has a unique answer set.
\end{proposition}


\newpage

\section*{Appendix B: Proofs}

\noindent
{\bf Theorem 1}
{\em Every PLP has a reduct.}

\vspace*{.2in}
\noindent
To proof Theorem 1, we need to introduce the concept of $<$-partition for a PLP.

\begin{definition}
Let ${\cal P}=(\Pi,{\cal N}<)$ be an arbitrary PLP.
A $<$-{\em partition} of $\Pi$ in
${\cal P}$ is a finite collection $\{\Pi_{1},\cdots, \Pi_{k}\}$,
where $\Pi=\Pi_{1}\cup\cdots \cup\Pi_{k}$ and
$\Pi_{i}$ and $\Pi_{j}$ are disjoint for any $i\neq j$, such that
\begin{enumerate}
\item
${\cal N}(r)<{\cal N}(r')$ $\in {\cal P}(<)$ implies that
there exist some $i$ and $j$ ($1\leq i<j$) such that
$r'\in \Pi_{j}$ and
$r\in \Pi_{i}$;
\item
for each rule $r' \in \Pi_{j}$ ($j>1$),
there exists some rule $r\in \Pi_{i}$ ($1\leq i < j$) such that\\
${\cal N}(r)<{\cal N}(r') \in {\cal P}(<)$.
\end{enumerate}
\end{definition}

\begin{example} 
Consider a PLP ${\cal P}_{3}=(\Pi,{\cal N},<)$:
\begin{quote}          
\noindent              
${\cal P}_{3}$:\\                    
\hspace*{.2in} $N_{1}: A\leftarrow$ {\em not} $B$, {\em not} $C$,\\
\hspace*{.2in} $N_{2}: B\leftarrow$ {\em not} $\neg C$,\\
\hspace*{.2in} $N_{3}: C\leftarrow$ {\em not} $A$, {\em not} $\neg C$,\\
\hspace*{.2in} $N_{4}: \neg C\leftarrow$ {\em not} $C$, \\
\hspace*{.2in} $N_{1}<N_{2}, N_{2}<N_{4}, N_{3}<N_{4}$.
\end{quote}                          
It is easy to verify that a $<$-partition of $\Pi$ in
${\cal P}_{3}$ is $\{\Pi_{1}, \Pi_{2}, \Pi_{3}\}$, where
\begin{quote}         
\noindent               
$\Pi_{1}$:  \\                       
\hspace*{.2in} $N_{1}: A\leftarrow$ {\em not} $B$, {\em not} $C$, \\
\hspace*{.2in} $N_{3}: C\leftarrow$ {\em not} $A$, {\em not} $\neg C$,\\
$\Pi_{2}$:\\                         
\hspace*{.2in} $N_{2}: B\leftarrow$ {\em not} $\neg C$,\\
$\Pi_{3}$:  \\                       
\hspace*{.2in} $N_{4}: \neg C\leftarrow$ {\em not} $C$.
\end{quote}                          
In fact, this program has a unique answer set $\{B, C\}$.
\end{example}    

\begin{lemma}
Every prioritized logic program has a $<$-partition.
\end{lemma}                        

\noindent
\begin{proof}
For a given PLP ${\cal P}=(\Pi,{\cal N},<)$,
we construct a series of subsets of $\Pi$ as follows:\\
\hspace*{.1in} $\Pi_{1}=\{r\mid \mbox{there does not exist a rule } r'\in \Pi$
$\mbox{ such that } {\cal N}(r')<{\cal N}(r)\}$;\\
\hspace*{.1in} $\Pi_{i}=\{r\mid \mbox{for all rules such that }$
${\cal N}(r')<{\cal N}(r)$, $r'\in \bigcup_{j=1}^{i-1}\Pi_{j}\}$.\\
We prove that $\{\Pi_{1},\Pi_{2},\cdots\}$ is a
$<$-partition of ${\cal P}$.         
First, it is easy to see that $\Pi_{i}$ and $\Pi_{j}$ are disjoint.
Now we show that this partition satisfies Conditions 1 and 2 described in
Definition 17. Let ${\cal N}(r)<{\cal N}(r')\in {\cal P}(<)$. If
there does not exist any rule $r''\in \Pi$ such that
${\cal N}(r'')<{\cal N}(r)$, then $r\in\Pi_{1}$. Otherwise, there exists some
$i$ ($1<i$) such that $r\in \Pi_{i}$ and for all rules satisfying
${\cal N}(r'')<{\cal N}(r)$ $r''\in \Pi_{1}\cup\cdots\cup\Pi_{i-1}$.
Let $r'\in \Pi_{j}$. Since ${\cal N}(r)<{\cal N}(r')$, it follows that $1<j$.
From the construction of $\Pi_{j}$, we also conclude
$r\in \Pi_{1}\cup\cdots\cup\Pi_{j-1}$. Since $r'\in \Pi_{i}$,
it follows $i\leq j-1$. That is, $i<j$.
Condition 2 directly follows         
from the construction of the partition described above.
                                     
Now we show that $\{\Pi_{1},\Pi_{2}, \cdots\}$ must be a finite set.
Firstly, if $\Pi$ is finite, it is clear
$\{\Pi_{1},\Pi_{2}, \cdots\}$ must be a finite set.
If $\Pi$ contains infinite rules, then according to our assumption presented in
Section 2,                           
${\cal P}$ must be the ground instantiation of some program, say
${\cal P}^{*}=(\Pi^{*}, {\cal N}^{*}, <^{*})$ where $\Pi^{*}$ is finite.
Then we can use the same way to define
a $<$-partition for ${\cal P}^{*}$. Since
$\Pi^{*}$ is finite, the partition of ${\cal P}^{*}$ must be
also finite: $\{\Pi_{1}^{*}, \Pi_{2}^{*}, \cdots, \Pi_{k}^{*}\}$.
As ${\cal P}^{*}$ is well formed, it implies that for each $i$ ($i=1, 2, \cdots$),
$\Pi_{i}$ is the ground instantiation of $\Pi_{i}^{*}$. So
$\{\Pi_{1},\Pi_{2},\cdots\} =\{\Pi_{1},\Pi_{2},\cdots, \Pi_{k}\}$ which
is finite.
\end{proof}      

\noindent
\begin{proof} 
({\em Proof of Theorem 1})
Let ${\cal P}=(\Pi,{\cal N}, <)$. From Lemma 2, we can assume
$\Pi$ has a partition $\Pi=\Pi_1\cup\cdots\cup\Pi_k$. We will show that 
${\cal P}$ has a fixpoint in the process of reduction according to Definition 2.
As $\Pi_1,\cdots,\Pi_k$ are disjoint and for any ${\cal N}(r)<{\cal N}(r')$, it
implies $r\in \Pi_i$ and $r'\in \Pi_j$ where $i<j$, 
we can use notation 
\begin{quote}
$\Pi_1 <\Pi_2<\cdots<\Pi_k$
\end{quote}
to illustrate this property. It is easy to see
that for each rule in $\Pi_{i}$ ($1<i<k$), there
must exist some $j$ and $h$ that $j<i<h$ such that
${\cal N}(r')<{\cal N}(r)<{\cal N}(r'')$ and  $r'\in \Pi_j$, $r''\in \Pi_h$.
Now we construct a sequence of
reductions that starts from those
least preferred rules in $\Pi_k$, then from rules in $\Pi_{k-1}\cup \Pi_k$, and so on
as illustrated below:
\begin{quote}
\noindent
$\Pi^{(0)}=\Pi=\Pi_1\cup\cdots\cup\Pi_k$;\\
$\Pi^{(1)}=\Pi^{(0)} -\{r_1,r_2,\cdots\mid r_1,r_2,\cdots \in \Pi_{k}$ and $r_1,r_2,\cdots$ satisfy 
the conditions \\
\hspace*{1.7in} as  stated in Definition 2$\}$;\\
$\Pi^{(2)} = \Pi^{(1)} - \{r_1,r_2,\cdots\mid r_1,r_2,\cdots \in \Pi_{k-1}\cup \Pi_k$ 
and $r_1,r_2,\cdots$ satisfy  \\ 
\hspace*{1.7in}  the conditions as stated in  Definition 2$\}$;\\
$\Pi^{(3)} = \Pi^{(2)} - \{r_1,r_2,\cdots\mid r_1,r_2,\cdots \in \Pi_{k-2}\cup \Pi_{k-1}\cup\Pi_k$
and $r_1,r_2,\cdots$ satisfy \\
\hspace*{1.7in} the conditions as stated in  Definition 2$\}$;\\
$\cdots$\\
$\Pi^{(k-1)} = \Pi^{(k-2)} -\{r_1,r_2,\cdots\mid r_1,r_2,\cdots \in \Pi_2\cup\cdots\cup\Pi_k$
and $r_1,r_2,\cdots$ satisfy \\
\hspace*{2in} the conditions as stated in Definition 2$\}$.
\end{quote}

It is observed that in the above 
reduction process,  after obtaining $\Pi^{(k-1)}$, no more rules can be 
eliminated from $\Pi^{(k-1)}$ by applying the conditions of Definition 2 because
after the $i$th reduction, all orderings inherited from
$\Pi_{k-i+1}<\cdots<\Pi_{k}$ will no longer play any roles in the further 
$(i+1)$th, $\cdots$, and $(k-1)$th reductions.
In particular,
in the $i$th reduction of obtaining $\Pi^{(i)}$,
all rules eliminated from $\Pi^{(i-1)}$ (note that
there may be infinite number of rules to be
eliminated in the $i$th reduction) are due to some
rules in $\Pi_{1}\cup\cdots\cup\Pi_{k-i}$ which are more preferred than 
those eliminated rules in $\Pi_{k-i+1}\cup\cdots\cup\Pi_{k}$. As $k$ is a finite number,
from Definition 2 $\Pi^{(k-1)}$ is also a reduct of ${\cal P}$.
\end{proof}

\noindent
{\bf Proposition 1}
{\em 
Let ${\cal D}$ be a domain description of ${\cal AT}^{i}$
and ${\cal P}^{{\cal AT}^{i}}({\cal D})$ ($i=0,1,2$) the corresponding PLP
translation of ${\cal D}$ specified previously.
${\cal D}$ is consistent if and only if
${\cal P}^{{\cal AT}^{i}}({\cal D})$ has a consistent answer set.
}

\vspace*{.1in}
\noindent
\begin{proof}
Here we only prove the result for ${\cal AT}^{0}$, proofs for other 
cases are similar. \\
Suppose ${\cal D}$ has a model $\Psi$. Then according to 
Definition 7, for any action string $\overline{A}$ such that
$\Psi(\overline{A})$ is defined  and any fluent $F$, 
$F$ and $\neg F$ cannot be both true in $\Psi(\overline{A})$. 
From the definition of $\Psi(\overline{A})$, i.e. Definition 6, it follows that
$\Psi(\overline{A'}\cdot A)\in {\cal R}(A,\Psi(\overline{A'}))$, where
$\overline{A}=\overline{A'}\cdot A$. Here we assume that
$\overline{A}$ is not empty (otherwise, $\Psi(\epsilon)=\hat{S}_{0}$ that we 
will consider next). Also since
$\Psi$'s domain is prefix closed, $\Psi(\overline{A'})$ is also defined.
Then from Defintiion 5 of transition function ${\cal R}$, it follows that 
${\cal R}(A,\Psi(\overline{A'})$ contains a consistent set of fluent expressions.
As this set is directly deduced from some answer set 
$Ans$ of ${\cal P}^{{\cal AT}^{0}}({\cal D})$,
it concludes that the subset of $Ans$ consisting of
all literals of the form $Holds(F,S)$ or
$\neg Holds(F,S)$ is consistent (note $S\neq S_{0}$).
Now we consider the case of empty action string. In this case
$\Psi(\epsilon)=\hat{S}_{0}$. As $\Psi$ is a model, $\hat{S}_{0}$ must be a consistent
set. Again, as $\hat{S}_{0}$ is deduced from some answer set $Ans$ of 
${\cal P}^{{\cal AT}^{0}}({\cal D})$, it concludes that
the subset of $Ans$ consisting all literals of the form 
$Holds(F,S_{0})$ or $\neg Holds(F,S_{0})$ is consistent. 
Therefore, the subset of $Ans$ of the following form is consistent:
\begin{quote}
$\{[\neg ] Holds(F,S_{0}),\cdots\}\cup\cdots\{[\neg ] Holds(F,S),\cdots\}$. 
\end{quote}
Recall that $Ans$ also contains a subset that consists of atoms 
of the forms $Effect^{+}(F,S)$, 
$Effect^{-}(F,S)$, $Caused^{+}(F,S)$ and $Caused^{-}(F,S)$. 
Clearly, this subset of $Ans$ is also consistent.
So $Ans$ is consistent.

Now suppose $Ans$ is a consistent answer set of
${\cal P}^{{\cal AT}^{0}}({\cal D})$. 
Then from Definitions 5, 6, and 7, we can construct a model $\Psi$ for 
${\cal D}$ in an obvious way.
\end{proof}

\noindent
{\bf Theorem 2}
{\em Every normal domain description of ${\cal AT}^{0}$ is consistent.
}

\vspace*{.1in}
\noindent
\begin{proof}
Let ${\cal D}$ be a normal domain description of ${\cal AT}^{0}$.
That is, ${\cal D}$ satisfies Conditions
(i), (ii) and (iii) in Definition 12. 
According to Proposition 1, we only need to show that the
PLP translation ${\cal P}^{{\cal AT}^{0}}({\cal D})$ of ${\cal D}$ 
has a consistent answer set.
Let ${\cal P}^{{\cal AT}^{0}}({\cal D})=(\Pi,{\cal N},<)$. 
First, from Condition (ii) and the construction of ${\cal P}^{{\cal AT}^{0}}({\cal D})$, 
it is observed that $\Pi$ does not contain rules of the following forms:
\begin{quote}
$r_1$: \hspace*{.2in} $L_1\leftarrow \cdots, not L^{\ast},\cdots$,\\
$r_2$: \hspace*{.2in} $L_2\leftarrow \cdots, L_1, \cdots$,\\
$\cdots$,\\
$r_{k}$: \hspace*{.2in} $L_{k}\leftarrow \cdots, L_{k-1}, \cdots$,\\
$r_{k+1}$: \hspace*{.05in} $L^{\ast}\leftarrow \cdots, L_{k},\cdots$.
\end{quote}
This actually ensures that $\Pi$ has an answer set $Ans$. To show this, 
we assume that $\Pi$ does not have an answer set. Then there must 
exist some literal $L^{\ast}$ satisfying the condition:
for any set $S$ of ground literals ($S$ can be empty) (a) if $L^{\ast}\not\in S$, then 
$L^{\ast}$ is in the answer set of program $\Pi^{S}$ ($\Pi^{S}$ is obtained from
$\Pi$ by doing Gelfond-Lifschitz transformation on $\Pi$ in terms of $S$); 
and (b) if $L^{\ast}\in S$, then
$L^{\ast}$ is not in the answer set of program $\Pi^{S}$.
It is worth to mention that since $\Pi^{S}$ does not contain rules including negation
as failure sign,  $\Pi^{S}$ always has an answer set.
From case (a), it is implied that $\Pi$ must contain a rule of the form:
\begin{quote}
$r'_{k+1}$: $L^{\ast}\leftarrow \cdots$.
\end{quote}
On the other hand, 
from case (b), it is easy to observe that 
all rules of the form $r_{k+1}'$ cannot be triggered in $\Pi^{S}$
due to $L^{\ast}\in S$. That is, some rule of the form
\begin{quote}
$r_1$: $L_1\leftarrow \cdots, not L^{\ast},\cdots$
\end{quote}
must be contained in $\Pi$
(we do not exclude the case that $L_1=L^{\ast}$).
This follows that rule $r'_{k+1}$ actually has a form:
\begin{quote}
$r'_{k+1}$: $L^{\ast}\leftarrow \cdots, L',\cdots$
\end{quote}
such that the deletion of $r_1$ from $\Pi$ will cause 
literal $L'$ not to be triggered and hence $L^{\ast}$ can 
not be derived from $\Pi^{S}$. 
Without loss of generality, we can assume that $\Pi$ contains a sequence of
rules $r_1,\cdots, r_{k+1}$ as described above.

Now we consider Condition (i). From Condition (i), 
we know that $\Pi$ does not contain a pair of rules of the forms:
\begin{quote}
$Holds(F,S_{0})\leftarrow$,\\
$\neg Holds(F,S_{0})\leftarrow$.
\end{quote}
This follows that a subset of $Ans$ in which each liteal is associated with 
initial situation $S_{0}$:
\begin{quote}
$\{[\neg ] Holds(F_1,S_{0}), \cdots, [\neg ] Holds(F_k,S_{0})\}$
\end{quote}
is consistent.
Now we consider a pair of complementary propositions $(l,l')$ in ${\cal D}$.
To simplify our presentation, for a rule of the form:
\begin{quote}
$r$: $L_0\leftarrow L_1,\cdots, L_m, not L_{m+1},\cdots, not L_n$
\end{quote}
we denote $pos(r)=\{L_{1},\cdots, L_m\}$ and
$neg(r)=\{L_{m+1},\cdots, L_n\}$. Then $r$ can be simply represented as
$L_{0}\leftarrow pos(r),neg(r)$.
Under this notation, a pair of complementary propositions
$l$ and $l'$ in ${\cal D}$
may have one of the following possible translations in $\Pi$:
\begin{quote}
\noindent
(a)\\
$r$: $Caused^{+}(F,s)\leftarrow pos(r), neg(r)$,\\
$r'$: $Caused^{-}(F,s)\leftarrow pos(r'), neg(r')$,\\
(b)\\
$r$: $Caused^{+}(F,s)\leftarrow pos(r), neg(r)$,\\
$r'$: $Effect^{-}(F,Result(A,s))\leftarrow pos(r'), neg(r')$,\\
(c)\\
$r$: $Effect^{+}(F,Result(A,s))\leftarrow pos(r), neg(r)$,\\
$r'$: $Effect^{-}(F,Result(A,s))\leftarrow pos(r'), neg(r')$.
\end{quote}
From Condition (iii), we know that in each case of (a), (b) and (c), 
$pos(r)$ and $pos(r')$ cannot be both true in answer set $Ans$. Hence, 
for any situation term $S$, none of these three pairs of atoms
$Caused^{+}(F,S)$ and $Caused^{-}(F,S)$, 
$Caused^{+}(F,S)$ and $Effect^{-}(F,S)$, or
$Effect^{+}(F,S)$ and $Effect^{-}(F,S)$ cannot both true in $Ans$. 
This concludes that $Ans$ does not contain any complementary literals
$Holds(F,S)$ and $\neg Holds(F,S)$ for any $F$ and $S$.
So $Ans$ is a consistent answer set of $\Pi$. Furthermore, every answer set of $\Pi$ is
also consistent \cite{lt:lp94}.
Finally, from the property that a PLP $(\Pi,{\cal N},<)$ has an answer set iff
$\Pi$ has an answer set and every answer set of $(\Pi,{\cal N},<)$ is also an
answer set of  $\Pi$ \cite{yan:00}, it concludes that ${\cal P}^{{\cal AT}^{0}}({\cal D})$
has a consistent answer set (and its every answer set is also consistent).
\end{proof}

\noindent
{\bf Theorem 3}
{\em
Every normal domain description of ${\cal AT}^{1}$ or
${\cal AT}^{2}$ is consistent.
}

\noindent
\begin{proof}
The proof is similar to the proof of Theorem 2 but with additional considerations on 
action explanation rules and action abnormal effect rules in 
${\cal P}^{{\cal AT}^{i}}({\cal D})$. We omit it here.
\end{proof}

\noindent
{\bf Theorem 4}
{\em Given a domain description ${\cal D}$ of ${\cal AT}^{0}$ and its
PLP translation ${\cal P}^{{\cal AT}^{0}}({\cal D})$,
the following results hold.
\begin{enumerate}
\item[(i)] If ${\cal P}^{{\cal AT}^{0}}({\cal D}) \models Holds(F,Result(A,S))$ and
${\cal P}^{{\cal AT}^{0}}({\cal D}) \not\models Holds(F,S)$, then\\ 
${\cal P}^{{\cal AT}^{0}}({\cal D}) \models Effect^{+}(F,Result(A,S))$ or
${\cal P}^{{\cal AT}^{0}}({\cal D}) \models Caused^{+}(F,Result(A,S))$;
\item[(ii)]
If ${\cal P}^{{\cal AT}^{0}}({\cal D}) \models\neg Holds(F,Result(A,S))$ and
${\cal P}^{{\cal AT}^{0}}({\cal D}) \not\models \neg Holds(F,S)$, then\\
${\cal P}^{{\cal AT}^{0}}({\cal D}) \models Effect^{-}(F,Result(A,S))$ or
${\cal P}^{{\cal AT}^{0}}({\cal D}) \models  Caused^{-}(F,Result(A,S))$.
\end{enumerate}
}

\noindent
\begin{proof}
It is sufficient to only prove (i). Since 
${\cal P}^{{\cal AT}^{0}}({\cal D})\models Holds(F,Result(A,S))$
and ${\cal P}^{{\cal AT}^{0}}({\cal D})\not\models Holds(F,S)$, it follows that 
for each answer set $Ans$ of ${\cal P}^{{\cal AT}^{0}}({\cal D})$, 
$Holds(F,Result(A,S))\in Ans$, and there exists some 
answer set $Ans'$ such that \\
$Holds(F,S)\not\in Ans'$.
Therefore, the fact that $Holds(F,Result(S,A))$ is true is {\em not} due to 
inertia rules (\ref{3.9}) and (\ref{3.10}) in ${\cal P}^{{\cal AT}^{0}}({\cal D})$, 
but due to 
action effect rules (\ref{3.5}) and (\ref{3.6}), or causal rules 
(\ref{3.1}) and (\ref{3.2}). That is, 
${\cal P}^{{\cal AT}^{0}}\models Effect^{+}(F,Result(A,s))$ or
${\cal P}^{{\cal AT}^{0}}\models Caused^{+}(F,Result(A,S))$.
\end{proof}

\noindent
{\bf Theorem 5}
{\em 
Let ${\cal D}$ be a domain description of ${\cal AT}^{1}$  and
${\cal P}^{{\cal AT}^{1}}({\cal D})$ be its PLP translation.
Suppose each observation proposition in ${\cal D}$ has the form
\begin{quote}
$L$ {\bf is observed if} $L_{1},\cdots, L_{m}$ {\bf with absence}
$\overline{L}, L_{m+1},\cdots, L_{n}$ {\bf after} $\overline{A}$,
\end{quote}
where $\overline{A}$ is not an empty string of actions.
Then the following results hold.
\begin{enumerate}
\item[(i)]
If ${\cal P}^{{\cal AT}^{1}}({\cal D})\models\neg Holds(F,S)$ and
${\cal P}^{{\cal AT}^{1}}({\cal D}) \models Holds(F,Result(A,S))$, then \\
${\cal P}^{{\cal AT}^{1}}({\cal D})\models Effect^{+}(F,Result(A,S))$ or
${\cal P}^{{\cal AT}^{1}}({\cal D}) \models Caused^{+}(F,Result(A,S))$;
\item[(ii)]
If ${\cal P}^{{\cal AT}^{1}}({\cal D})\models Holds(F,S)$ and
${\cal P}^{{\cal AT}^{1}}({\cal D}) \models \neg Holds(F,Result(A,S))$, then \\
${\cal P}^{{\cal AT}^{1}}({\cal D}) \models Effect^{-}(F,Result(A,S))$ or
${\cal P}^{{\cal AT}^{1}}({\cal D}) \models Caused^{-}(F,Result(A,S))$.
\end{enumerate}
}

\noindent
\begin{proof}
It is sufficient to only prove (i). As
${\cal P}^{{\cal AT}^{1}}({\cal D})\models \neg Holds(F,S)$ and
${\cal P}^{{\cal AT}^{1}}({\cal D})\models Holds(F,Result(A,S))$, it is clear that 
the fact that $Holds(F,Result(A,S))$ is true in each answer set
of ${\cal P}^{{\cal AT}^{1}}({\cal D})$ is {\em not} due to 
inertia rules (\ref{3.9}), (\ref{3.10}), (\ref{4.5}) and
(\ref{4.6}), but due to 
\begin{enumerate}
\item[(1)] some observation rules in ${\cal P}^{{\cal AT}^{1}}({\cal D})$, or
\item[(2)] action effect rules (\ref{3.5}) and (\ref{3.6}), or
\item[(3)] causal rules (\ref{3.1}) and (\ref{3.2}).
\end{enumerate}
Consider case (1). We suppose there exists some observation rule
in ${\cal P}^{{\cal AT}^{1}}({\cal D})$ of the form
\begin{eqnarray*}
Holds(F,Result(A,S)) & \leftarrow &
[\neg ] Holds(F_{1},S),\cdots, [\neg ] Holds(F_{m},S), \nonumber \\
 & & not \hspace*{.05in} [\neg ] Holds(F_{m+1},S),\cdots,
not \hspace*{.05in} [\neg ] Holds(F_{n},S),
\end{eqnarray*}
But from the condition, we know that the above observation
rule must be of the form:
\begin{eqnarray*}
Holds(F,Result(A,S)) & \leftarrow & \cdots, not \neg Holds(F,Result(A,S)), \cdots.
\end{eqnarray*}
This results in a conflict with inertia rule (\ref{3.10}):
\begin{eqnarray*}
Holds(f,Result(a,s)) & \leftarrow & Holds(f,s),
not \hspace*{.05in} \neg Holds(f,Result(a,s))
\end{eqnarray*}
in ${\cal P}^{{\cal AT}^{1}}({\cal D})$. As we specify inertia rules have higher
priorities than observation rules in ${\cal P}^{{\cal AT}^{1}}({\cal D})$,
it turns out that 
$\neg Holds(F,Result(A,S))$ is derived. So case (1) is impossible.
Hence, only cases (2) or (3) is possible to derive
$Holds(F,Result(A,s))$. That is, 
${\cal P}^{{\cal AT}^{1}}({\cal D})\models Effect^{+}(F,Result(A,S))$, or
${\cal P}^{{\cal AT}^{1}}({\cal D})\models Caused^{+}(F,Result(A,S))$.
\end{proof}

\noindent
{\bf Theoem 6}
{\em 
Let ${\cal D}$ be a domain description of ${\cal AT}^{2}$ and
${\cal P}^{{\cal AT}^{2}}({\cal D})$ be its PLP translation.
Suppose each observation proposition in ${\cal D}$ has the form
\begin{quote}
$L$ {\bf is observed if} $L_{1},\cdots, L_{m}$ {\bf with absence}
$\overline{L}, L_{m+1},\cdots, L_{n}$ {\bf after} $\overline{A}$,
\end{quote}
where $\overline{A}$ is not an empty string of actions.
Then the following results hold.
\begin{enumerate}
\item[(i)]
If ${\cal P}^{{\cal AT}^{2}}({\cal D}) \models\neg Holds(F,S)$ and
${\cal P}^{{\cal AT}^{2}}({\cal D}) \models Holds(F,Result(A,S))$, 
then one of following
results holds:\\
${\cal P}^{{\cal AT}^{2}}({\cal D}) \models Effect^{+}(F,Result(A,S))$; \\
${\cal P}^{{\cal AT}^{2}}({\cal D}) \models  AbEffect^{+}(F,Result(A,S))$; or\\
${\cal P}^{{\cal AT}^{2}}({\cal D}) \models Caused^{+}(F,Result(A,S))$;
\item[(ii)]
If ${\cal P}^{{\cal AT}^{2}}({\cal D}) \models Holds(F,S)$ and
${\cal P}^{{\cal AT}^{2}}({\cal D}) \models \neg Holds(F,Result(A,S))$,
then one of following results holds:\\
${\cal P}^{{\cal AT}^{2}}({\cal D}) \models Effect^{-}(F,Result(A,S))$;\\
${\cal P}^{{\cal AT}^{2}}({\cal D}) \models AbEffect^{-}(F,Result(A,S))$; or\\
${\cal P}^{{\cal AT}^{2}}({\cal D}) \models Caused^{-}(F,Result(A,S))$.
\end{enumerate}
}

\noindent
\begin{proof}
The proof of Theorem 6 is similar to that of Theorem 5 as
described above.
\end{proof}

\noindent
{\bf Theorem 7}
{\em
Let ${\cal D}$ be a domain description ${\cal AT}^{0}$.
${\cal D}$ is O-monotonic if
\begin{enumerate}
\item[(i)] each causal proposition in ${\cal D}$ is of the form
\begin{quote}
$L$ {\bf is caused if} $L_{1},\cdots, L_{m}$, and
\end{quote}
\item[(ii)] ${\cal F}^{+}_{Initial}\cap ({\cal F}^{-}_{Effect}$
$\cup {\cal F}^{-}_{Caused})=\emptyset$,
${\cal F}^{-}_{Initial}\cap ({\cal F}^{+}_{Effect}$
$\cup {\cal F}^{+}_{Caused})=\emptyset$, and
$({\cal F}^{+}_{Effect}\cup {\cal F}^{+}_{Caused})\cap$
$({\cal F}^{-}_{Effect}\cup {\cal F}^{-}_{Caused})=\emptyset$.
\end{enumerate}
}

\noindent
\begin{proof}
Let ${\cal D'}$ be an augment of ${\cal D}$, 
${\cal P}^{{\cal AT}^{0}}({\cal D})$ and
${\cal P}^{{\cal AT}^{0}}({\cal D'})$ be the PLP translations of
${\cal D}$ and
${\cal D'}$ respectively. To prove the result,
it is sufficient to prove that
${\cal P}^{{\cal AT}^{0}}({\cal D})\models Holds(F,S)$ implies 
${\cal P}^{{\cal AT}^{0}}({\cal D'})\models Holds(F,S)$.
From the construction of 
${\cal P}^{{\cal AT}^{0}}({\cal D})$, it is clear that 
${\cal P}^{{\cal AT}^{0}}({\cal D})\models Holds(F,S)$ implies
\begin{enumerate}
\item[(1)] ${\cal P}^{{\cal AT}^{0}}({\cal D})\models Effect^{+}(F,S)$, 
\item[(2)] ${\cal P}^{{\cal AT}^{0}}({\cal D})\models Caused^{+}(F,S)$, or
\item[(3)] ${\cal P}^{{\cal AT}^{0}}({\cal D})\models Holds(F,S')$
due to
inertia rules in ${\cal P}^{{\cal AT}^{0}}({\cal D})$,
where $S=Result(A,S')$.
\end{enumerate}
Adding more observation propositions into ${\cal D}$ to 
form ${\cal D'}$, the new program
${\cal P}^{{\cal AT}^{0}}({\cal D'})$
then may have the following effects:
\begin{enumerate}
\item[(a)] initiating some action effect rules in ${\cal P}^{{\cal AT}^{0}}({\cal D})$;
\item[(b)] initiating some casual rules in ${\cal P}^{{\cal AT}^{0}}({\cal D})$;
\item[(c)] defeating some casual rules in ${\cal P}^{{\cal AT}^{0}}({\cal D})$; 
\item[(d)] not initiating any action effect and causal rules in 
${\cal P}^{{\cal AT}^{0}}({\cal D})$.
\end{enumerate}
Firstly, since each causal proposition in ${\cal D}$ has the form
\begin{quote}
$L$ {\bf is caused if} $L_{1},\cdots, L_{m}$,
\end{quote}
this follows that each corresponding causal rule in
${\cal P}^{{\cal AT}^{0}}({\cal D})$
is non-defeasible, i.e. no negation as failure sign {\em not} is
included in the body. Hence, the
effect (c) will not be presented. On the other hand, since
both causal rules and action effect rules are non-defeasible in 
${\cal P}^{{\cal AT}^{0}}({\cal D})$, 
it is clear that
initiating more action effect rules 
or causal rules in ${\cal AT}^{0}$ will 
{\em not} affect the truth values of literals $Effect^{+}(F,S)$ and
$Caused^{+}(F,S)$ if
${\cal P}^{{\cal AT}^{0}}({\cal D})\models Effect^{+}(F,S)$ 
and ${\cal P}^{{\cal AT}^{0}}({\cal D})\models Caused^{+}(F,S)$ respectively.

Now suppose ${\cal P}^{{\cal AT}^{0}}({\cal D})\models Holds(F,S)$ is due to 
some inertia rule in ${\cal P}^{{\cal AT}^{0}}({\cal D})$:
\begin{quote}
$N: Holds(F,S)\leftarrow Holds(F,S'), not\neg Holds(F,S)$,
\end{quote}
where $S=Result(A,S')$, and ${\cal P}^{{\cal AT}^{0}}({\cal D})\models Holds(F,S')$.
We prove ${\cal P}^{{\cal AT}^{0}}({\cal D'}) \models Holds(F,S)$.

\noindent
\underline{\em Case 1}. Suppose $S=S_{0}$ and
${\cal P}^{{\cal AT}^{0}}({\cal D'}) \models Holds(F,S_{0})$.
Since no inertia rule is needed to drive $Holds(F,S_{0})$, 
the only possibility to have 
${\cal P}^{{\cal AT}^{0}}({\cal D'})\models Holds(F,S_{0})$ 
is either $Holds(F,S_{0})\leftarrow$ is in ${\cal P}^{{\cal AT}^{0}}({\cal D'})$, 
or
$Caused^{+}(F,s)\leftarrow \cdots$ is in 
${\cal P}^{{\cal AT}^{0}}({\cal D'})$ (note that 
such causal rule is non-defeasible). Obviously,
in ${\cal P}^{{\cal AT}^{0}}({\cal D'})$,
the truth value of
$Holds(F,S_{0})$ will not be affected. Hence the result holds.

\noindent
\underline{\em Case 2}.  
Now consider the case that $S$ is not the initial situation.
Suppose ${\cal P}^{{\cal AT}^{0}}({\cal D}) \models Holds(F,S)$. It implies 
that there exists some action constant $A$ such that
${\cal P}^{{\cal AT}^{0}}({\cal D}) \models Holds(F,S')$
due to the inertia rule in ${\cal P}^{{\cal AT}^{0}}({\cal D})$,
where $S=Result(A,S')$.

Now suppose ${\cal P}^{{\cal AT}^{0}}({\cal D'})\not\models Holds(F,S)$.
So the inertia rule:
\begin{quote}
$N': Holds(F,S)\leftarrow Holds(F,S'), not \neg Holds(F,S)$,
\end{quote}
where $S=Result(A,S')$, is defeated
in ${\cal P}^{{\cal AT}^{0}}({\cal D'})$.
Hence it must be the case that
$F\in ({\cal F}^{-}_{Effect}\cup {\cal F}^{-}_{Caused})$.
On the other hand, from the fact that
that ${\cal P}^{{\cal AT}^{0}}({\cal D})\models Holds(F,S')$,
it follows that  
$F\in {\cal F}^{+}_{Initial}$ or
$F\in ({\cal F}^{+}_{Effect}\cup {\cal F}^{+}_{Caused})$.
But this contradicts conditions of Theorem 7.
So it must have ${\cal P}^{{\cal AT}^{0}}({\cal D'})\models Holds(F,S)$.
\end{proof}

\noindent
{\bf Lemma 1}
{\em
A domain description ${\cal D}$ of
language ${\cal AT}^{i}$ ($i=0,1,2$) is temporally definite
if its PLP translation
${\cal P}^{{\cal AT}^{i}}({\cal D})$ has a unique answer set.
}

\noindent
\begin{proof}
Let ${\cal P}^{{\cal AT}^{i}}({\cal D})$ be the PLP translation of
${\cal D}$. 
From the definition of  temporal definiteness, it is sufficient to prove that
${\cal P}^{{\cal AT}^{i}}({\cal D})\models Holds(F,S)$ implies
${\cal P}^{{\cal AT}^{i}}({\cal D})\models Holds(F,Result(A,S))$ or 
${\cal P}^{{\cal AT}^{i}}({\cal D})\models \neg Holds(F,Result(A,S))$ 
for any action constant
$A$. Suppose ${\cal P}^{{\cal AT}^{i}}({\cal D})\models Holds(F,S)$ and 
${\cal P}^{{\cal AT}^{i}}({\cal D})$ has a unique answer set $Ans^{i}$. So 
$Holds(F,S)\in Ans^{i}$. Then, it is clear that
if one of the following cases holds, the result is true:
\begin{enumerate}
\item[(1)] $Effect^{+}(F,Result(A,S))$ or $Effect^{-}(F,Result(A,S))$
is in $Ans^{i}$;
\item[(2)] $AbEffect^{+}(F,Result(A,S))$ or
$AbEffect^{-}(F,Result(A,S))$ is in $Ans^{i}$, here
$i=2$; 
\item[(3)] $Caused^{+}(F,Result(A,S))$ or
$Caused^{-}(F,Result(A,S))$ is in $Ans^{i}$; 
\item[(4)] $\neg Holds(F,Result(A,S))$ is in $Ans^{i}$.
\end{enumerate}
Now suppose none of the above cases is held. Then from the 
instance of inertia rule in ${\cal P}^{{\cal AT}^{i}}({\cal D})$:
\begin{quote}
$Holds(F,Result(A,S))\leftarrow Holds(F,S), not \neg Holds(F,Result(A,S))$,
\end{quote}
it follows that $Holds(F,Result(A,S))$ is in 
$Ans^{i}$. So the result is still true.
\end{proof}

\noindent
{\bf Theorem 8}
{\em 
A domain description ${\cal D}$ of ${\cal AT}^{i}$ ($i=0, 1, 2$) is
temporally definite if
its PLP translation ${\cal P}^{{\cal AT}^{i}}({\cal D})$
has a unique reduct $\Delta^{i}$ and
$Trans(\Delta^{i})$ is locally stratified.
}

\noindent
\begin{proof}
The proof is directly from Lemma 1 and Proposition
3 in Appendix A.
\end{proof}

\bibliography{tplp02R9}

\comment{

\newpage

}

\end{document}